\documentclass[10pt,twocolumn,letterpaper]{article}

\usepackage[accsupp]{axessibility}
\usepackage{iccv}              
\newcommand*{\ARXIV}{}%

\usepackage{graphicx}
\usepackage{tikz}
\usepackage{amsmath}
\usepackage{amssymb}
\usepackage{booktabs}
\usepackage{multirow}

\usepackage{color,xcolor}
\usepackage{epsfig}
\usepackage{graphicx}
\usepackage{times}

\usepackage{comment}

\usepackage{pifont}

\usepackage{microtype}
\usepackage[font=small]{caption}
\usepackage{arydshln}
\usepackage{tabularx}
\usepackage{adjustbox}
\usepackage{array}
\usepackage{booktabs}
\usepackage{colortbl}
\usepackage{float,wrapfig}
\usepackage{hhline}
\usepackage{multirow}
\usepackage{subcaption} %
\usepackage[percent]{overpic}

\usepackage{stmaryrd}
\usepackage{breqn}

\usepackage{amsmath,amsfonts,amssymb}

\usepackage{bm}
\usepackage{nicefrac}
\usepackage{microtype}
\usepackage{dsfont}
\usepackage{changepage}
\usepackage{extramarks}
\usepackage{fancyhdr}
\usepackage{setspace}
\usepackage{soul}
\usepackage{xspace}
\usepackage{hhline}
\usepackage{algorithmicx}
\usepackage{algorithm}
\usepackage{algpseudocode}
\usepackage{pifont}
\usepackage{amssymb}

\usepackage[pagebackref,breaklinks,colorlinks]{hyperref}

\usepackage{enumitem}
\usepackage[title]{appendix}

\def\L{\mathcal{L}} %

\def\G{{G}}

\newcommand{\reffig}[1]{Figure~\ref{fig:#1}}
\newcommand{\refsec}[1]{Section~\ref{sec:#1}}
\ifdefined\ARXIV
\newcommand{\refapp}[1]{Appendix~\ref{sec:#1}}
\else
\newcommand{\refapp}[1]{Section~\ref{sec:#1}}
\fi
\newcommand{\reftbl}[1]{Table~\ref{tbl:#1}}

\newcommand{\refeq}[1]{Eqn.~\ref{eq:#1}}

\newcommand{\lblalg}[1]{\label{alg:#1}}

\newcommand{\ignorethis}[1]{}

\newcommand{\myparagraph}[1]{\vspace{0pt} \noindent \textbf{#1} \ }

\def\1{\bm{1}}

\newcommand{\R}{\mathbb{R}}

\newcolumntype{L}[1]{>{\raggedright\let\newline\\\arraybackslash\hspace{0pt}}m{#1}}
\newcolumntype{C}[1]{>{\centering\let\newline\\\arraybackslash\hspace{0pt}}m{#1}}
\newcolumntype{R}[1]{>{\raggedleft\let\newline\\\arraybackslash\hspace{0pt}}m{#1}}

\newcommand{\ignore}[1]{}

\makeatletter
\DeclareRobustCommand\onedot{\futurelet\@let@token\@onedot}
\def\@onedot{\ifx\@let@token.\else.\null\fi\xspace}
\def\eg{\emph{e.g}\onedot,\xspace}

\def\ie{\emph{i.e}\onedot,\xspace}

\def\etc{\emph{etc}\onedot} 
 
\def\etal{\emph{et al}\onedot}
\makeatother

\usepackage{duckuments}
\usepackage[capitalize]{cleveref}
\crefname{section}{Sec.}{Secs.}
\Crefname{section}{Section}{Sections}
\Crefname{table}{Table}{Tables}
\crefname{table}{Tab.}{Tabs.}

\iccvfinalcopy 

\newcommand{\figablationalign}{
\begin{figure}[]
\centering
\includegraphics[width=1.0\linewidth]{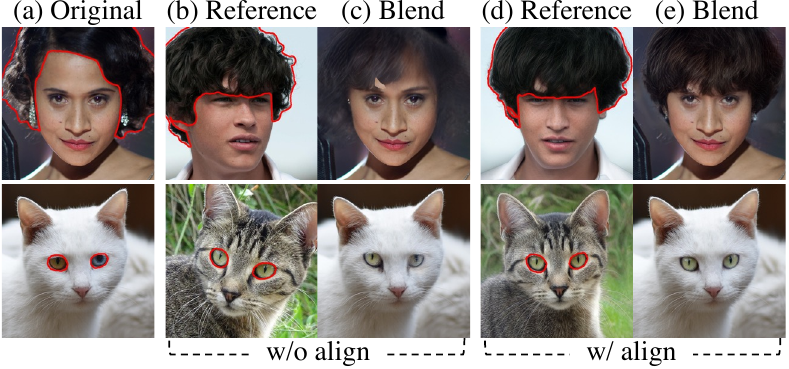}
\caption{{\bf The effect of our 3D-aware alignment}. Aligned reference images (d) have the same pose as the original images (a). With our alignment, blending results (e) look more realistic and reflect the reference well than those without alignment (c).
} 
\label{fig:ablationalign}
\end{figure}
}
\newcommand{\densityResult}{
\begin{figure}[]
\centering
\includegraphics[width=1.0\linewidth]{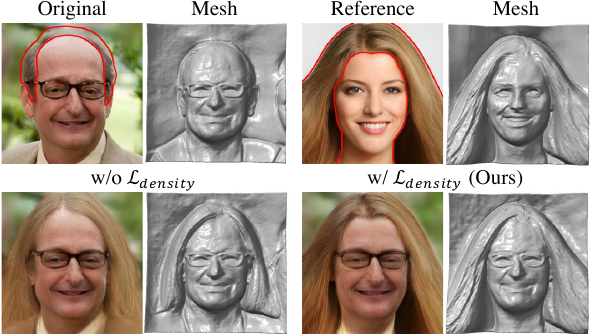} 
\caption{{\bf The effect of our density-blending loss}. 
Without the loss, 3D information is not considered, resulting in inaccurate blending in 3D space. In the bottom left figure, the hair mesh is not properly reflected without the density-blending loss, resulting in inaccurate blending and missing fine details.} 
\label{fig:densityResult}
\end{figure}
}
\newcommand{\figOursWWOPB}{
\begin{figure}[]
\centering

\includegraphics[width=1.0\linewidth]{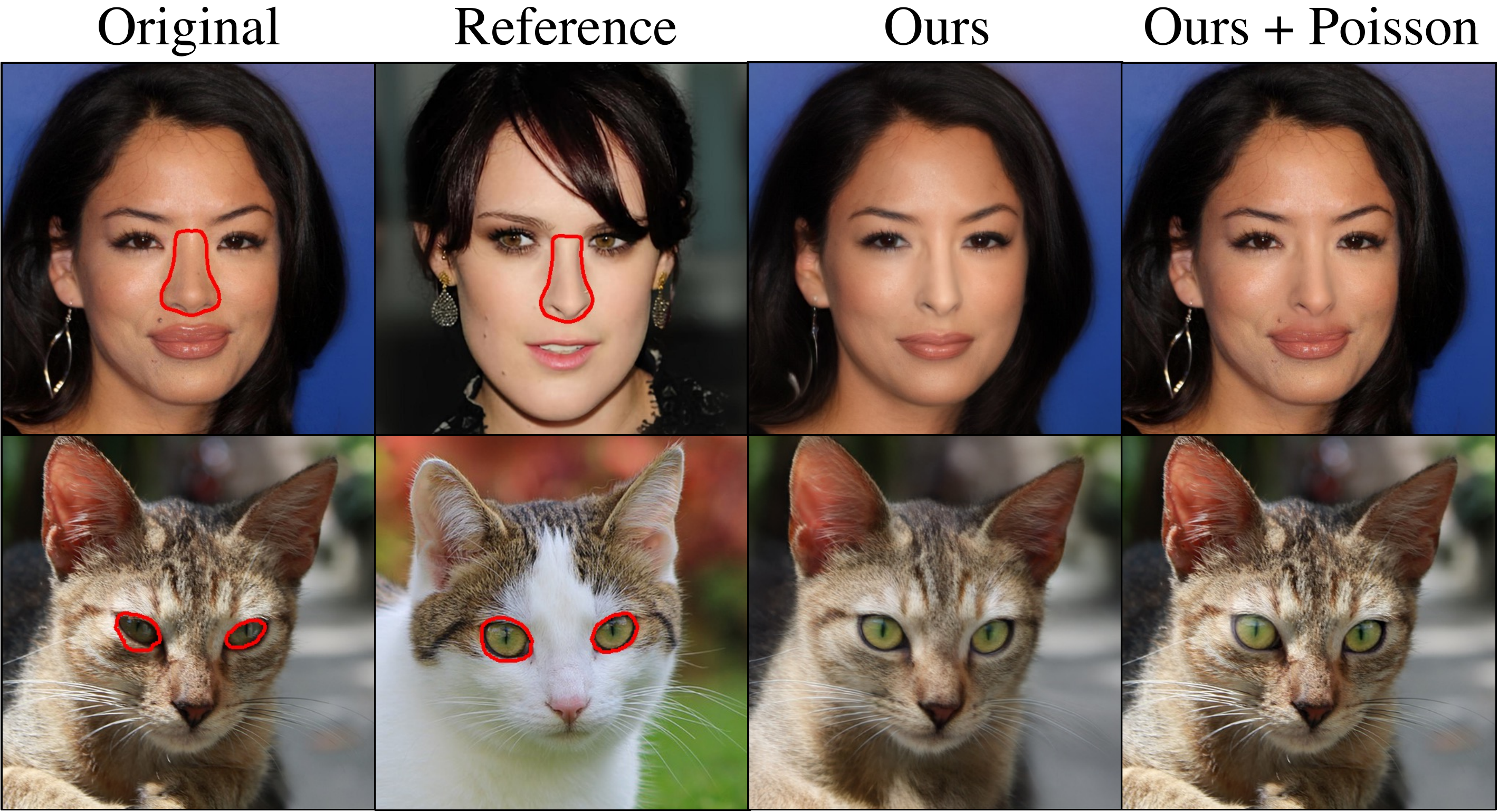}\\
\caption{
\textbf{Ours with Poisson blending}~\cite{perez2003poisson}.
Ours shows satisfying blending results but a lack of preservation in details. In the first row, the earring is missing in our method. The high-frequency details such as hair and fur are less pronounced in our method alone compared to when it is combined with Poisson blending. 
} 
\label{fig:figOurs_w_wo_PB}
\end{figure}
}

\newcommand{\figmotivation}{
\begin{figure}[]
\centering
\includegraphics[width=1.0\linewidth]{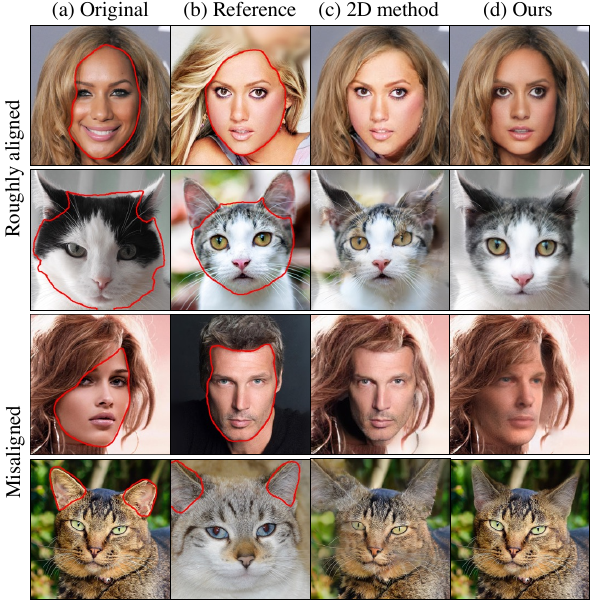} 
\caption{
Image blending is challenging for unaligned original and reference images. 
Existing 2D-based methods~\cite{kim2021exploiting} struggle to synthesize realistic results due to the 3D object pose differences between foreground and background.  
In contrast, we propose a 3D-aware blending method that aligns and composes unaligned images without manual effort. 
} 
\label{fig:motivation}
\end{figure}
}

\newcommand{\figglobalalign}{
\begin{figure}[]
\centering
\includegraphics[width=1.0\linewidth]{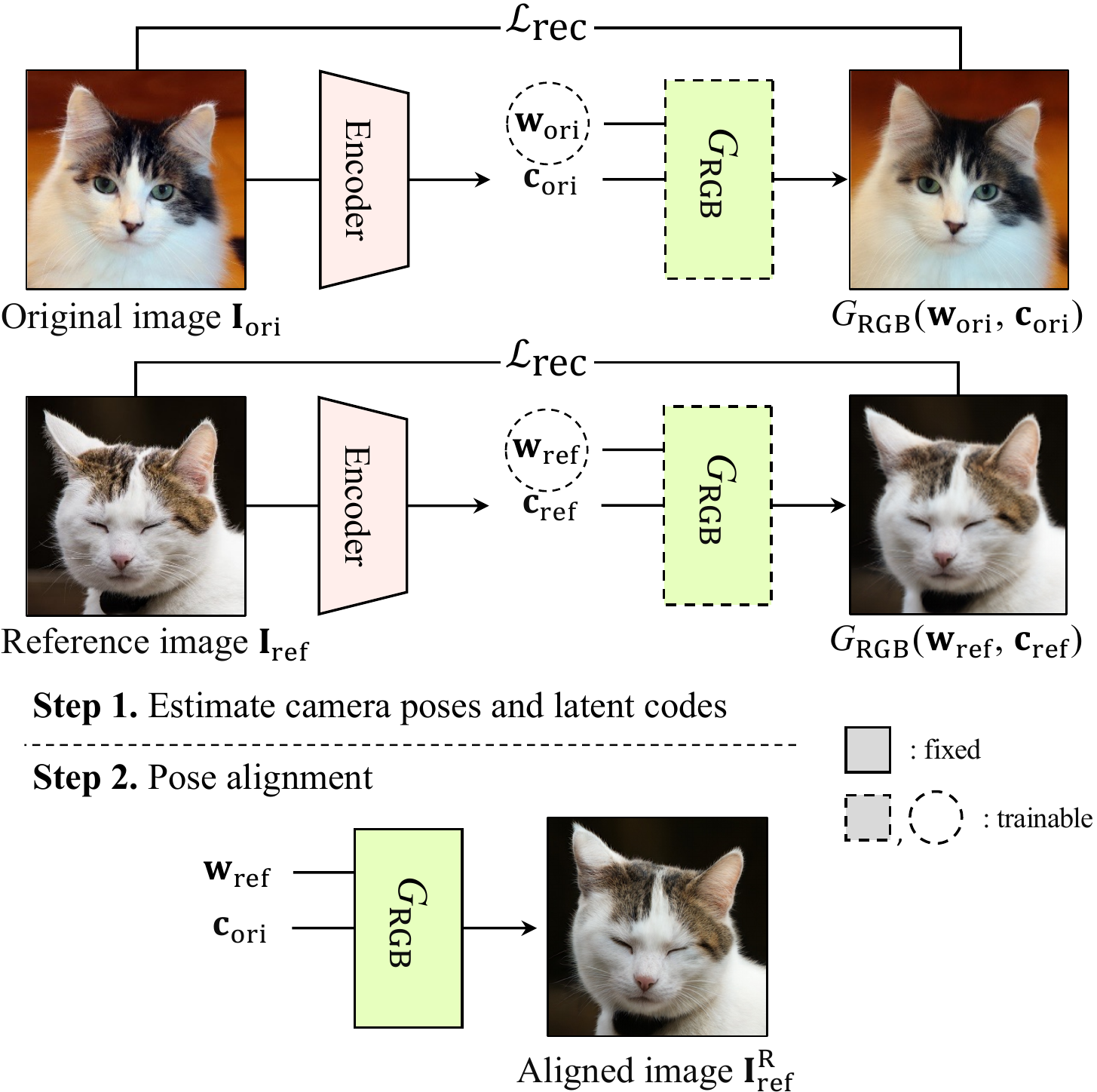} 
\caption{{\bf 3D-aware alignment}: we first use a CNN encoder to infer the camera pose of each input image. \textbf{Step 1.} Given the camera pose $\cam{}$, we estimate the latent code $\w{}$ for each input using a reconstruction loss $\mathcal{L}_{\text{rec}}$. \textbf{Step 2.} Given the estimated camera pose $\cam{ori}$ and latent code $\w{ref}$, we align the reference image to match the pose of the original image.
} 
\label{fig:figglobalalign}
\end{figure}
}

\newcommand{\figblending}{
\begin{figure*}[]
\centering
\includegraphics[width=1.0\linewidth]{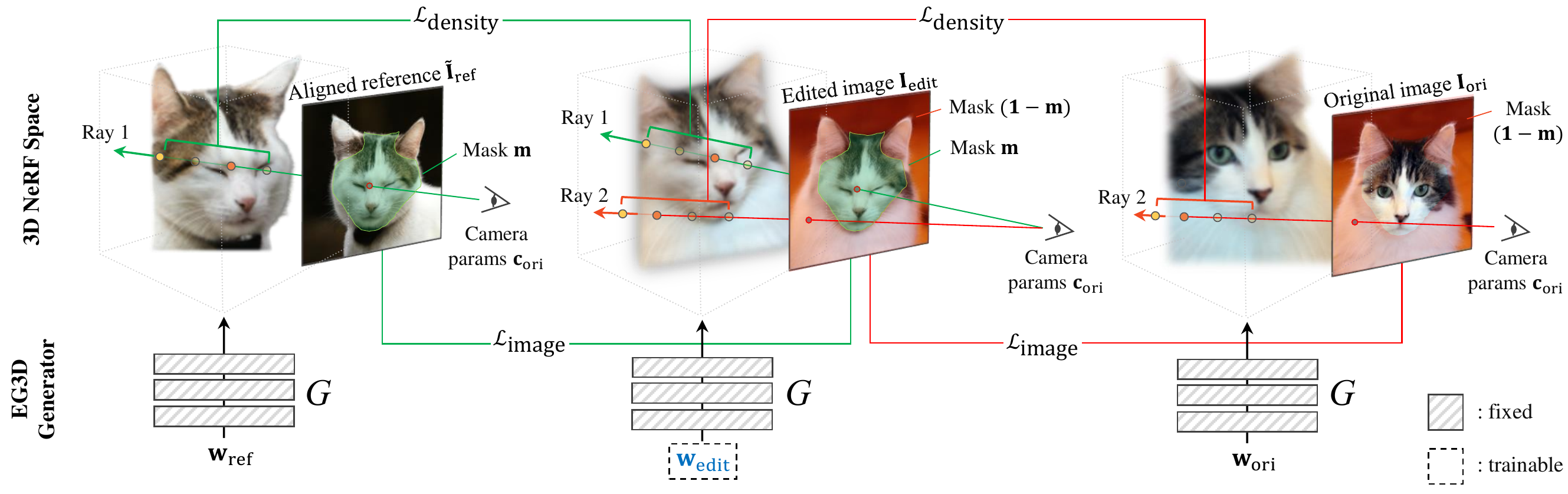}\\
\caption{{\bf Our 3D-aware blending pipeline.} 
We employ density-blending loss ($\L{density}$) in the volume density of 3D NeRF space, as well as the image-blending loss ($\L{image}$) in 2D image space. 
\textcolor[RGB]{31,196,111}{Green rays} pass through the interior of the mask ($\mask{}$) and \textcolor{red}{red rays} pass through the exterior of the mask ($\bm{1}-\mask{}$). $\L{image}$ and $\L{density}$ are used to optimize the latent code $\w{edit}$ to generate the well-blended image $\I{edit}$. 
} 

\label{fig:blending}
\end{figure*}
}

\newcommand{\figcomparison}{
\begin{figure*}[]
\centering
\includegraphics[width=1.0\linewidth]{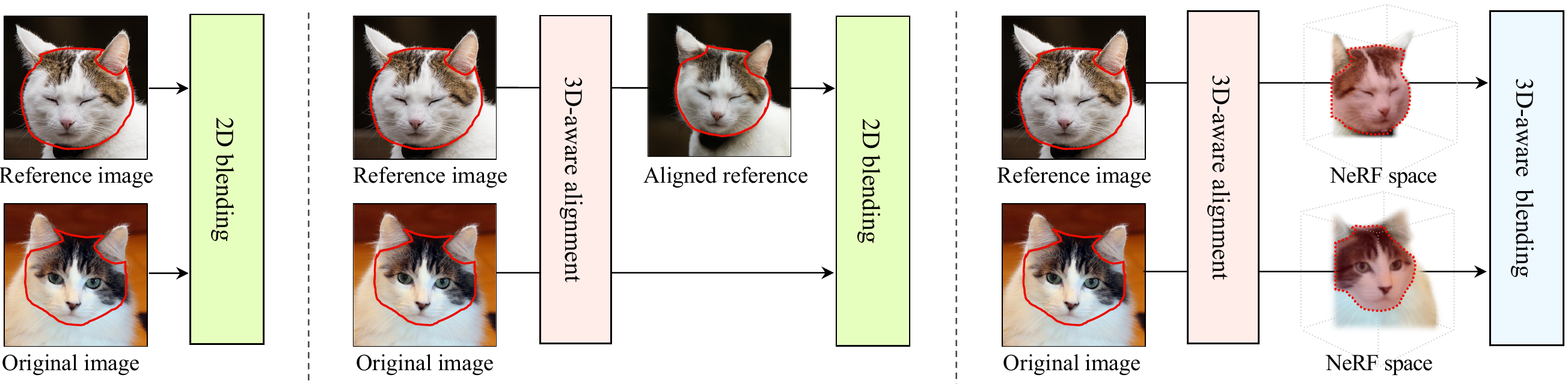}\\
\makebox[0.2\linewidth][c]{\footnotesize{\textbf{(a)} 2D blending}}\hfill
\makebox[0.4\linewidth][c]{\footnotesize{\textbf{(b)} 2D blending with our 3D-aware alignment}}\hfill
\makebox[0.4\linewidth][c]{\footnotesize{\textbf{(c)} Proposed method (Ours)}}\hfill
\\
\caption{{\bf Comparison with the existing blending methods.} Red lines denote target blending parts. \textbf{(a) 2D blending.} 2D blending methods compose two images without any 3D-aware alignment. \textbf{(b) 2D blending with 3D-aware alignment.} To address misalignment, we apply our 3D-aware alignment method to existing 2D blending methods. \textbf{(c) Proposed method.} We propose 3D-aware blending after applying our 3D-aware alignment. Note that all methods do not use 3D labels or 3D morphable models.} 
\label{fig:comparison}
\end{figure*}
}

\newcommand{\figdisentangle}{
\begin{figure}[]
\centering
\includegraphics[width=1.0\linewidth]{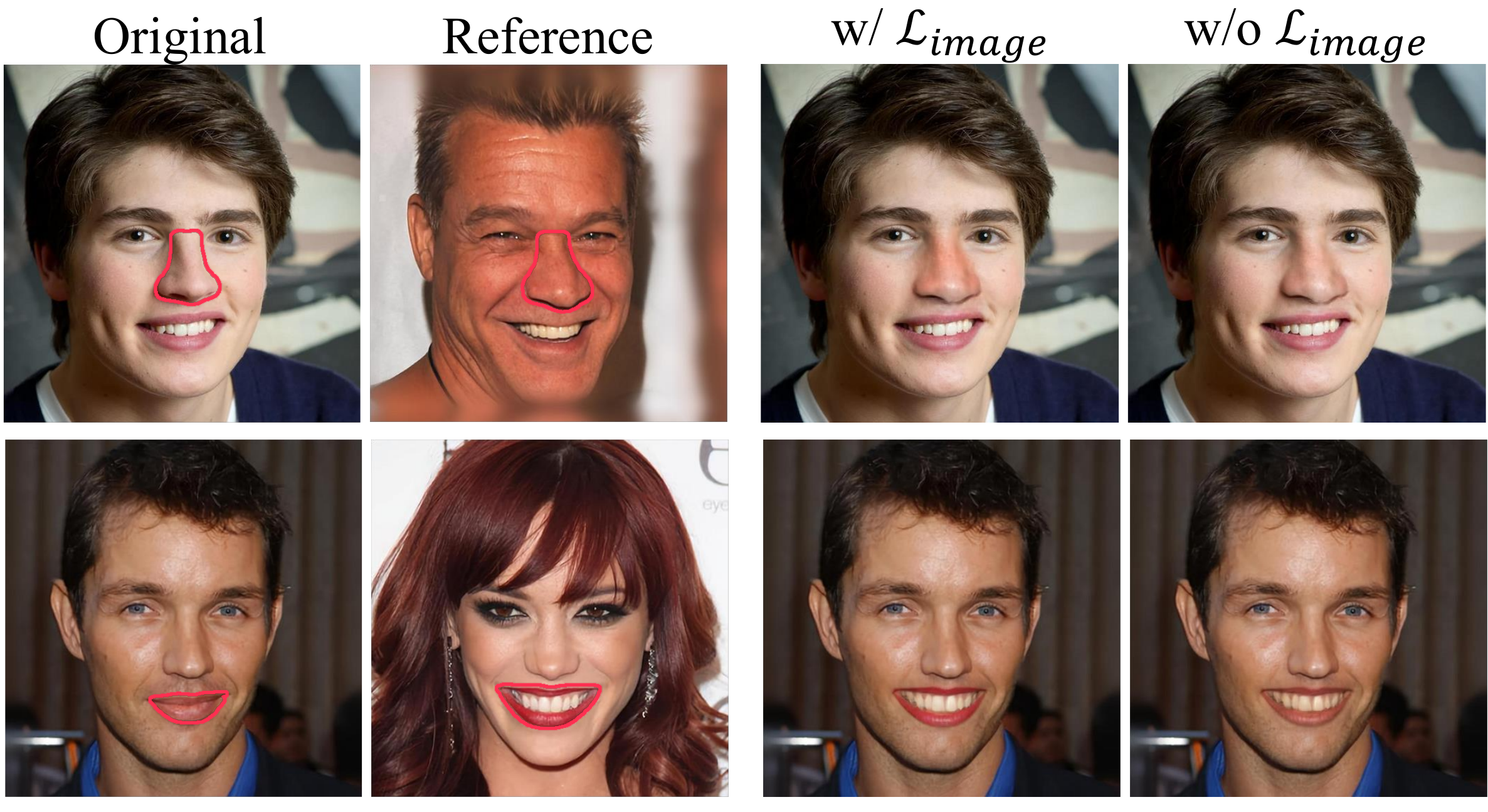} 
\caption{{\bf Color-geometry disentanglement with our model}. We can adjust the reflection of the reference image's color by adjusting the weight $\lambda_2$ in the image-blending loss. Without image blending loss on reference, we can focus on object shapes, as shown in the rightmost column.
} 
\label{fig:disentangle}
\end{figure}
}

\newcommand{\figmultiview}{
\begin{figure}[]
\centering
\includegraphics[width=1.0\linewidth]{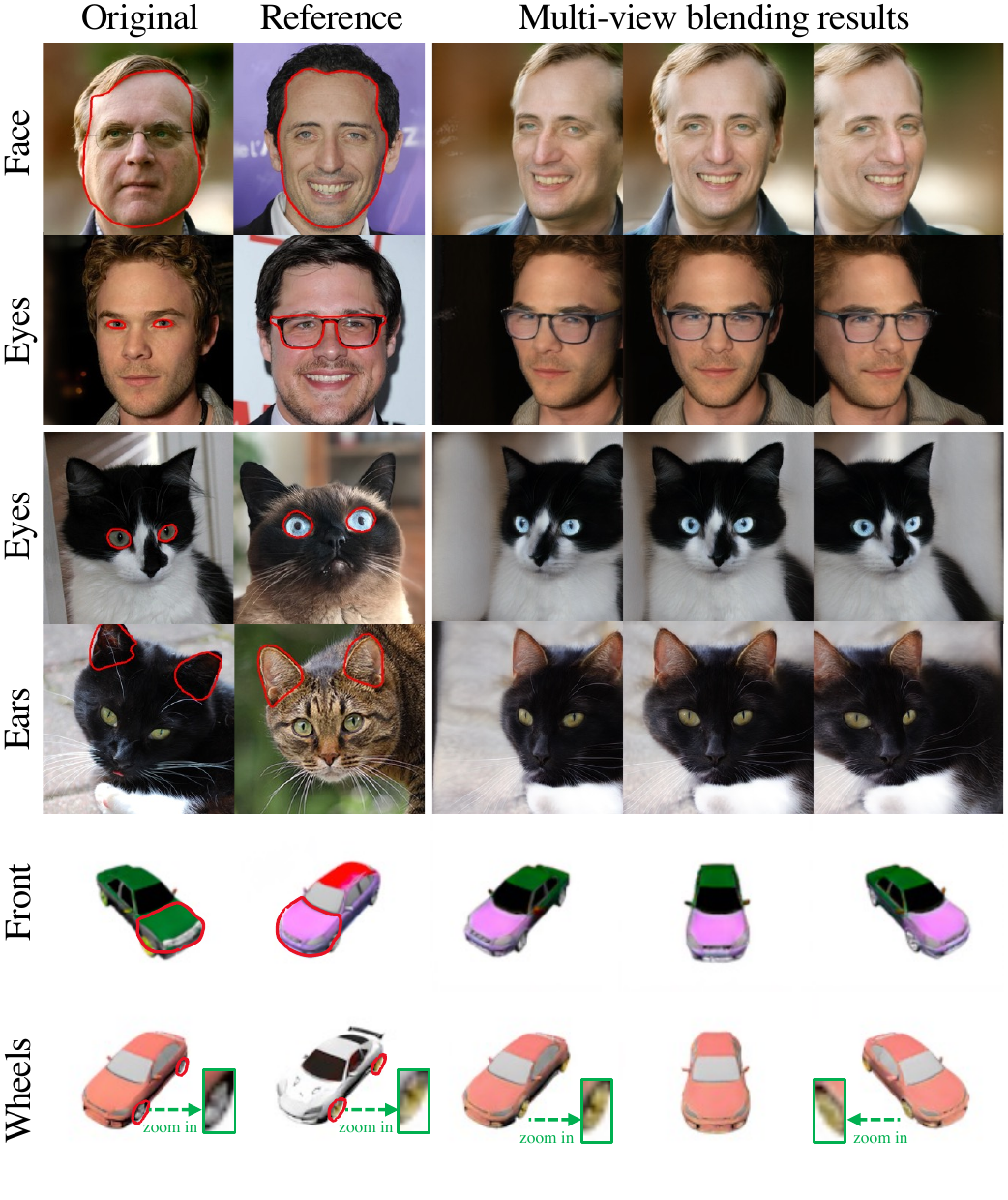} 
\caption{{\bf Multi-view blending results} in various datasets: CelebA-HQ, AFHQv2-Cat, and ShapeNet-Car. Since we optimize the latent code of the generative NeRF, we can synthesis images of the blended object in different poses through the generative NeRF. 
} 
\label{fig:multiview}
\end{figure}
}

\newcommand{\figLimitInv}{
\begin{figure}[]
\centering
\includegraphics[width=0.95\linewidth]{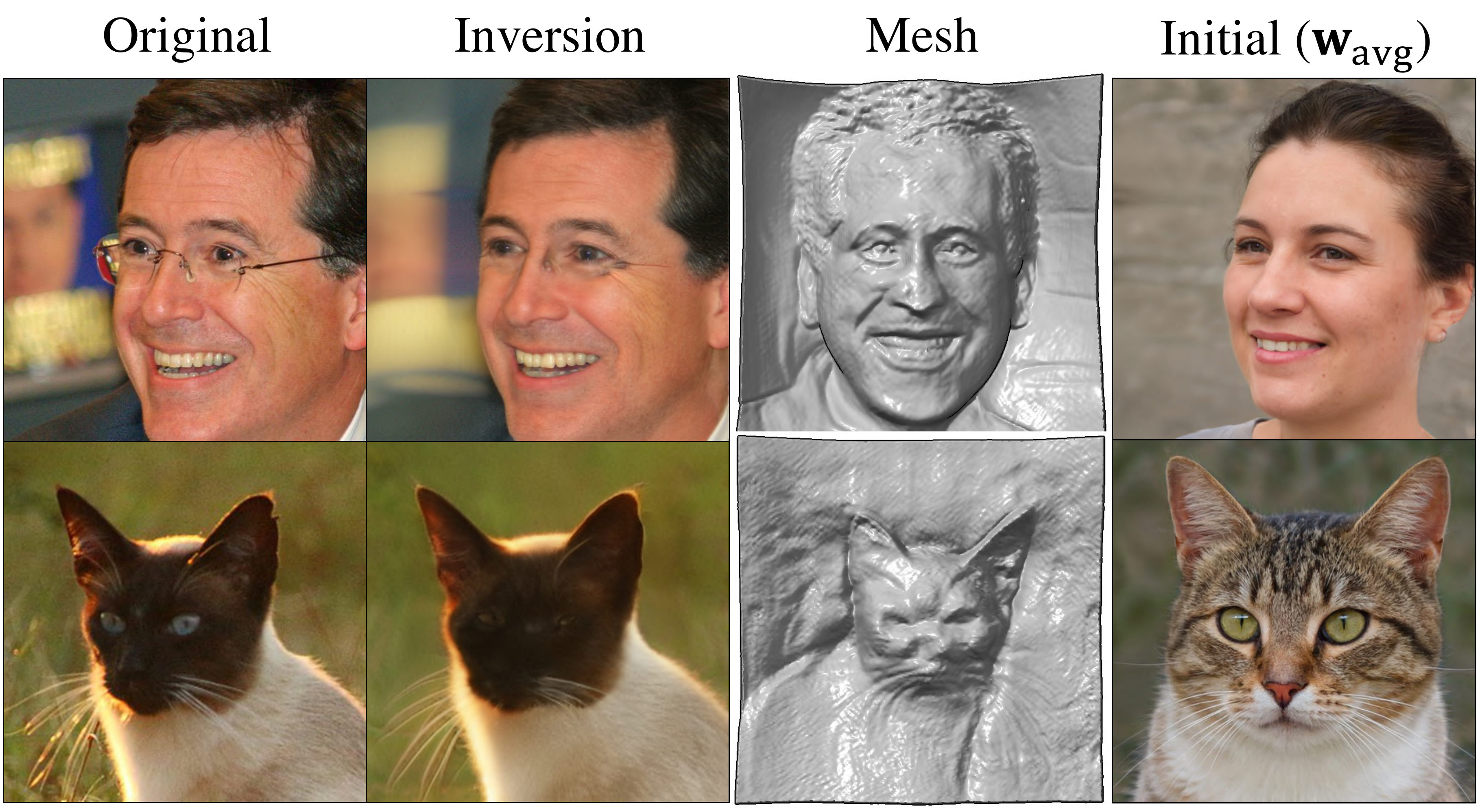}\\
\caption{Failure cases of inversion.
If an input image has a large variance in scale relative to the mean face or the estimated pose from the encoder is not valid, inversion sometimes fails. The first row shows a failure to reconstruct eyeglasses, and the second row shows a crushed face of a cat in the reconstructed image and mesh.
} 
\ifdefined\ARXIV
\else
\vspace{-2mm}
\fi
\label{fig:figLimitInv}
\end{figure}
}

\newcommand{\figPoissonBaseline}{
\begin{figure*}[t]
\centering
\includegraphics[width=0.8\linewidth]{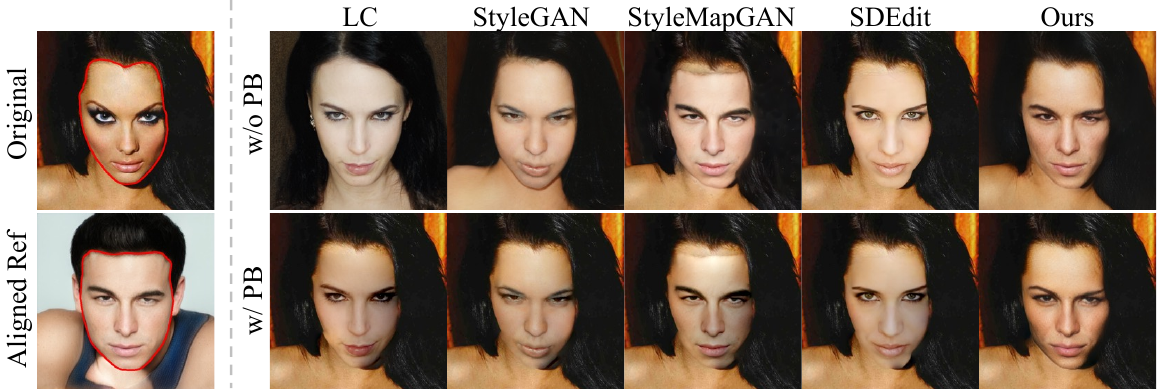}
\caption{
Ablation study of Poisson blending (PB) in baselines. Despite combining Poisson blending with the baselines, StyleMapGAN still generates artifacts, and other baselines fail to preserve the identity of the reference. Our method with Poisson blending keeps the original image intact while accurately reflecting the reference image.
}
\label{fig:poisson_baseline}
\end{figure*}
}

\newcommand{\tabBaselineEditCelebAHQ}{
\newcommand{\himg}{0.117}
\newcommand{\himgtwo}{0.12}
\begin{table*}[t!]
\centering
\begin{minipage}{\linewidth}
    \centering
    \hspace{3mm}
    \makebox[\himg\linewidth][c]{~~\small{Original}}
    \makebox[\himg\linewidth][c]{~~\small{Reference}}
    \makebox[\himg\linewidth][c]{~\small{Poisson}}\hfill
    \makebox[\himg\linewidth][c]{\small{LC}}\hfill
    \makebox[\himg\linewidth][c]{\small{StyleGAN3 $\mathcal{W}$~~}}\hfill
    \makebox[\himg\linewidth][c]{\small{StyleMapGAN}~~}\hfill 
    \makebox[\himg\linewidth][c]{\small{SDEdit}}\hfill 
    \makebox[\himg\linewidth][c]{\small{Ours + PB~~}}\hfill 
    
    \hspace{-3.5mm}
    \rotatebox{90}{\makebox[20mm][c]{\small{~~~~w/o align}}}\vspace{-2.4mm}\hspace{0.5mm}
    \includegraphics[width=\himg\linewidth]{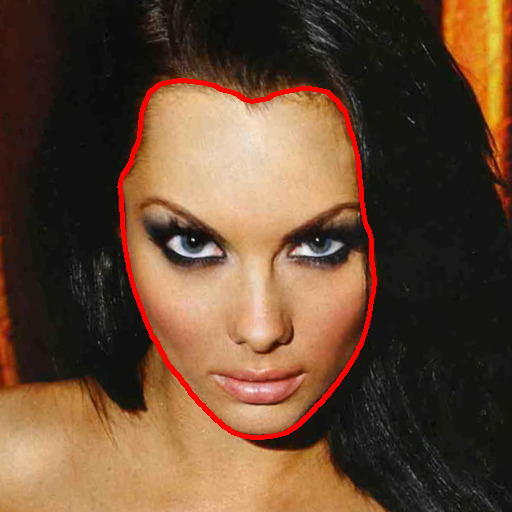}
    \includegraphics[width=\himg\linewidth]{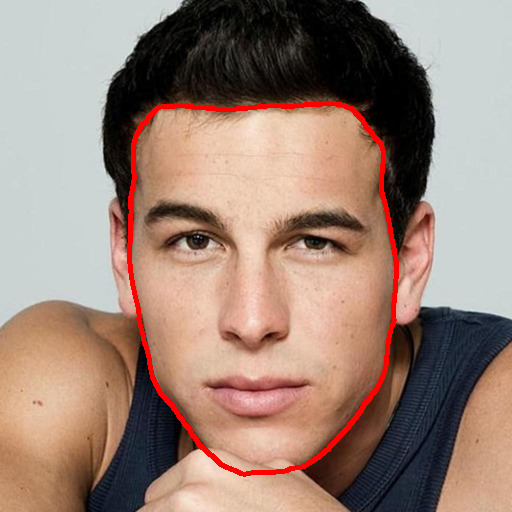}
    \includegraphics[width=\himg\linewidth]{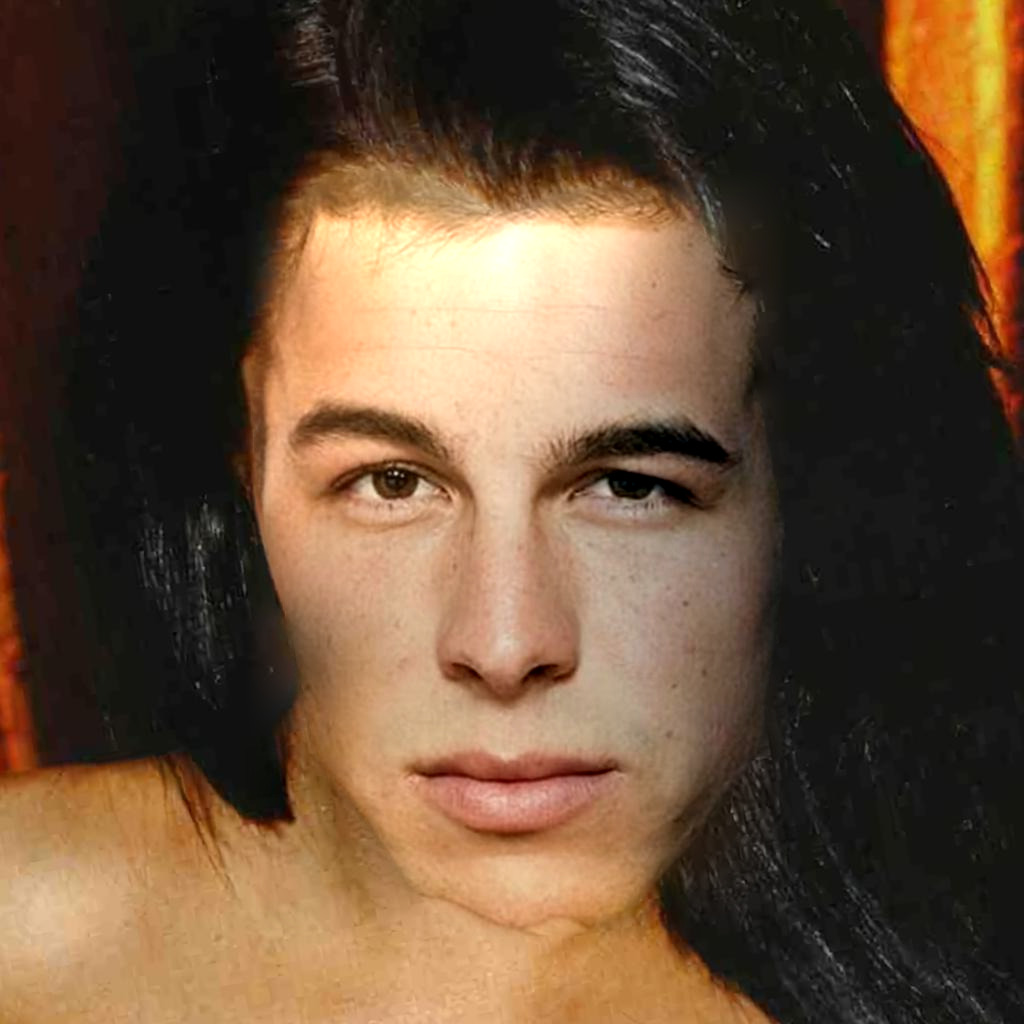}
    \includegraphics[width=\himg\linewidth]{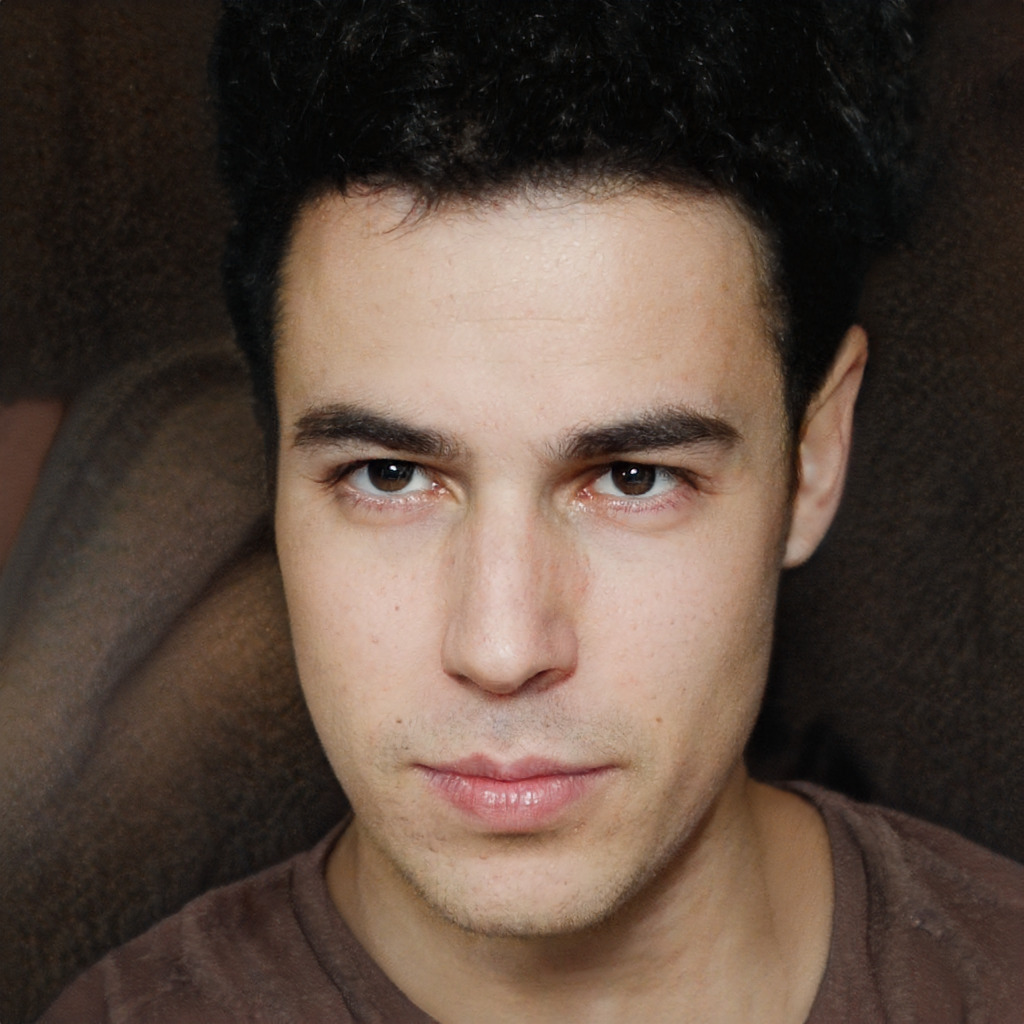}
    \includegraphics[width=\himg\linewidth]{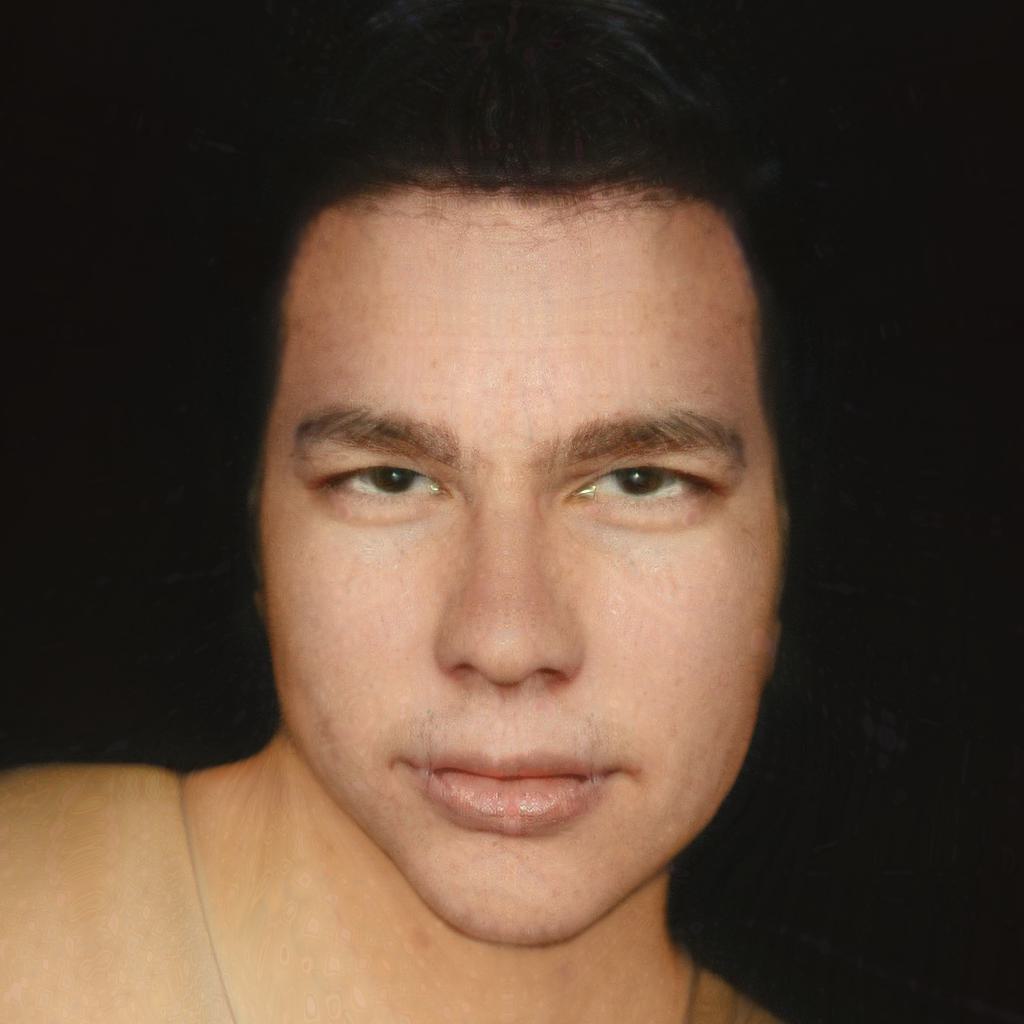}
    \includegraphics[width=\himg\linewidth]{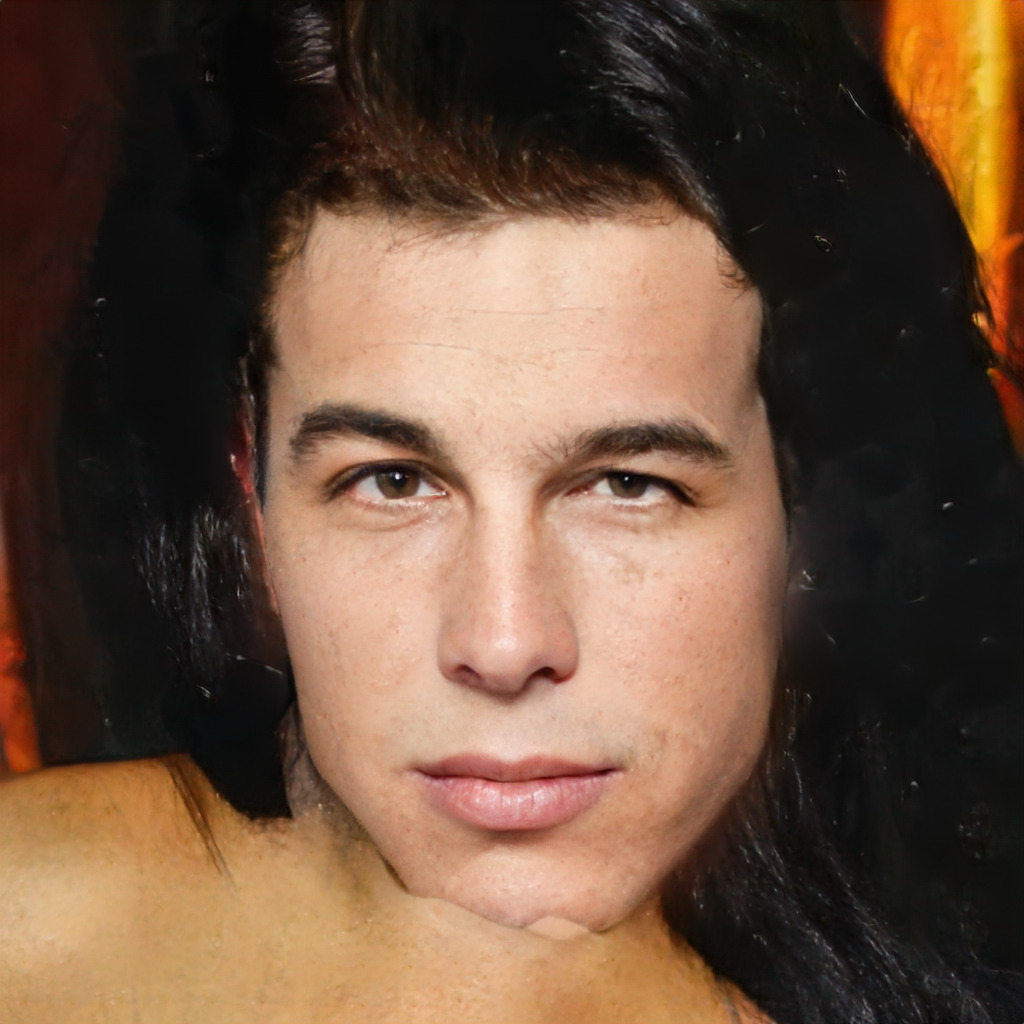}
    \includegraphics[width=\himg\linewidth]{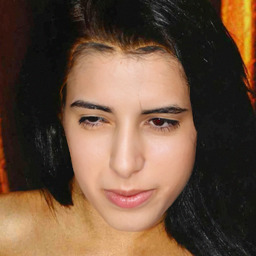}
      \begin{tikzpicture}
        \node[draw=black, text centered, minimum width=2.03cm, minimum height=2.03cm, inner sep=0pt, line width=0.2pt] {\texttt{N/A}};
      \end{tikzpicture}%
    \vspace{2mm}
    \hspace{-2.5mm}
    
    \rotatebox{90}{\makebox[20mm][c]{\small{w/ align}}}
    \vspace{0.0mm}\hspace{-0.5mm}
    \includegraphics[width=\himg\linewidth]{figures/table1/tab1_ori.jpg}
    \includegraphics[width=\himg\linewidth]{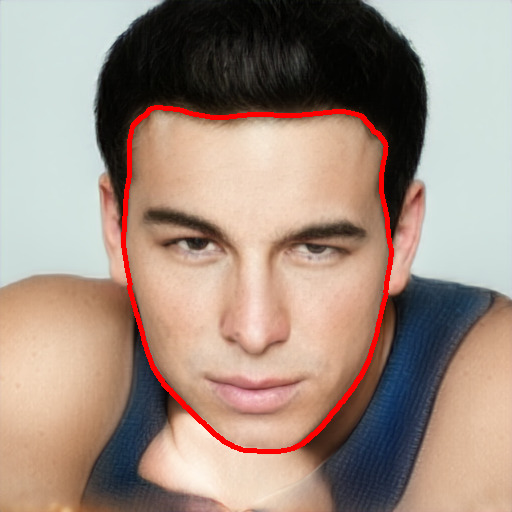}
    \includegraphics[width=\himg\linewidth]{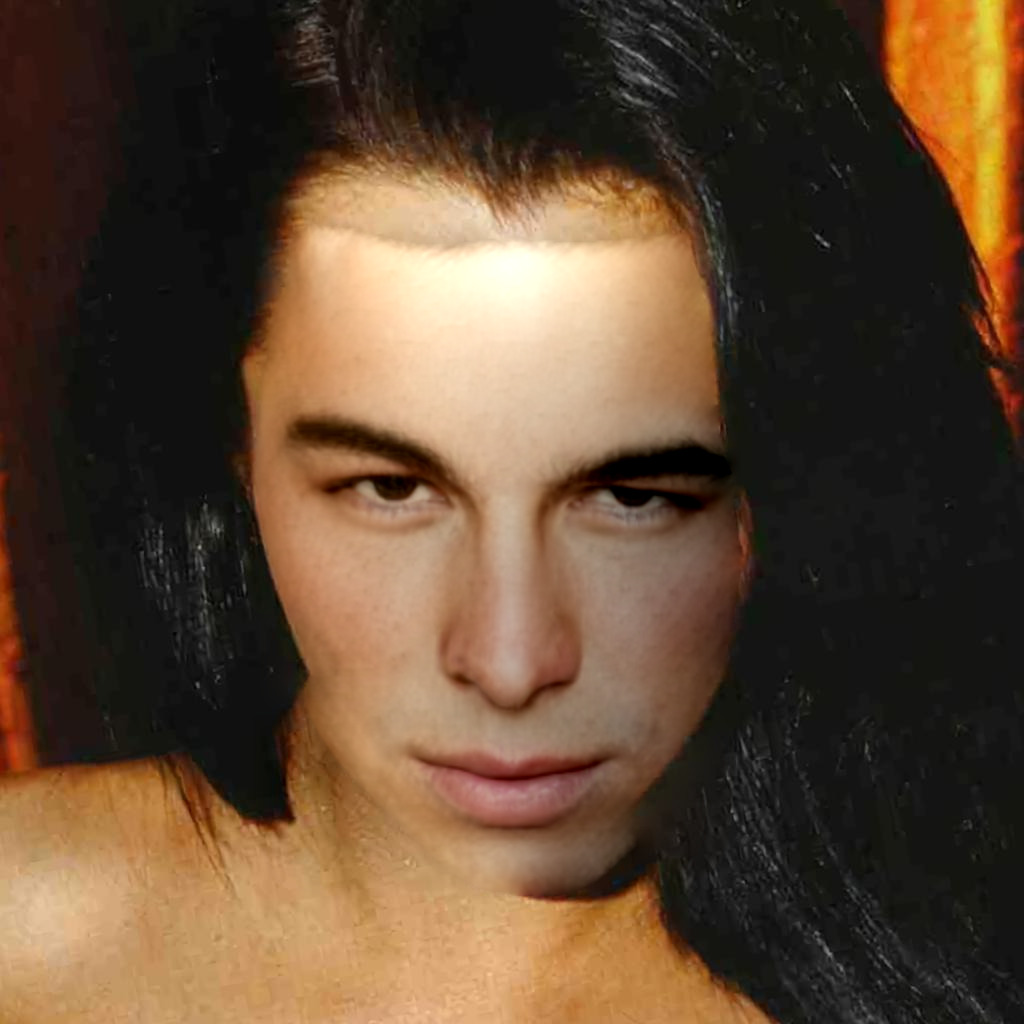}
    \includegraphics[width=\himg\linewidth]{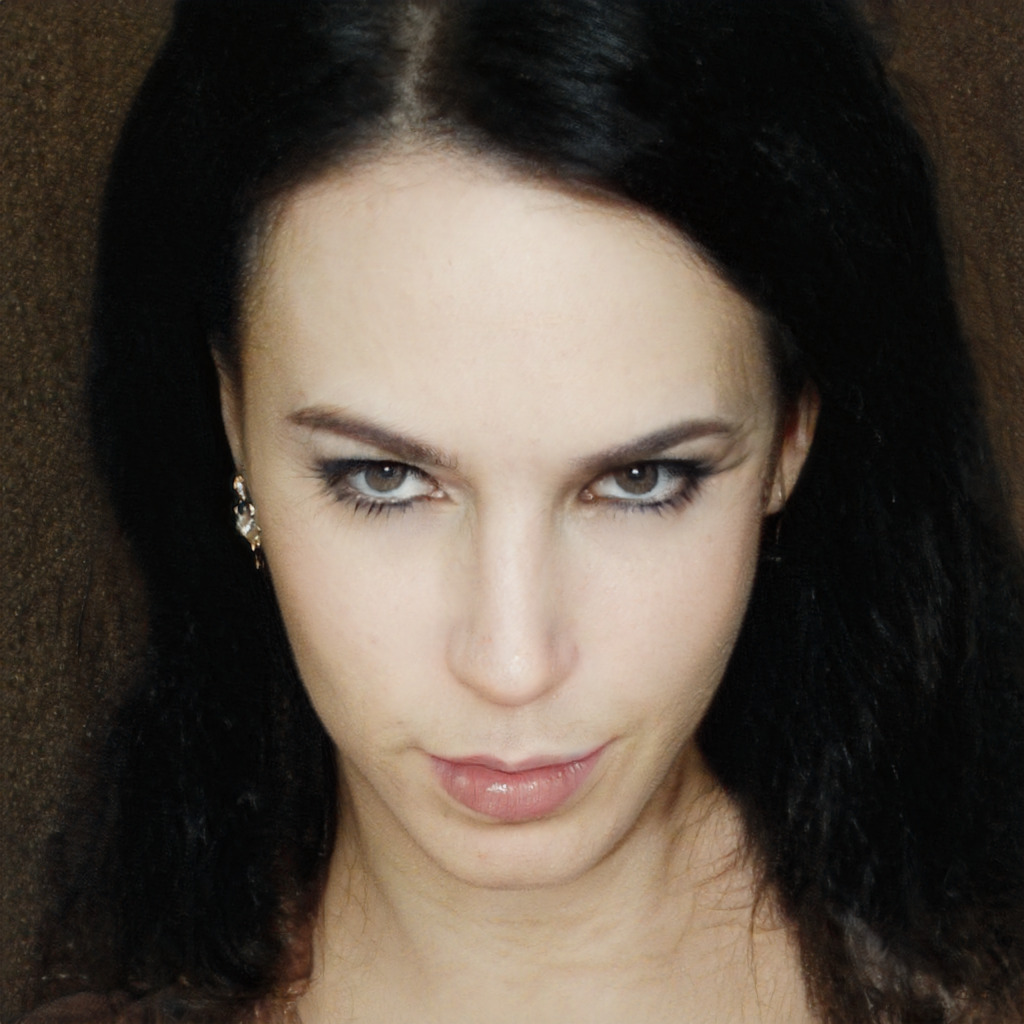}
    \includegraphics[width=\himg\linewidth]{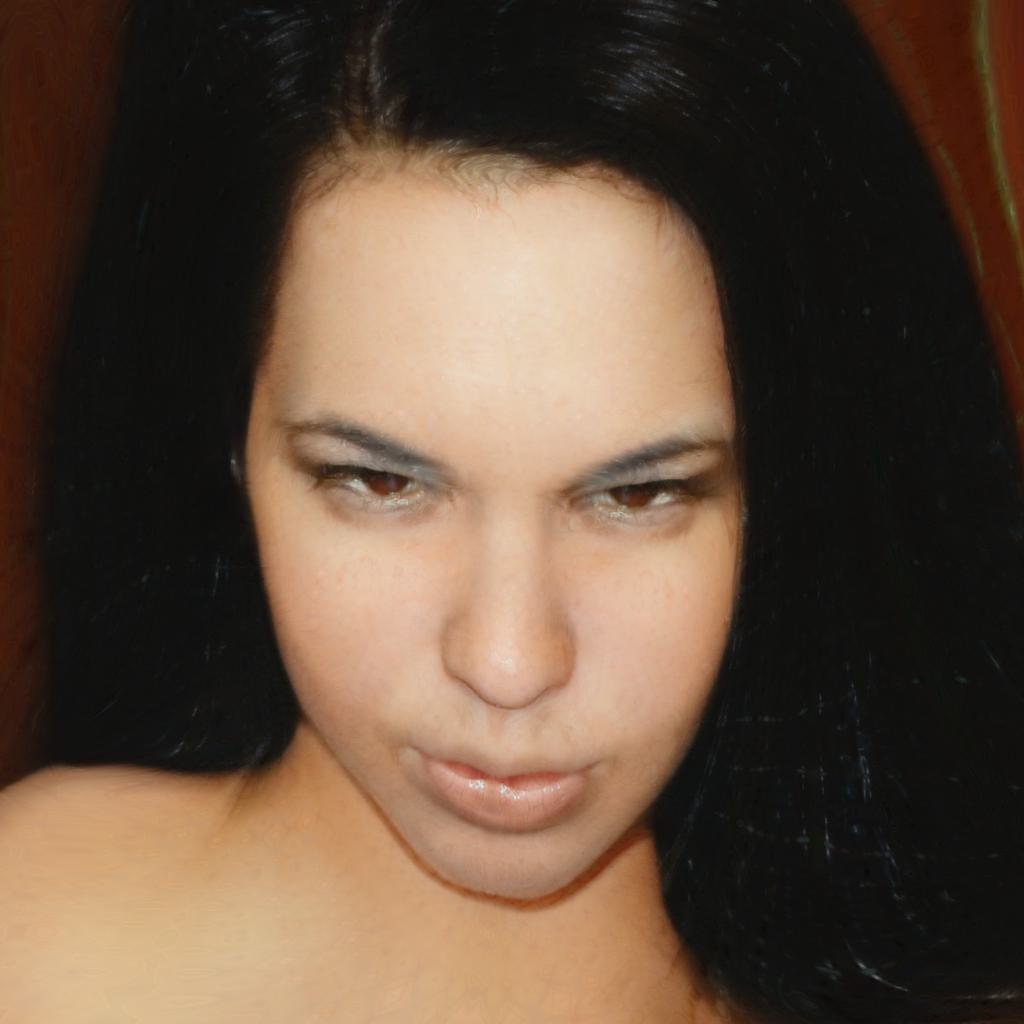}
    \includegraphics[width=\himg\linewidth]{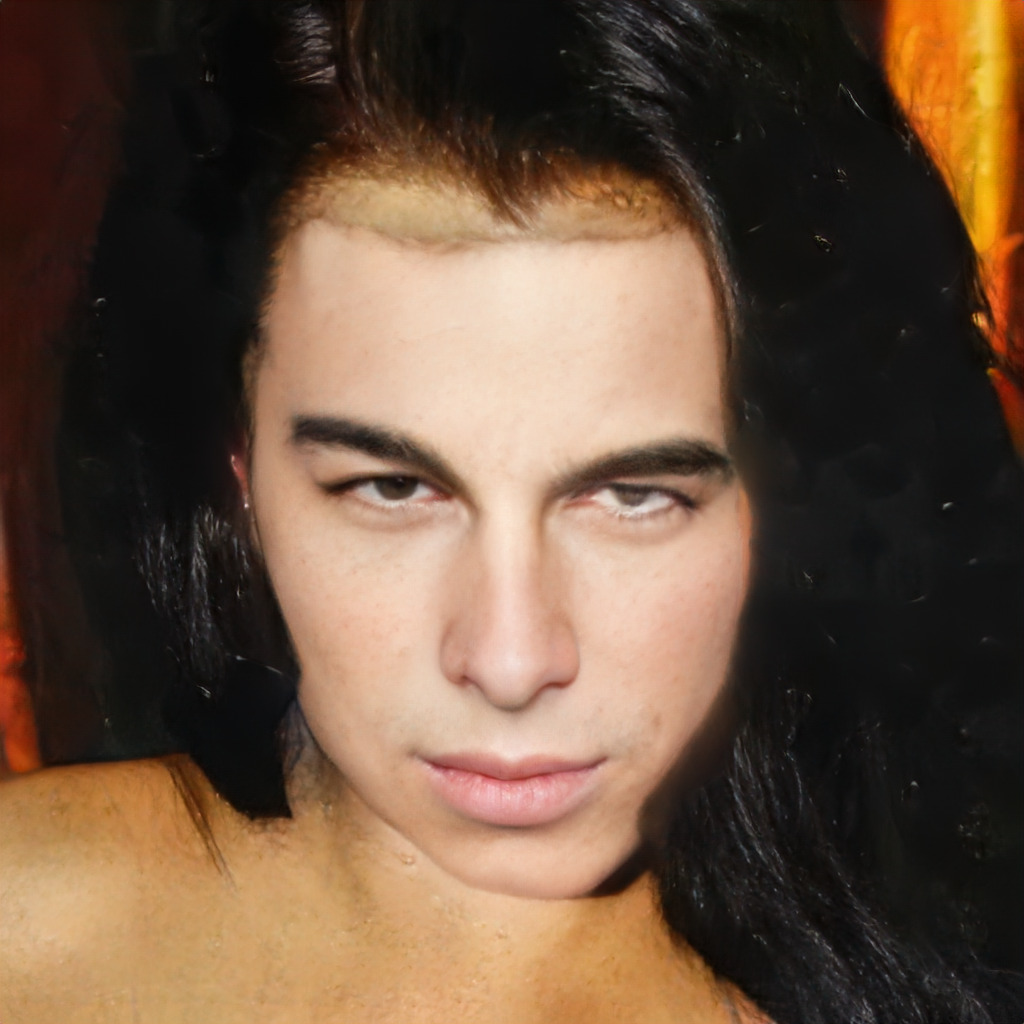}
    \includegraphics[width=\himg\linewidth]{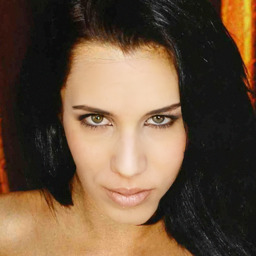}
    \includegraphics[width=\himg\linewidth]{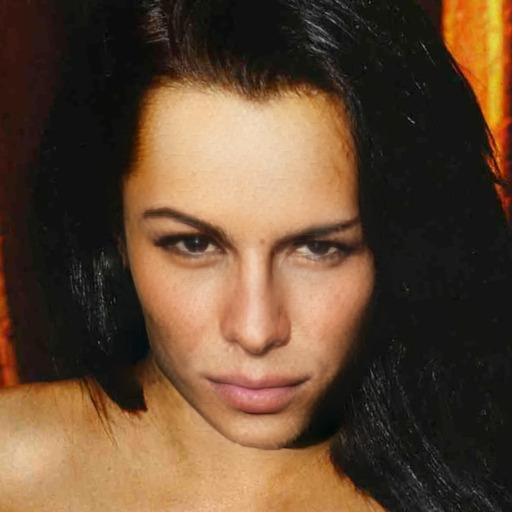} 
    \vspace{2mm}
\end{minipage}
\newcolumntype{x}{>{\centering\arraybackslash\hspace{0pt}}p{18mm}}
\begin{tabular}{lcxxxxxx}
\toprule
\multirow{2}{*}{Method}
    & \multicolumn{3}{c}{w/o align \small{(baseline only)}}
    & \multicolumn{3}{c}{w/ 3D-aware align}
\\
 
    & KID $\downarrow$ & LPIPS$_m$ $\downarrow$ & m$L_2$ $\downarrow$ 
    & KID $\downarrow$ & LPIPS$_m$ $\downarrow$ & m$L_2$ $\downarrow$ 
\\ 
\cmidrule(lr){0-0}
\cmidrule(lr){2-4}
\cmidrule(lr){5-7}

Poisson Blending~\cite{perez2003poisson} \ \ 
    & \underline{0.006}      & 0.4203     & 0.0069
    & \underline{0.005}      & 0.2355     & 0.0051
\\[0.3mm]

Latent Composition~\cite{chai2021latent} \ \ 
    & 0.012      & 0.4735     & 0.0388 
    & 0.012      & 0.4487     & 0.0321
\\[0.3mm]

StyleGAN3 $\mathcal{W}$~\cite{karras2021alias} \ \ 
    & 0.016      & 0.4379     & 0.0353
    & 0.017      & 0.3921     & 0.0307
\\[0.3mm]
StyleGAN3 $\mathcal{W}+$~\cite{karras2021alias} \ \ 
    & 0.025      & 0.4634     & 0.0462   
    & 0.023     & 0.4086     & 0.0391
\\[0.3mm]
StyleMapGAN ($32 \times 32$)~\cite{kim2021exploiting} \ \ 
    & 0.007      & 0.3792     & 0.0118
    & 0.006      & \underline{0.1989}     & 0.0045
\\[0.3mm]   
SDEdit~\cite{meng2021sdedit} \ \ 
    & 0.011      & 0.3857     & 0.0076 
    & 0.008      & 0.3427     & \textbf{0.0003}
\\[0.3mm]
\cmidrule(lr){0-0}
\cmidrule(lr){2-4}
\cmidrule(lr){5-7}
Ours \ \ 
    & 0.013      & \underline{0.2046}     & \underline{0.0050}   
    & 0.013      & 0.2046     & 0.0050   
\\[0.3mm]       
Ours + Poisson Blending \ \ 
    & \textbf{0.002}      & \textbf{0.1883}    & \textbf{0.0007}    
    & \textbf{0.002}      &\textbf{0.1883}     & \underline{0.0007}    
\\[0.3mm]
\bottomrule
\end{tabular}\vspace{0mm}
\caption{{\bf Comparison with baselines in the CelebA-HQ test set}. The first and second rows of the \emph{figure} show the blending results without and with our 3D-aware alignment, respectively. Metric scores on the left side of the \emph{table} show the results without alignment. We apply our 3D-aware alignment to the baselines on the right side of the table. Lower scores denote better performance in all metrics. The best and second-best scores are bold and underlined. Our method outperforms baselines in all metrics. 
LC and PB stand for Latent Composition~\cite{chai2021latent} and Poisson Blending~\cite{perez2003poisson}, respectively. Note that our method always operates 3D-aware alignment, as it is an integral part of our algorithm.}
\label{tbl:baseline_edit_celeba}
\end{table*}
}

\newcommand{\tabBaselineEditAFHQvtwo}{
\begin{table*}[t!]
\centering

\begin{minipage}{\linewidth}
    \centering
    \hspace{3mm}
    \makebox[\himg\linewidth][c]{~~\small{Original}}
    \makebox[\himg\linewidth][c]{~~\small{Reference}}
    \makebox[\himg\linewidth][c]{~\small{Poisson}}\hfill
    \makebox[\himg\linewidth][c]{\small{StyleGAN3 $\mathcal{W}$}}\hfill
    \makebox[\himg\linewidth][c]{\small{StyleGAN3 $\mathcal{W}+$}}\hfill
    \makebox[\himg\linewidth][c]{\small{StyleMap$^{8\times8}$}}\hfill 
    \makebox[\himg\linewidth][c]{\small{StyleMap$^{16\times16}$}}\hfill 
    \makebox[\himg\linewidth][c]{\small{Ours + PB~~}}\hfill 
    
    \hspace{-3.5mm}
    \rotatebox{90}{\makebox[20mm][c]{\small{~~~~w/o align}}}\vspace{-2.4mm}\hspace{0.5mm}
    \includegraphics[width=\himg\linewidth]{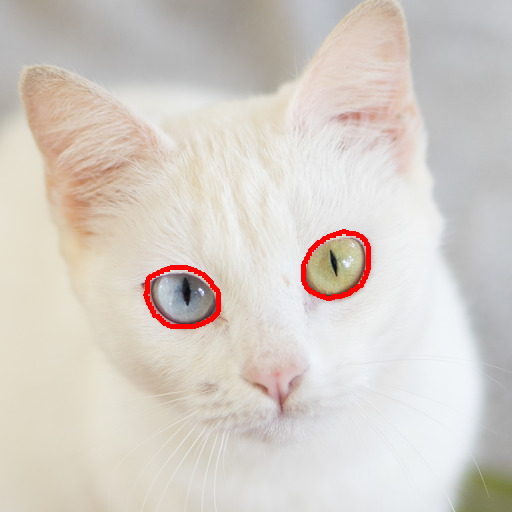}
    \includegraphics[width=\himg\linewidth]{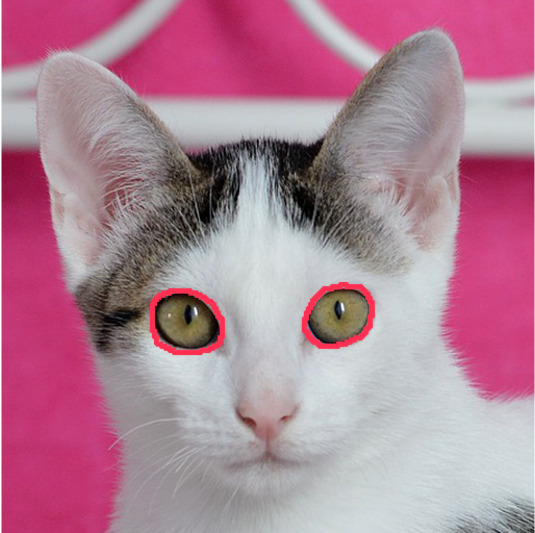}
    \includegraphics[width=\himg\linewidth]{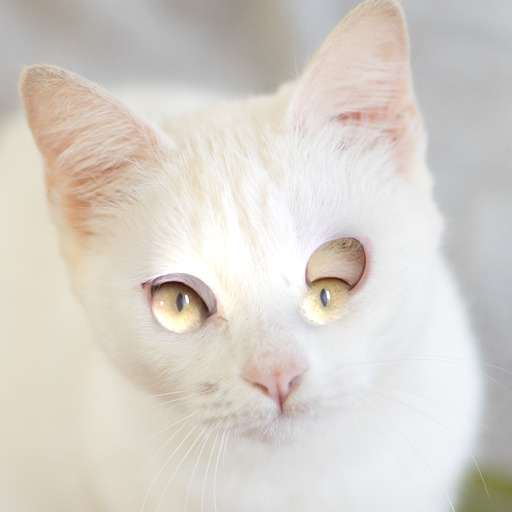}
    \includegraphics[width=\himg\linewidth]{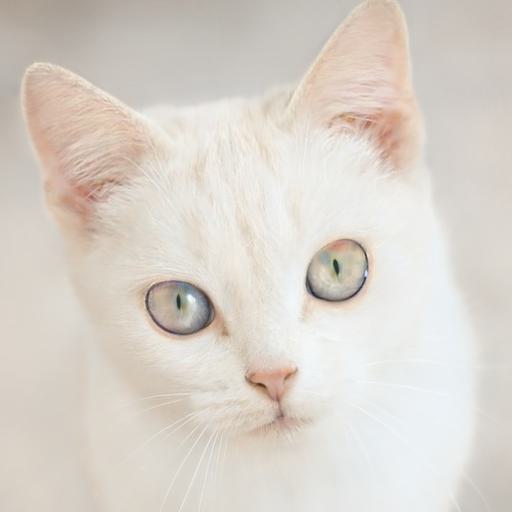}
    \includegraphics[width=\himg\linewidth]{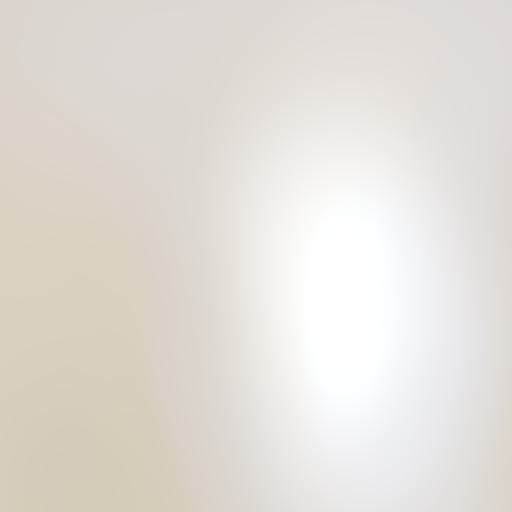}
    \includegraphics[width=\himg\linewidth]{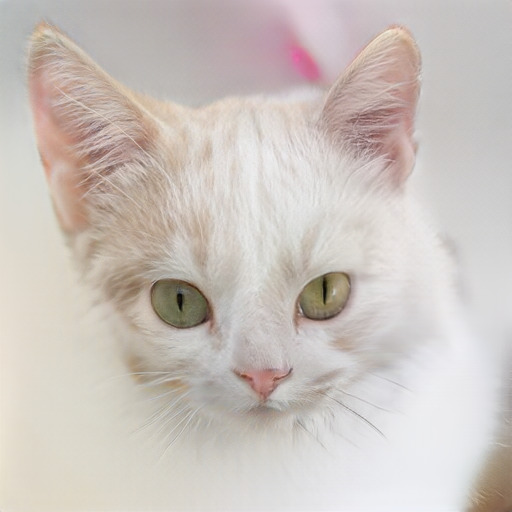}
    \includegraphics[width=\himg\linewidth]{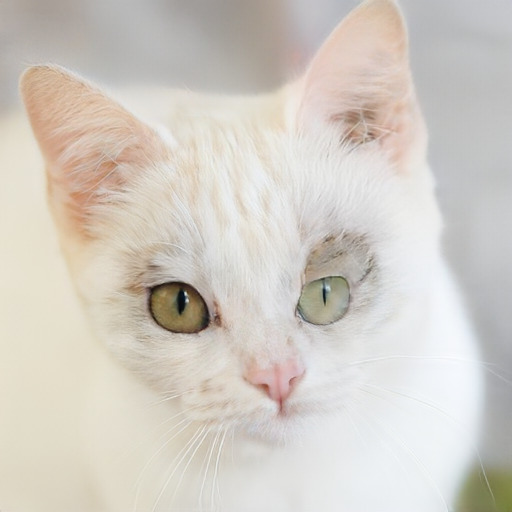}
      \begin{tikzpicture}
        \node[draw=black, text centered, minimum width=2.03cm, minimum height=2.03cm, inner sep=0pt, line width=0.2pt] {\texttt{N/A}};
      \end{tikzpicture}    
    \vspace{2mm}
    \hspace{-2.5mm}
    
    \rotatebox{90}{\makebox[20mm][c]{\small{w/ align}}}
    \vspace{0.0mm}\hspace{-0.5mm}
    \includegraphics[width=\himg\linewidth]{figures/table2/tab2_ori.jpg}
    \includegraphics[width=\himg\linewidth]{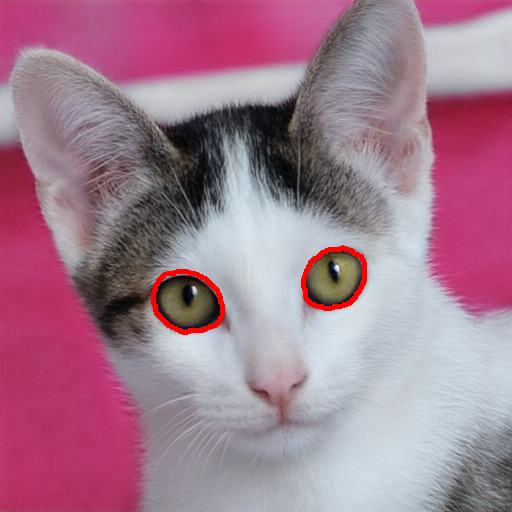}
    \includegraphics[width=\himg\linewidth]{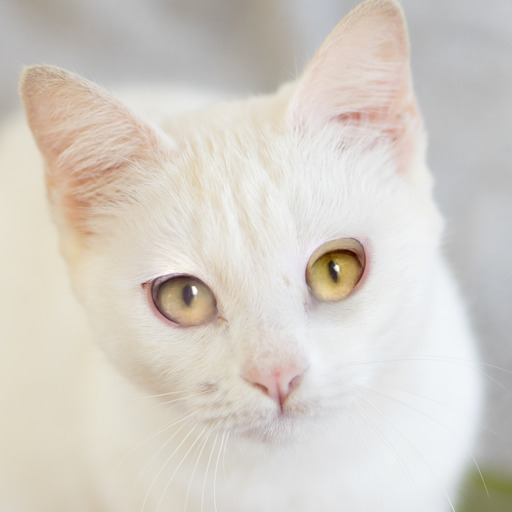}
    \includegraphics[width=\himg\linewidth]{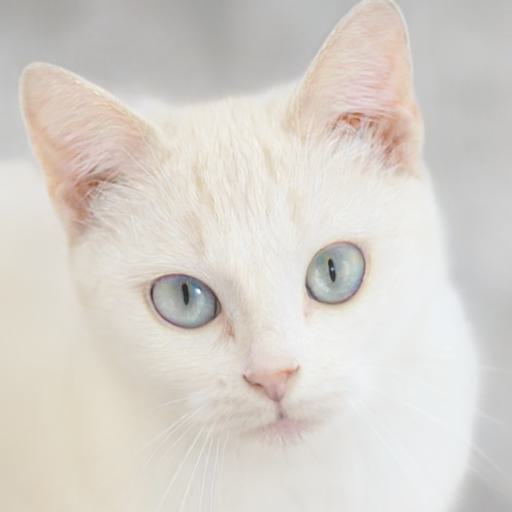}
    \includegraphics[width=\himg\linewidth]{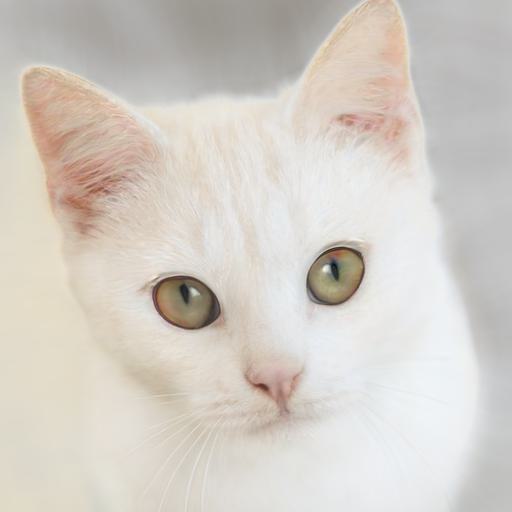}
    \includegraphics[width=\himg\linewidth]{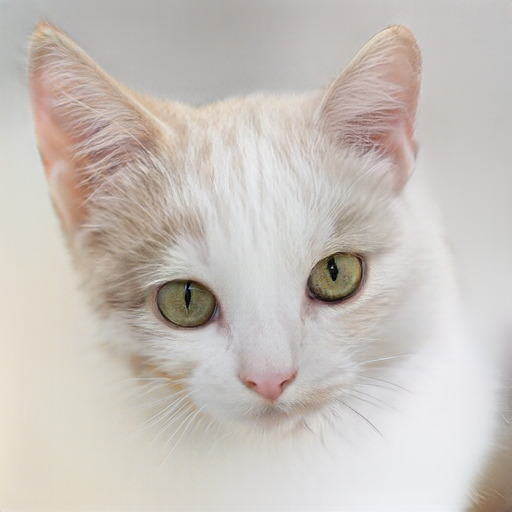}
    \includegraphics[width=\himg\linewidth]{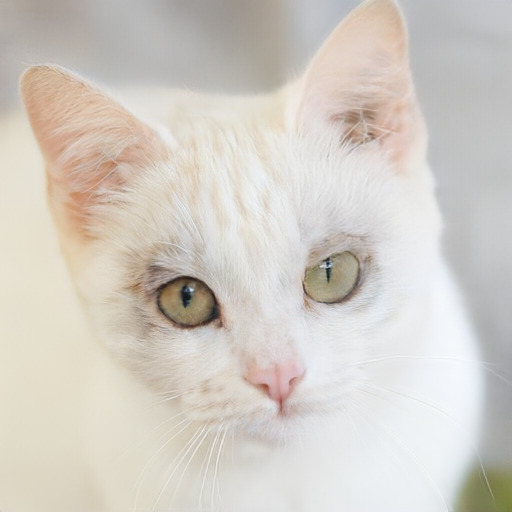}
    \includegraphics[width=\himg\linewidth]{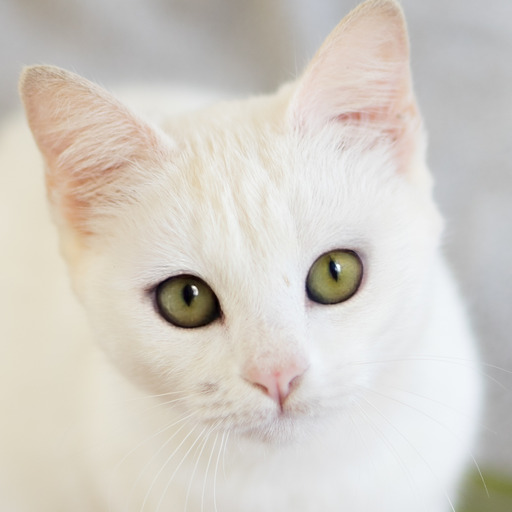} 
    \vspace{2mm}
\end{minipage}
\newcolumntype{x}{>{\centering\arraybackslash\hspace{0pt}}p{18mm}}
\begin{tabular}{lcxxxxxx}
\toprule
\multirow{2}{*}{Method}
    & \multicolumn{3}{c}{w/o align \small{(baseline only)}}
    & \multicolumn{3}{c}{w/ 3D-aware align}
\\
 
    & KID $\downarrow$ & LPIPS$_m$ $\downarrow$ & m$L_2$ $\downarrow$ 
    & KID $\downarrow$ & LPIPS$_m$ $\downarrow$ & m$L_2$ $\downarrow$ 
\\ 
\cmidrule(lr){0-0}
\cmidrule(lr){2-4}
\cmidrule(lr){5-7}

Poisson Blending~\cite{perez2003poisson} \ \ 
    & \textbf{0.002}      & 0.4956     & \underline{0.0024}
    & \textbf{0.002}      & \underline{0.2656}     & \textbf{0.0004}
\\[0.3mm]
StyleGAN3 $\mathcal{W}$~\cite{karras2021alias} \ \ 
    & 0.006      & 0.4588     & 0.0316 
    & 0.006      & 0.3802     & 0.0268
\\[0.3mm]
StyleGAN3 $\mathcal{W}+$~\cite{karras2021alias} \ \ 
    & 0.014      & 0.4941     & 0.0298
    & 0.013      & 0.3903     & 0.0236
\\[0.3mm]
StyleMapGAN ($8\times8$)~\cite{kim2021exploiting} \ \ 
    & 0.013      & 0.4840     & 0.0574  
    & 0.013      & 0.3221     & 0.0526
\\[0.3mm]
StyleMapGAN ($16\times16$)~\cite{kim2021exploiting} \ \ 
    & 0.006      & 0.4746     & 0.0225
    & 0.004      & 0.2707     & 0.0160
\\[0.3mm]   
\cmidrule(lr){0-0}
\cmidrule(lr){2-4}
\cmidrule(lr){5-7}
Ours \ \ 
    & 0.005     & \underline{0.2739} & 0.0073    
    & 0.005      & 0.2739     & 0.0073    
\\[0.3mm]       
Ours + Poisson Blending \ \ 
    & \textbf{0.002}      & \textbf{0.2229}    & \textbf{0.0013}    
    & \textbf{0.002 }     &\textbf{ 0.2229}     & \underline{0.0013}    
\\[0.3mm]
\bottomrule
\end{tabular}
\caption{{\bf Comparison with baselines in the AFHQv2-Cat test set}. Formats of the figure and table are the same as \reftbl{baseline_edit_celeba}. 
}
\label{tbl:baseline_edit_afhq}
\end{table*}
}

\newcommand{\UserCelebA}{
\begin{table}[t!]
\footnotesize
\centering
\begin{tabular}{@{\hspace{1mm}}lcccc@{\hspace{1mm}}}
\toprule
\multirow{2}{*}{\textbf{Method}}
    & \multicolumn{2}{c}{\textbf{Ours}}
    & \multicolumn{2}{c}{\textbf{Ours + Poisson Blending}}
\\
    & w/o  & w/ align
    & w/o  & w/ align 
\\ 
\cmidrule(lr){1-1}
\cmidrule(lr){2-3}
\cmidrule(lr){4-5}

Poisson~\cite{perez2003poisson} \ \ 
    & 79.9\%  &  59.9\%  
    & 80.9\%~(\textcolor{green}{+1.0})   & 67.5\%~(\textcolor{green}{+7.6})
\\[0.3mm]
StyleMap~\cite{kim2021exploiting} \ \ 
    & 72.3\%    & 62.0\%   
    & 75.4\%~(\textcolor{green}{+3.1})  & 66.3\%~(\textcolor{green}{+4.3}) 
\\[0.3mm]
SDEdit~\cite{meng2021sdedit} \ \ 
    & 61.0\%   & 55.7\%  
    & 61.1\%~(\textcolor{green}{+0.1})   & 50.2\%~(\textcolor{red}{-5.5})
    
\\[0.3mm]   
\bottomrule
\end{tabular}\vspace{0mm}
\caption{
\textbf{User study in CelebA-HQ} regarding the photorealism of the blended image. 
The percentage denotes how often MTurk workers prefer our method to each baseline in pairwise comparison. Values larger than 50\% mean ours outperforms the baseline. 
Our method, both with and without Poisson blending, outperforms all baselines even if we improve the baselines using our 3D-aware alignment. 
Incorporating Poisson blending further enhances the realism score of our method as shown in \textcolor{green}{green} numbers.
}
\ifdefined\ARXIV
\else
\vspace{10mm}
\fi

\label{tbl:userstudyCelebA}
\end{table}
}

\newcommand{\UserAFHQ}{
\begin{table}[t!]
\footnotesize
\centering
\begin{tabular}{@{\hspace{1mm}}lcccc@{\hspace{1mm}}}
\toprule
\multirow{2}{*}{\textbf{Method}}
    & \multicolumn{2}{c}{\textbf{Ours}}
    & \multicolumn{2}{c}{\textbf{Ours + Poisson Blending}}
\\
    & w/o   & w/ align
    & w/o  & w/ align 
\\ 
\cmidrule(lr){0-0}
\cmidrule(lr){2-3}
\cmidrule(lr){4-5}

Poisson~\cite{perez2003poisson} \ \ 
    & 91.2\%  &  76.2\%  
    & 87.6\%~(\textcolor{red}{-3.6})   & 82.8\%~(\textcolor{green}{+6.6})
\\[0.3mm]
StyleMap~\cite{kim2021exploiting} \ \ 
    & 91.0\%    & 82.3\%   
    & 91.6\%~(\textcolor{green}{+0.6})  & 83.6\%~(\textcolor{green}{+1.3}) 
\\[0.3mm]
\bottomrule
\end{tabular}\vspace{0mm}
\caption{\textbf{User study in AFHQv2-Cat} regarding the photorealism of the blended image. All details are the same as in~\reftbl{userstudyCelebA}. Our approach surpasses all baselines, and the incorporation of Poisson blending further improves the realism score.
}
\label{tbl:userstudyAFHQ}
\end{table}
}

\newcommand{\ablationWarping}{
\begin{figure}[]
\centering
\includegraphics[width=1.0\linewidth]{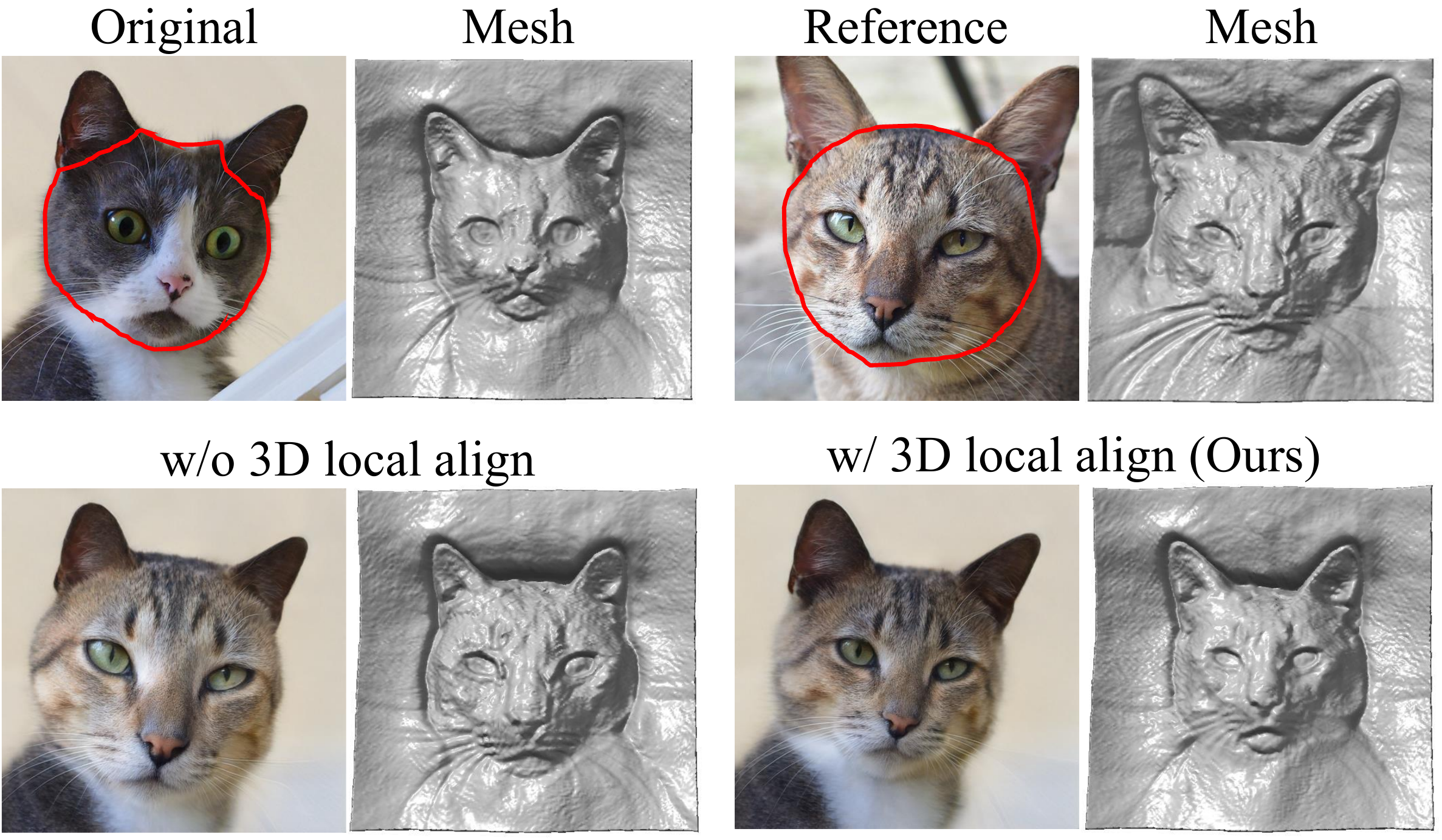} 
\caption{The effect of local alignment. Generative NeRF can align global object poses, but cannot handle differences in local parts. For example, in AFHQ-Cat, the proportions of a cat's face and ears vary. (Bottom left) It cannot be handled with pose estimation only, so local parts can be blended in the wrong places. (Bottom right) Our 3D local alignment method alleviates this issue and produces more natural results.} 
\label{fig:ablationWarping}
\end{figure}
}

\newcommand{\suppfiglocalalign}{
\begin{figure}[]
\centering
\includegraphics[width=1.0\linewidth]{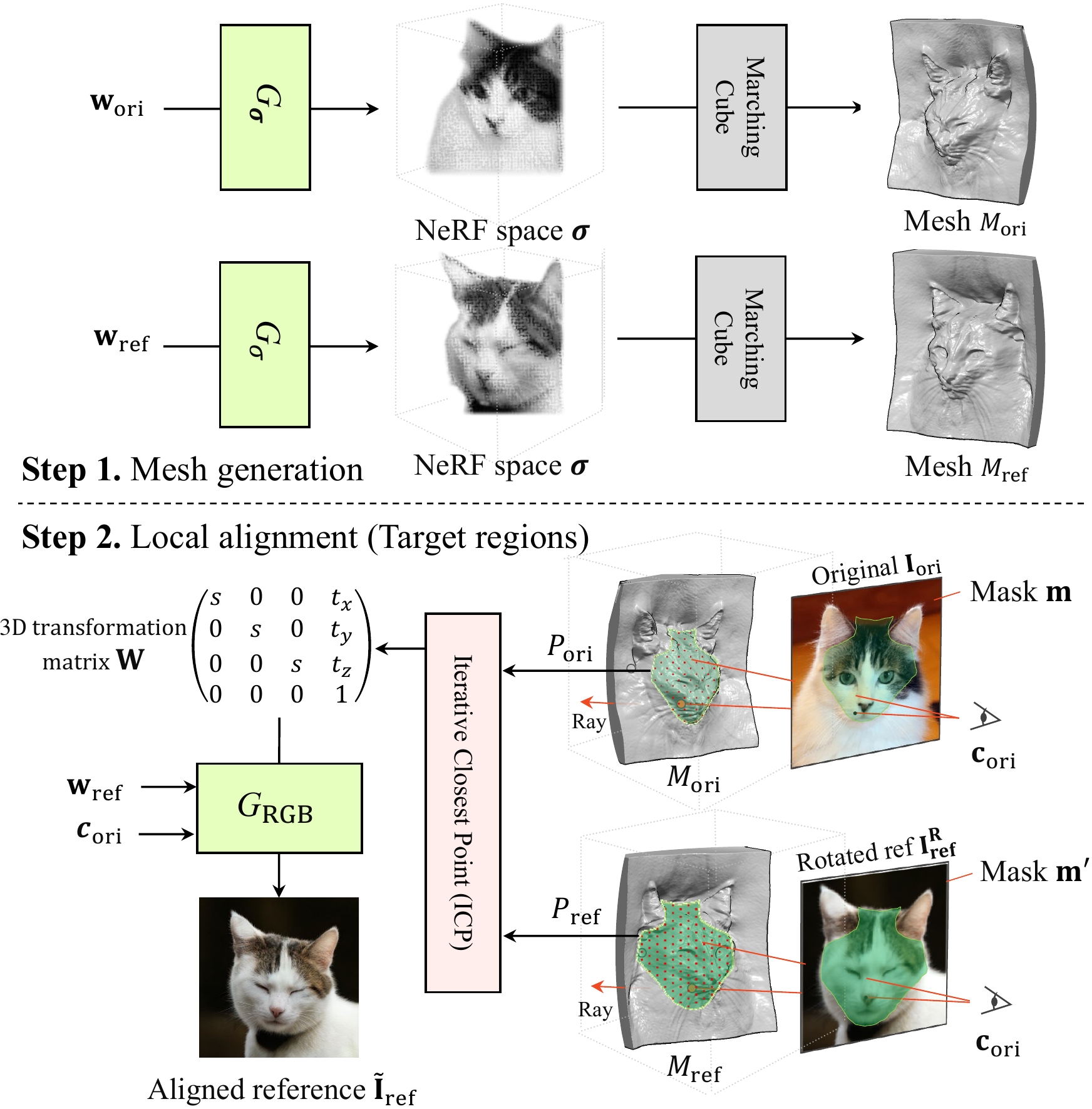} 
\caption{{\bf Local alignment}: In {\bf Step 1}, we first generate 3D meshes using the density field generator $G_{\sigma}$ and the Marching Cube~\cite{lorensen1987marching} algorithm. In {\bf Step 2}, we 
calculate the intersected 3D points between the 2D mask $\m$ and the corresponding mesh $\Mesh{}$. Then, we use Iterative Closest Point (ICP) algorithm~\cite{besl1992method,chen1992object} to estimate the 3D transformation matrix $\mathbf{W} \in \R^{4\times4}$ to align the 3D point clouds ($\Pt{ori}$ and $\Pt{ref}$) in terms of the scale and the translation. Finally, we locally align the reference image $\Ialigned{ref}$ by $\G{RGB}(\w{ref}, \cam{ori}; \mathbf{W})$.
} 

\label{fig:suppfiglocalalign}
\end{figure}
}

\newcommand{\suppfigchoiceBlending}{
\begin{figure}[]
\centering
\includegraphics[width=1.0\linewidth]{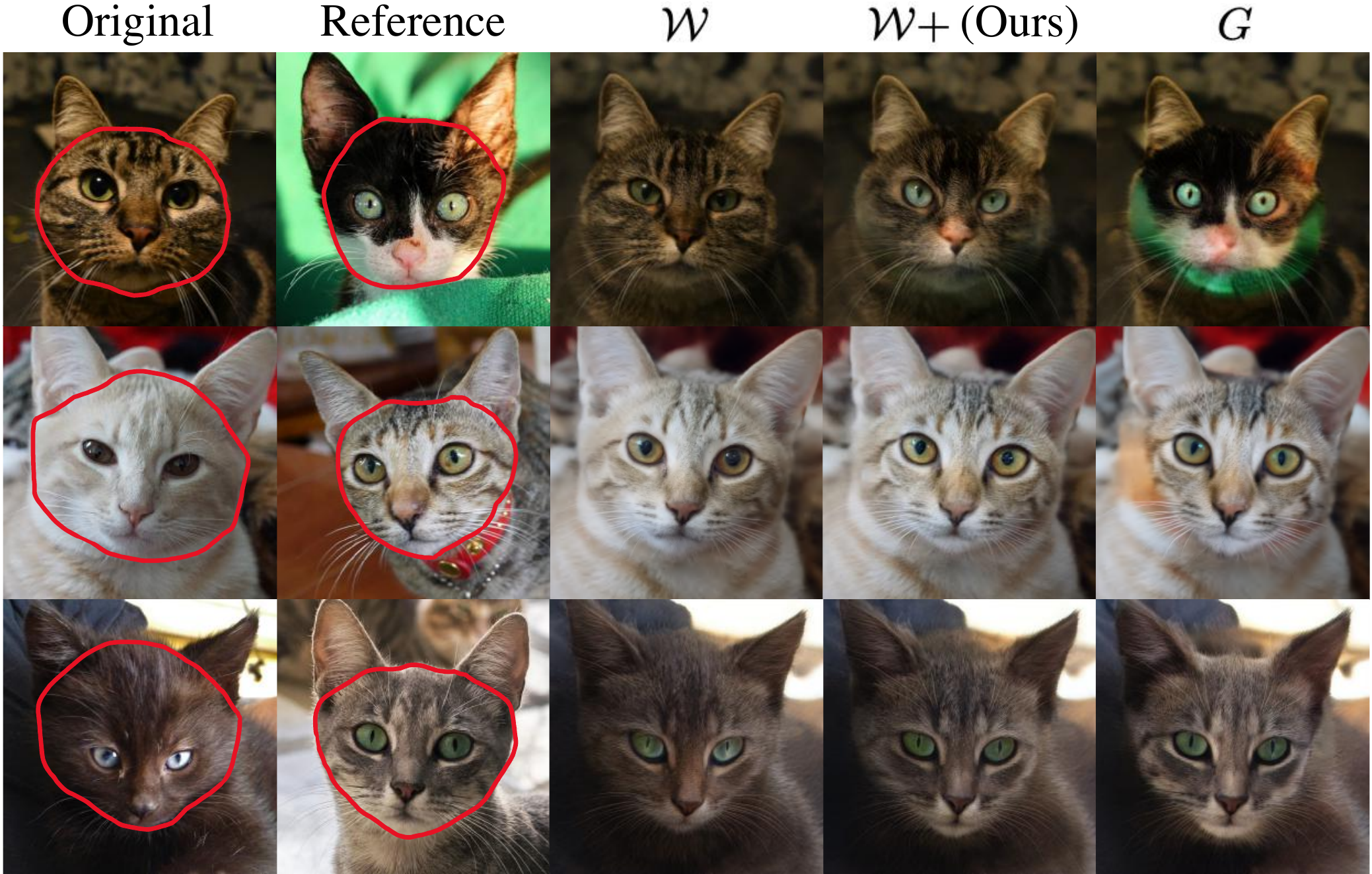}\\
\caption{Blending comparison among the optimization spaces in \refapp{blendingDetails}. The first and second columns denote original and reference images, respectively.
The blending results of $\mathcal{W}$ space (3rd column) show realistic but less faithful to the reference images; see the details of eyes. The blending results of $\G{}$ space (rightmost column) show faithful but less realistic results. $\mathcal{W}+$ space (4th column) shows favorable results in both realism and faithfulness.
} 
\vspace{-3mm}
\label{fig:choiceBlending}
\end{figure}
}

\newcommand{\suppfigablationWarping}{
\begin{figure*}[]
\centering
\includegraphics[width=0.8\linewidth]{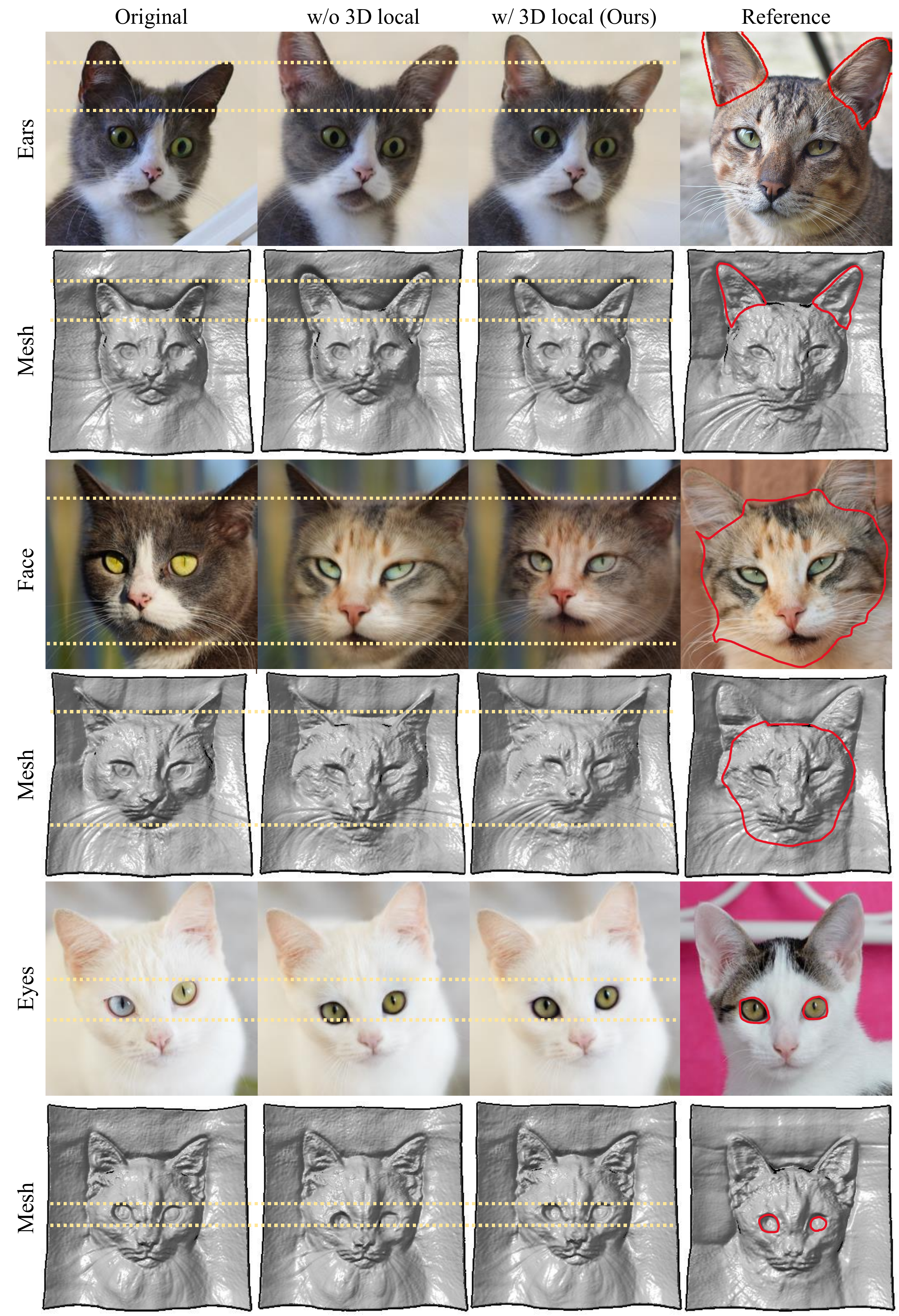}\\
\caption{Ablation study of our 3D local alignment method in \refapp{local_align}. The leftmost and rightmost columns denote the original and reference images, respectively. Images in even rows are meshes corresponding to odd rows. Red lines denote the target blending mask, and yellow dotted lines are guidelines to be aware of alignment easily. Our 3D local alignment method (3rd column) shows more realistic blending results than those without local alignment (2nd column). 
} 
\label{fig:suppablationWarping}
\end{figure*}
}

\newcommand{\suppfigStyleSDFblend}{
\begin{figure}[]
\centering
\includegraphics[width=1.0\linewidth]{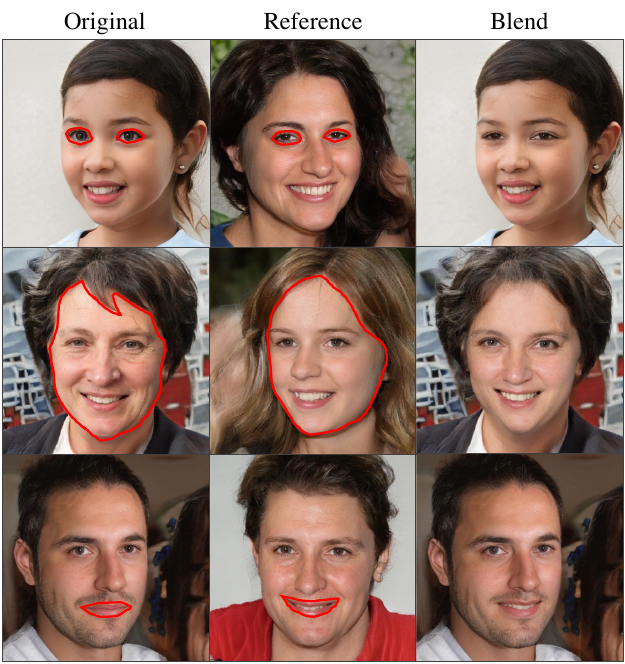}\\
\caption{Our 3D-aware blending results of StyleSDF~\cite{or2022stylesdf} in \refapp{stylesdf}. The third column is blending results in $1024^2$ resolution. Our method can be applied to the SDF-based 3D-aware generator beyond NeRFs.
} 
\label{fig:suppfigStyleSDFblend}
\end{figure}
}

\newcommand{\suppfigStyleSDFdensity}{
\begin{figure}[]
\centering
\includegraphics[width=1.0\linewidth]{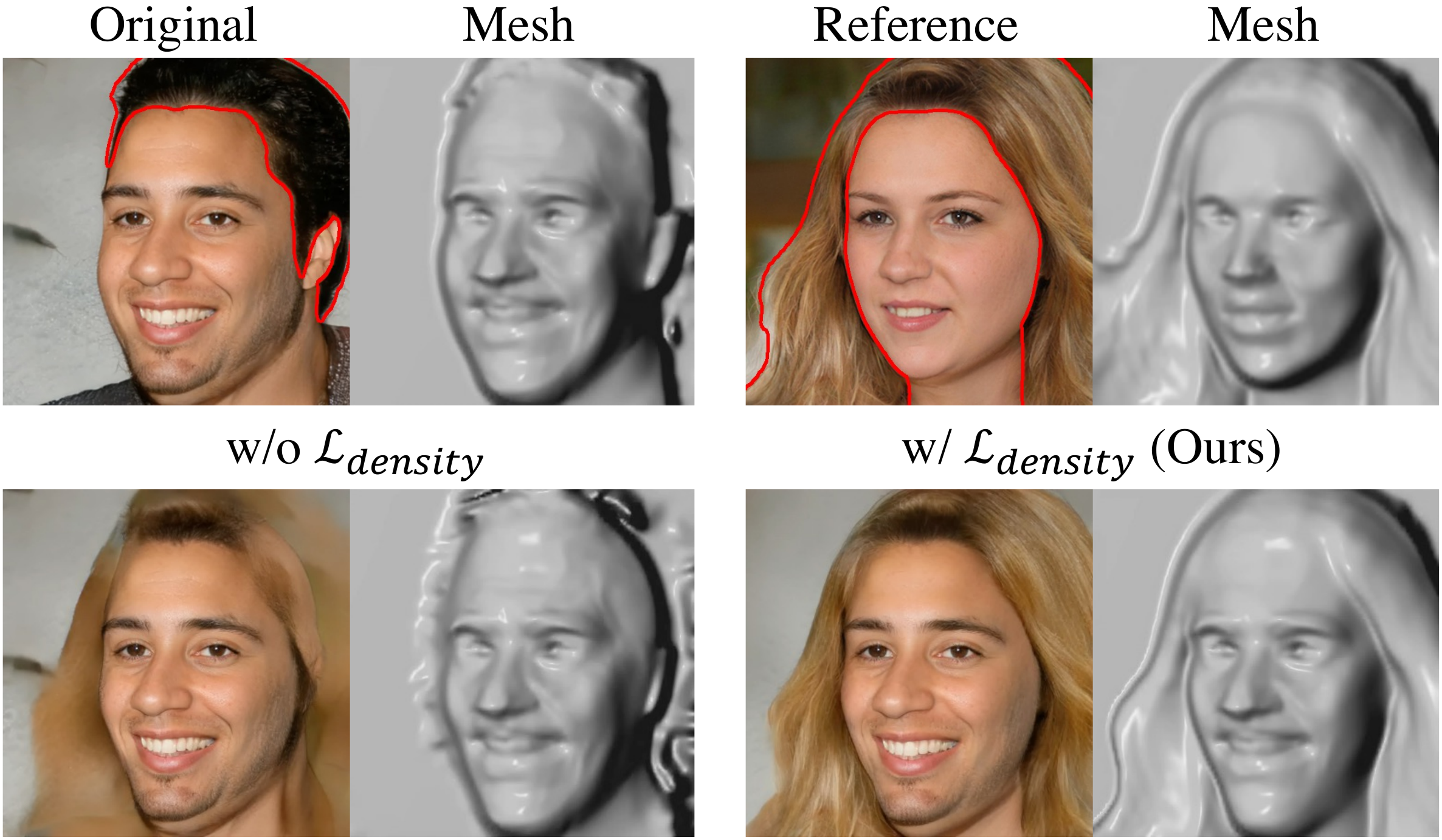}\\
\caption{Ablation study of density-blending loss $\L{density}$ in StyleSDF (\refapp{stylesdf}). This experiment uses SDF instead of volume density $\sigma$.
Without $\L{density}$, the blending result shows a blurry image, and the corresponding mesh can not reflect the geometry of the hair of the reference image.
3D signals such as density and SDF are key components in the 3D-aware blending method.
} 
\vspace{-3mm}
\label{fig:suppfigStyleSDFdensity}
\end{figure}
}

\newcommand{\suppfigStyleSDFimage}{
\begin{figure}[]
\centering
\includegraphics[width=1.0\linewidth]{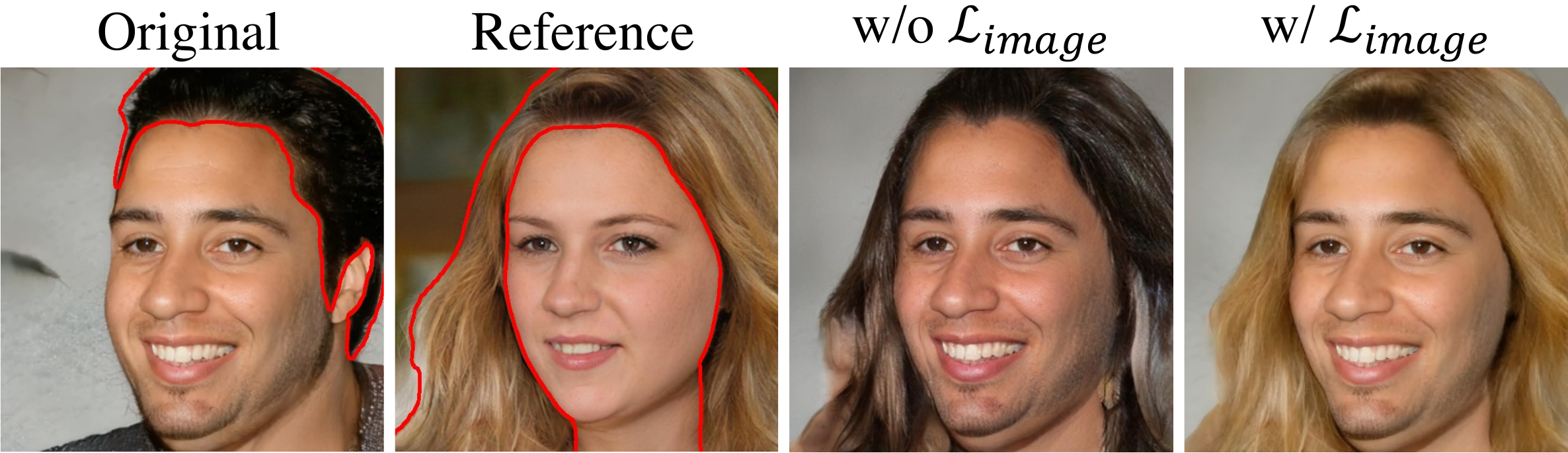}\\
\caption{Ablation study of image-blending loss $\L{image}$ in StyleSDF (\refapp{stylesdf}). 
We can control the degree to which the color of the reference object is reflected in the blended image, by adjusting the weight of the image-blending loss with respect to the reference image.
} 
\label{fig:suppfigStyleSDFimage}
\end{figure}
}

\newcommand{\suppfigAMTintro}{
\begin{figure}[t]
\centering
\includegraphics[width=1.0\linewidth]{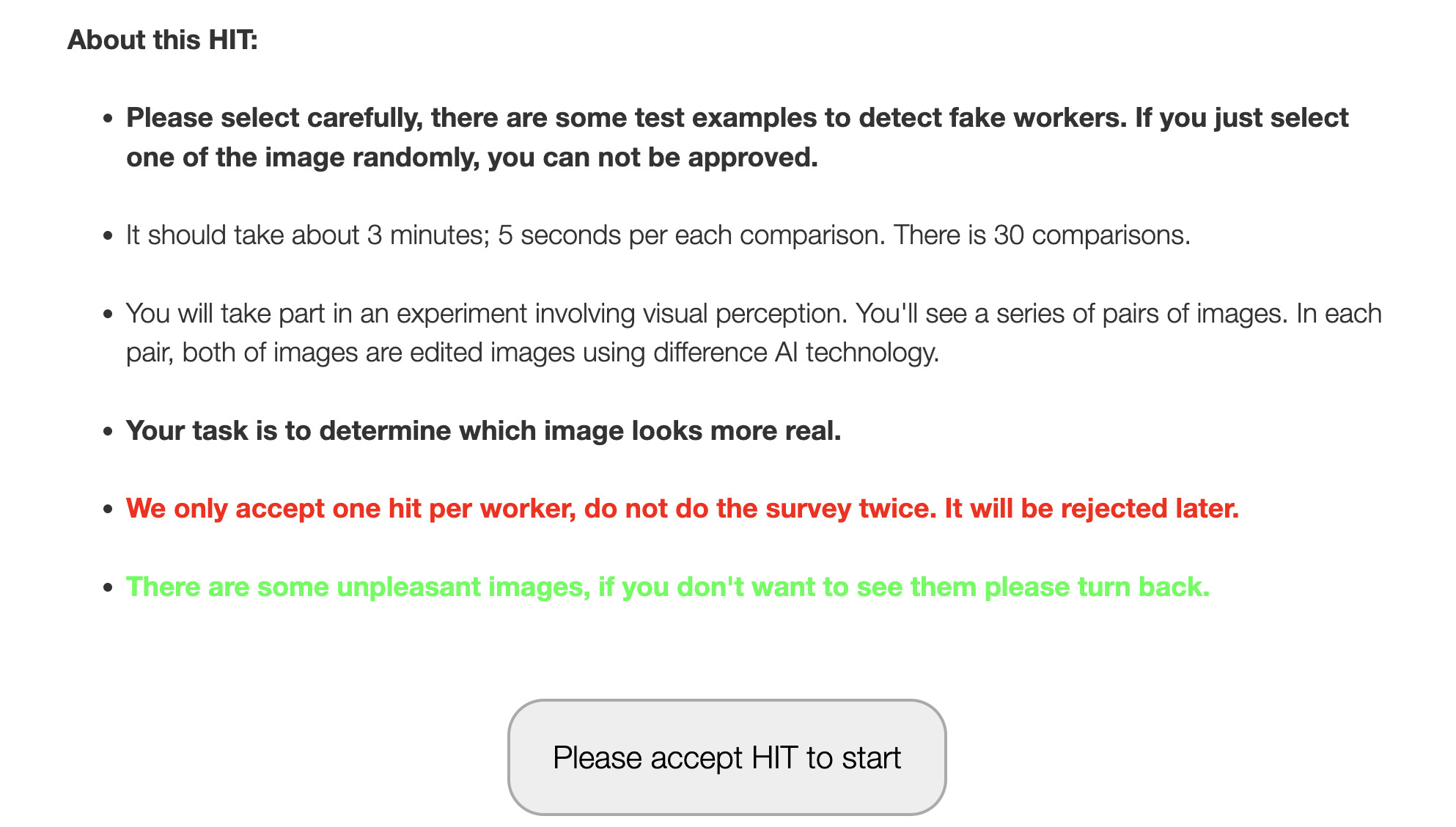}\\
\caption{An introduction page of user studies in Amazon Mechanical Turk (MTurk).
} 
\label{fig:suppfigAMTintro}
\end{figure}
}

\newcommand{\suppfigAMTbaseline}{
\begin{figure}[h]
\centering
\includegraphics[width=0.85\linewidth]{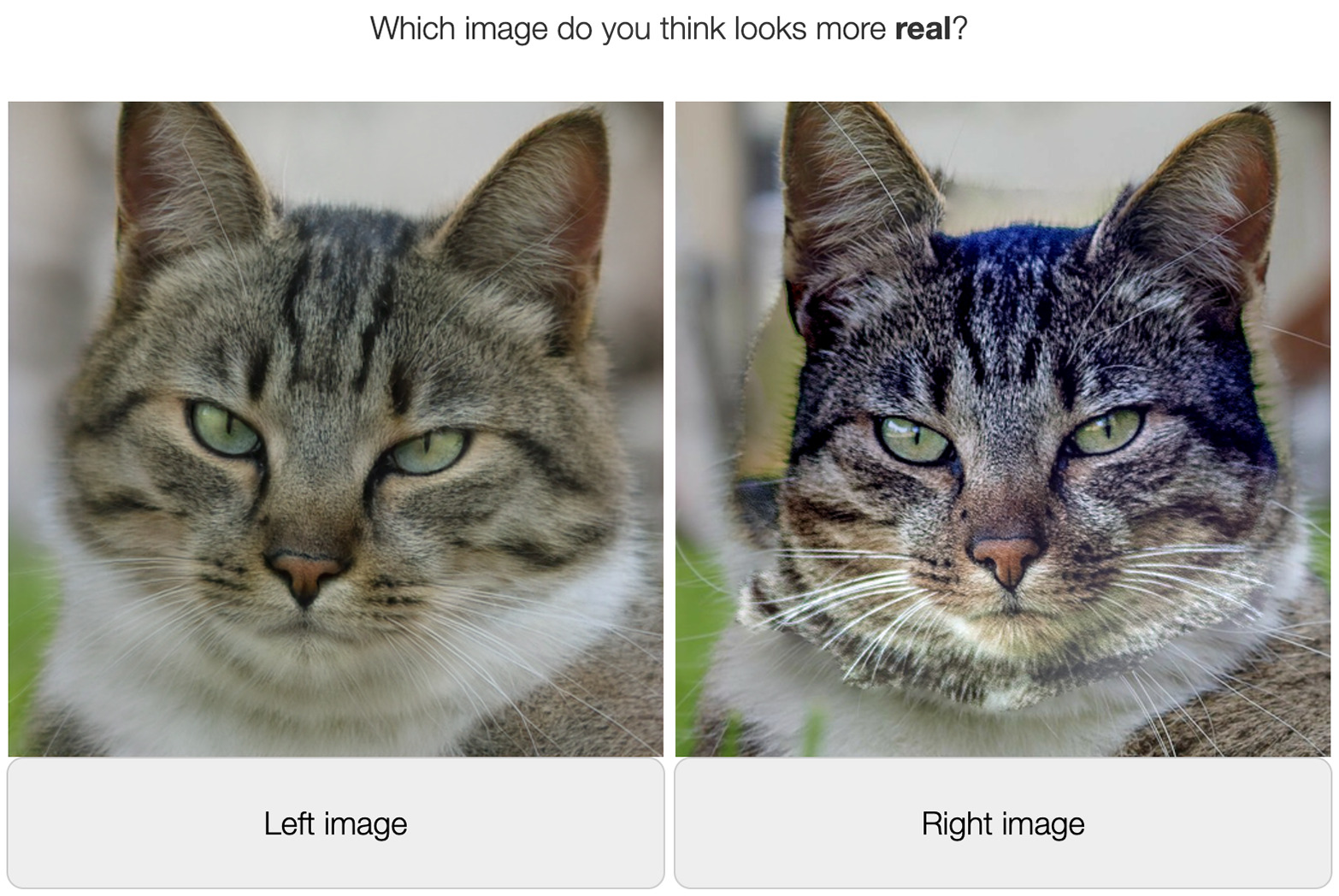}\\
\caption{A comparison page between ours and baselines in MTurk.} 
\vspace{-2mm}
\label{fig:suppfigAMTbaseline}
\end{figure}

}

\newcommand{\suppfigAMTwarping}{
\begin{figure}[h]
\centering
\includegraphics[width=0.85\linewidth]{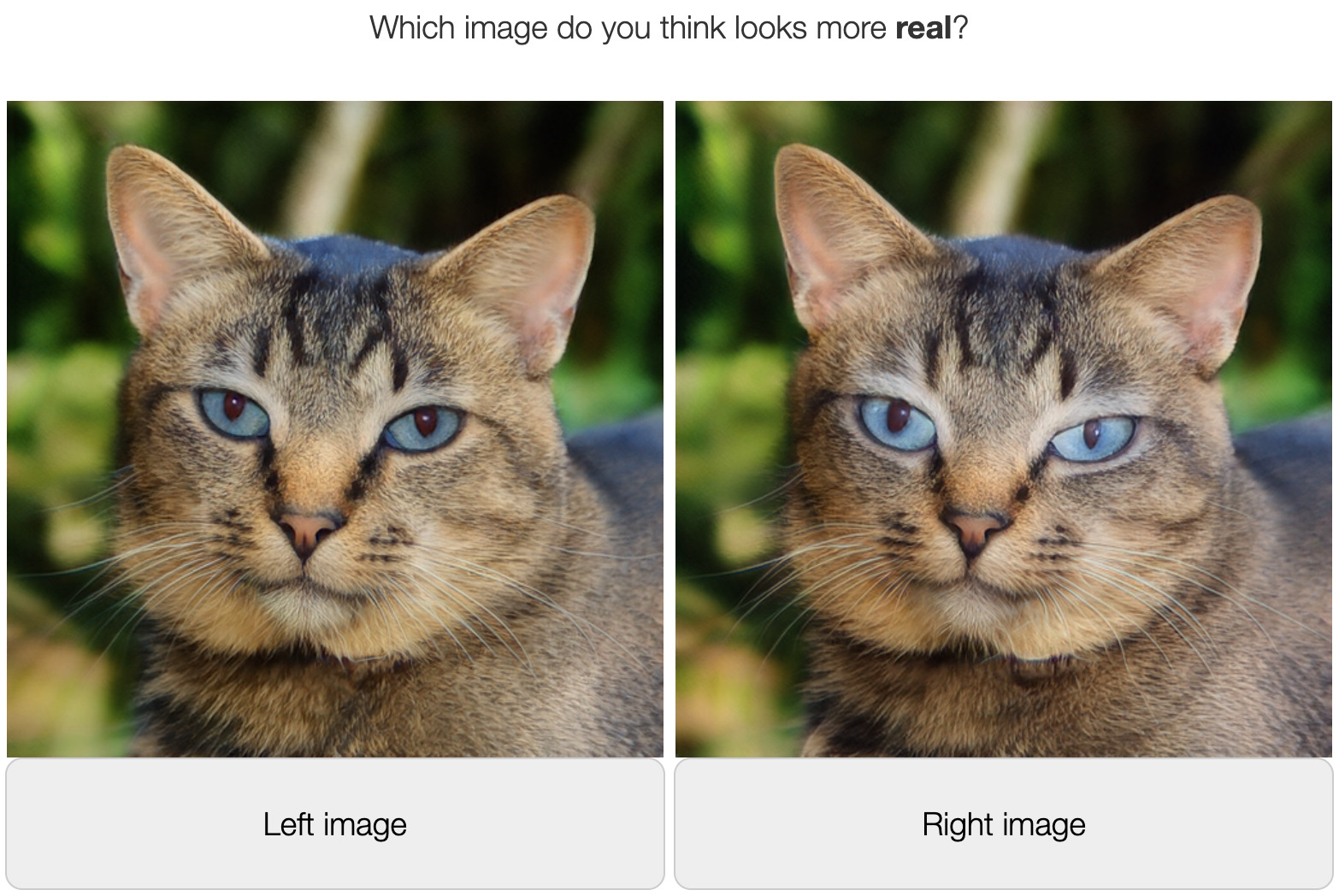}\\
\caption{A comparison page of ablation study of our local alignment in MTurk.} 
\label{fig:suppfigAMTwarping}
\end{figure}
}

\newcommand{\suppfigLimitICP}{
\begin{figure}[]
\centering
\includegraphics[width=1.0\linewidth]{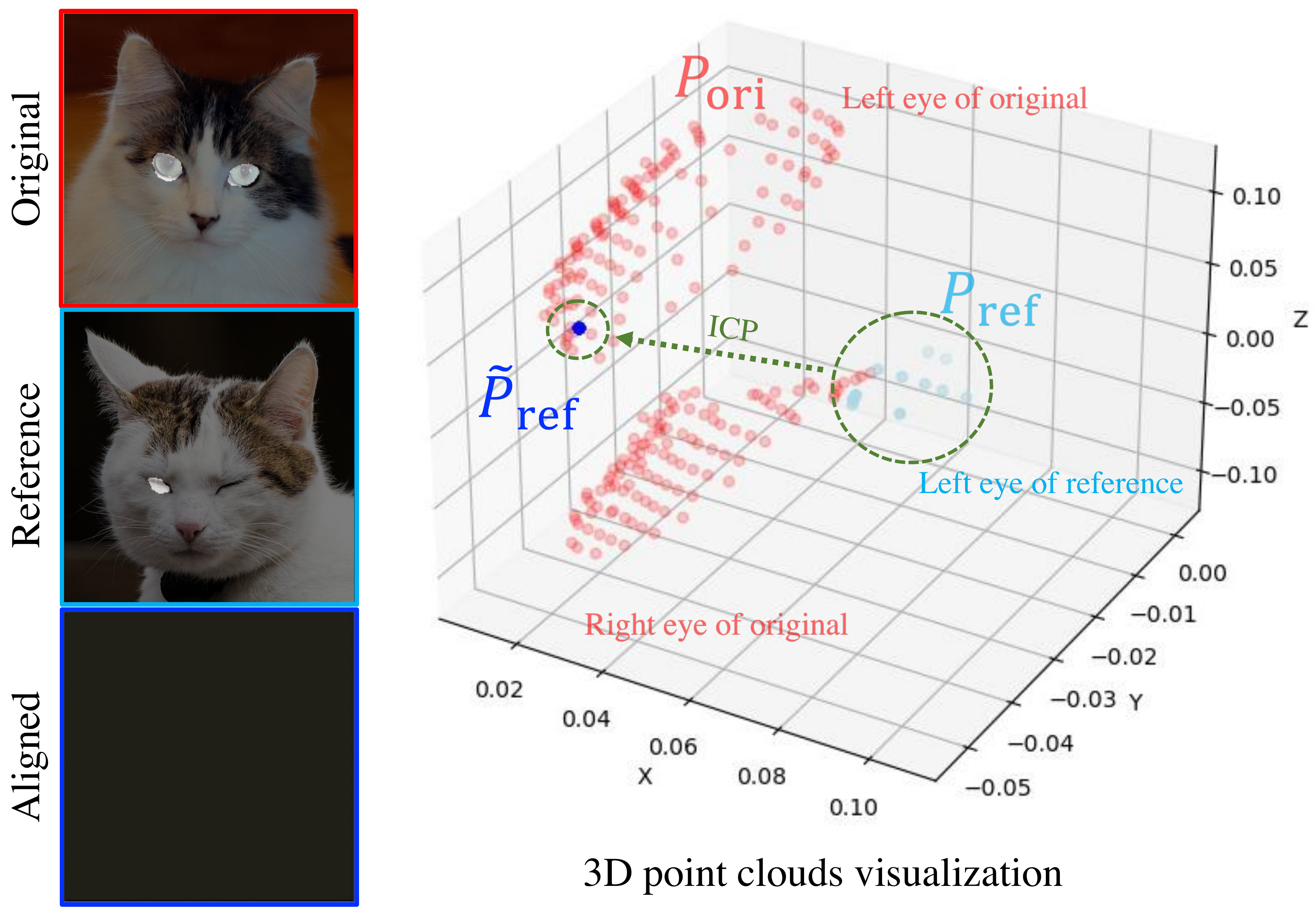}\\
\caption{A failure case of the Iterative Closest Point (ICP) algorithm in \refapp{limit}. 
If a user gives inappropriate masks or the target region of the mesh has indistinctive geometry, ICP may fail and generates a degraded aligned reference image. 
Masks are given as both eyes for the original image and the left eye for the reference image. 
The right figure shows the point cloud visualization. \textcolor{red}{Red} dots stand for the point cloud $\Pt{ori}$ of both eyes of the original object. \textcolor{cyan}{Light blue} dots stand for the point cloud $\Pt{ref}$ of the left eye of the reference object. \textcolor{blue}{Blue} dots denote the transformed point cloud $\tilde{P}_{\text{ref}}$ of the reference after applying ICP. 
} 
\vspace{-3mm}
\label{fig:suppfigLimitICP}
\end{figure}
}

\newcommand{\suppfigCompareCelebA}{
\begin{figure*}[t]
\centering
\includegraphics[width=1.0\linewidth]{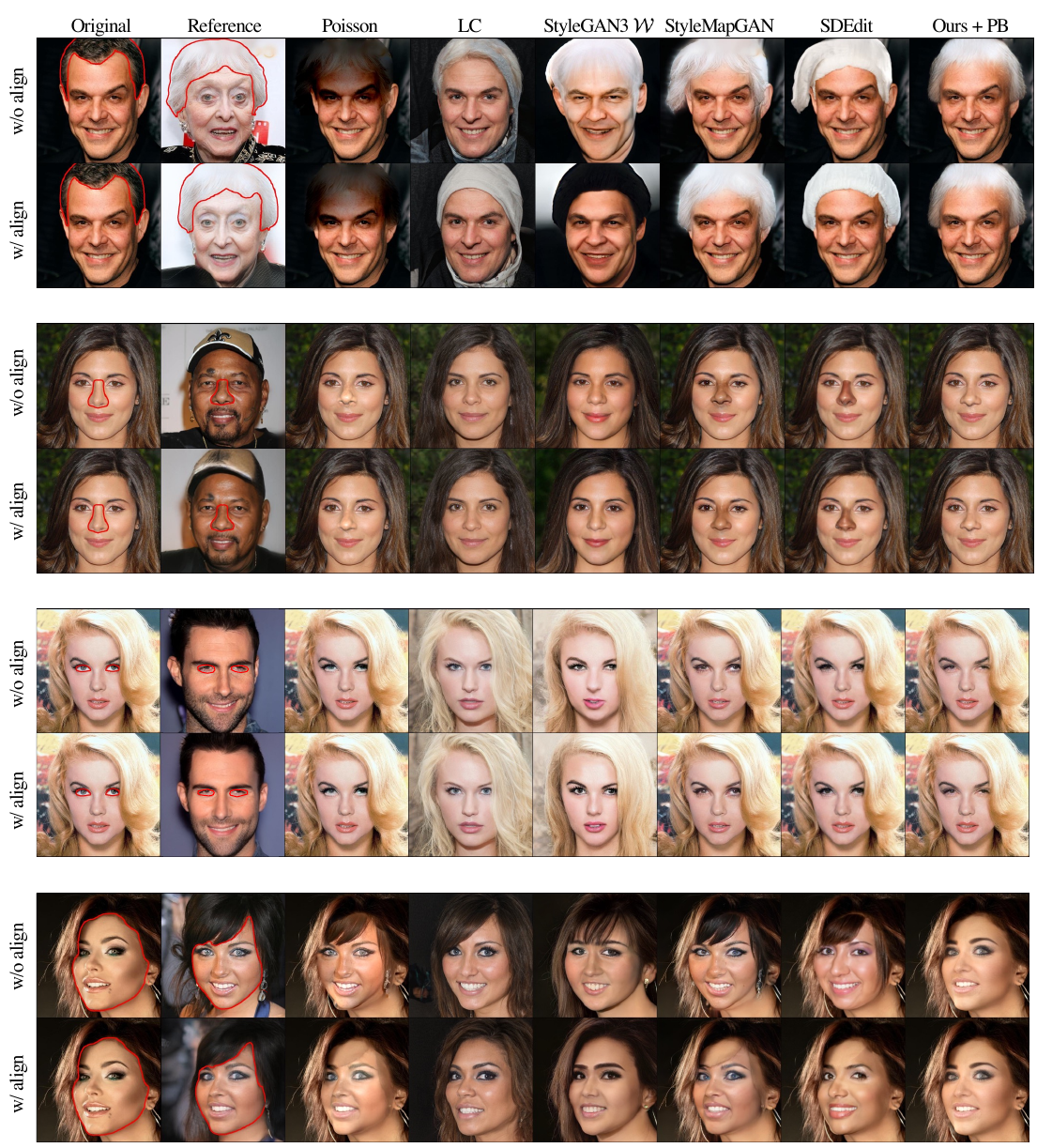}\\
\caption{Blending comparison with baselines in the CelebA-HQ test set. It shows supplementary comparison results of 
\ifdefined\ARXIV
   \reftbl{baseline_edit_celeba}.
\else
   Table 1 in the paper.
\fi
 LC and PB stand for Latent Composition~\cite{chai2021latent} and Poisson Blending~\cite{perez2003poisson}, respectively.
} 
\label{fig:suppfigCompareCelebA}
\end{figure*}

}

\newcommand{\suppfigCompareAFHQ}{
\begin{figure*}[t]
\centering
\includegraphics[width=1.0\linewidth]{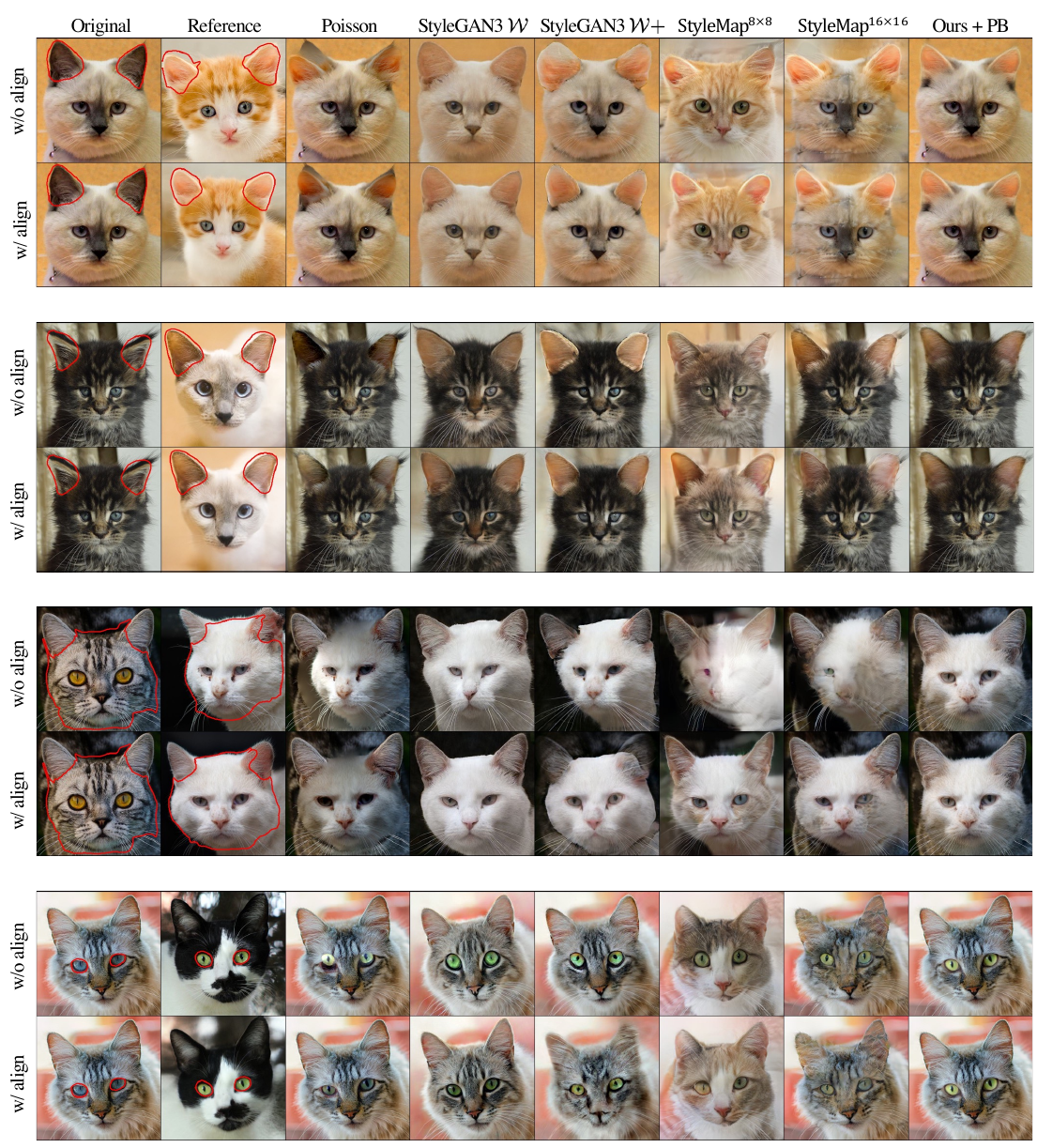}\\
\caption{Blending comparison with baselines in the AFHQv2-Cat test set. It shows supplementary comparison results of 
\ifdefined\ARXIV
   \reftbl{baseline_edit_afhq}.
\else
   Table 2 in the paper.
\fi
} 
\label{fig:suppfigCompareAFHQ}
\end{figure*}
}

\newcommand{\suppfigMultiviewCelebA}{
\begin{figure}[t]
\centering
\includegraphics[width=1.0\linewidth]{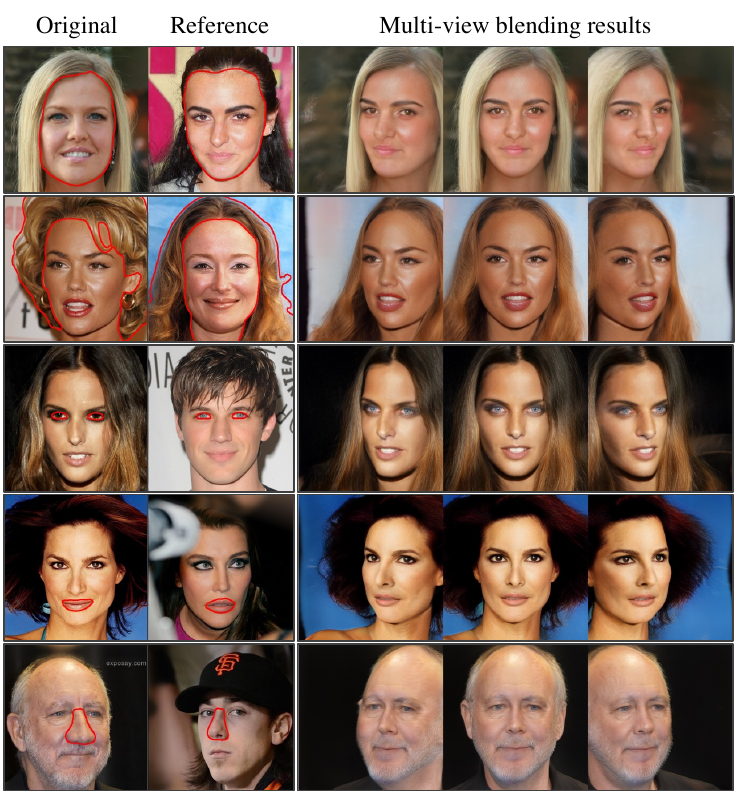}\\
\caption{Multi-view blending results in CelebA-HQ using EG3D.
} 
\label{fig:suppfigMultiviewCelebA}
\end{figure}
}

\newcommand{\suppfigMultiviewAFHQ}{

\begin{figure}[]
\centering
\vspace{-12.0mm}
\includegraphics[width=1.0\linewidth]{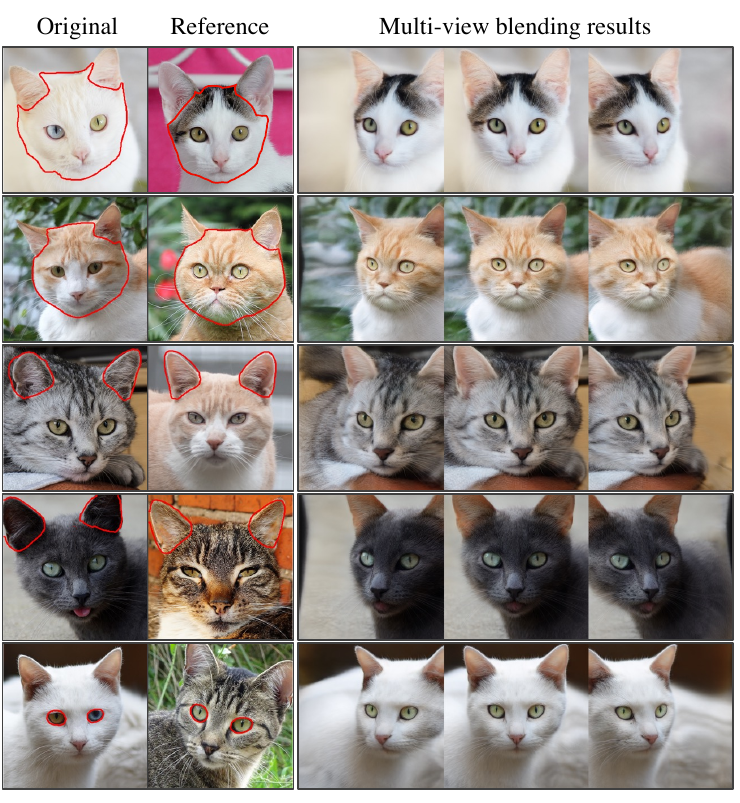}\\
\caption{Multi-view blending results in AFHQv2-Cat using EG3D.
} 
\label{fig:suppfigMultiviewAFHQ}
\end{figure}

}

\newcommand{\suppfigMultiviewShapeNet}{
\begin{figure}[]
\centering
\includegraphics[width=1.0\linewidth]{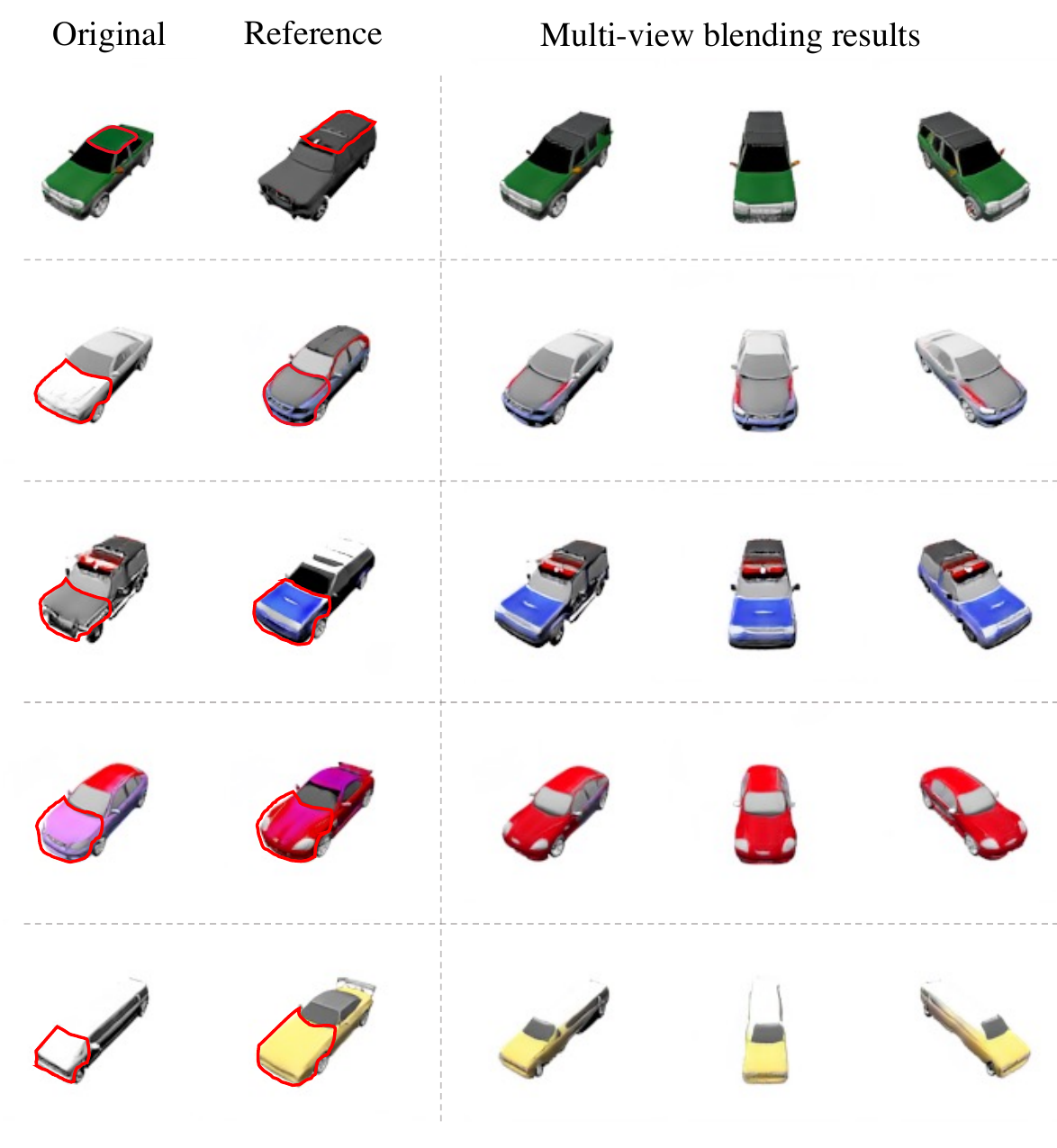}\\
\caption{Multi-view blending results in ShapeNet-Car~\cite{chang2015shapenet} using EG3D. The last two rows demonstrate that our method can achieve natural blending even when the sizes of the blending regions differ. Our 3D local alignment also performs well on the ShapeNet-Car dataset.
} 
\label{fig:suppfigMultiviewShapeNet}
\end{figure}
}

\newcommand{\suppfigMultiviewStyleSDF}{
\begin{figure}[]
\centering
\vspace{-28mm}
\includegraphics[width=1.0\linewidth]{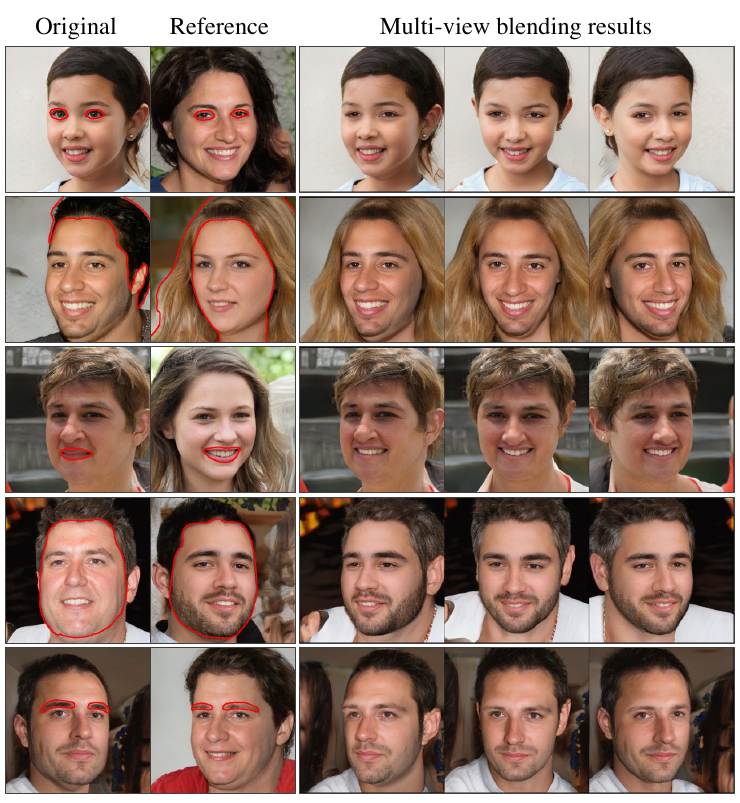}\\
\caption{Multi-view blending results in generated images using FFHQ-pretrained StyleSDF~\cite{or2022stylesdf}.
} 
\label{fig:suppfigMultiviewStyleSDF}
\end{figure}
}

\newcommand{\suppfigEyeglasses}{
\begin{figure}[t]
\centering
\ifdefined\ARXIV
\vspace{-8mm}
\fi
\includegraphics[width=1.0\linewidth]{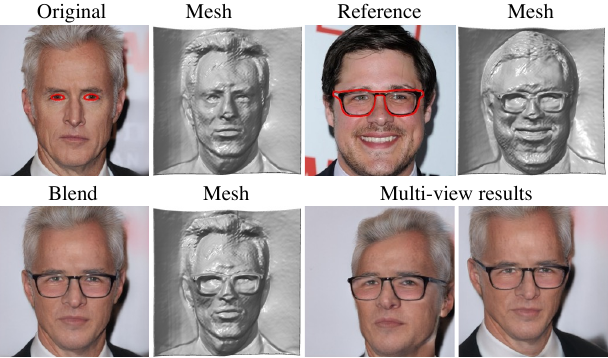}
\caption{Blending highly structured parts such as eyeglasses. Our blending method can reflect the high-fidelity 3D shape of eyeglasses and generate multi-view consistent results.}
\label{fig:eyeglasses}
\end{figure}
}

\newcommand{\suppfigOverlap}{
\begin{figure}[t]
\centering
\includegraphics[width=1.0\linewidth]{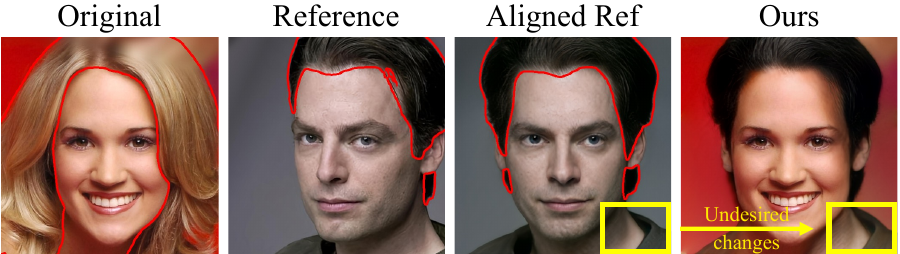}
\caption{A failure case of blending from long hair to short hair.}
\label{fig:overlap}
\end{figure}
}

\newcommand{\suppfigSDEditAFHQ}{
\begin{figure}[t]
\centering
\includegraphics[width=1.0\linewidth]{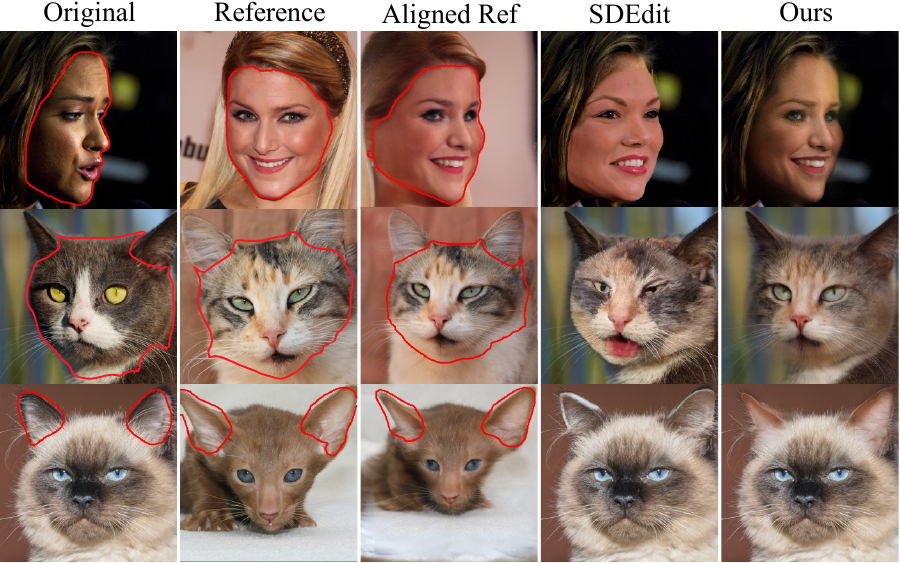}
\caption{Comparison with SDEdit in CelebA-HQ and AFHQv2-Cat test sets. SDEdit blending results often do not match reference images well. The identities of faces or ears have been changed in both CelebA-HQ and AFHQ datasets.}
\label{fig:sdeditAFHQ}
\end{figure}
}

\newcommand{\suppfigADDcomparison}{

\ifdefined\ARXIV
\begin{figure*}[t]
\centering
\includegraphics[width=1.0\linewidth]{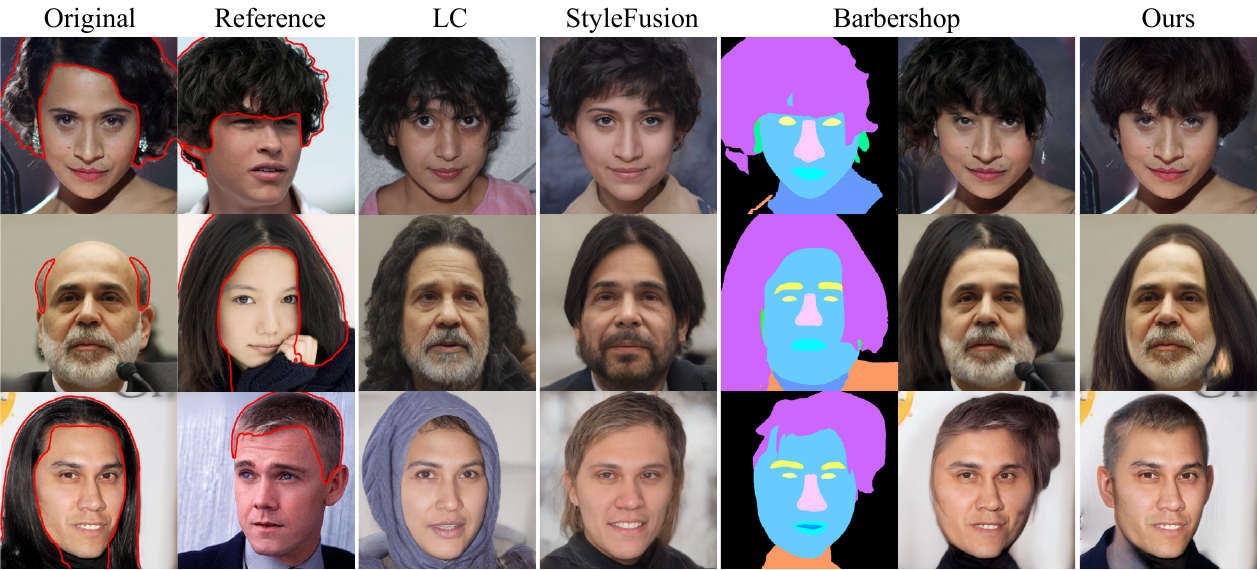}
\caption{Additional hair blending comparison with baselines. Latent Composition (LC) generates real-looking images but far from the input images. StyleFusion alters the hair length of the reference images. In StyleFusion and Barbershop, the hair and face poses do not match in each blending result. Ours shows the best results regarding 3D-aware alignment and identity preservation. 
}
\label{fig:identity}
\end{figure*}
\else
\begin{figure}[t]
\centering
\includegraphics[width=1.0\linewidth]{rebuttal/figures/identity_comparison.pdf}
\caption{Additional hair blending comparison with baselines. Latent Composition (LC) generates real-looking images but far from the input images. StyleFusion alters the hair length of the reference images. In StyleFusion and Barbershop, the hair and face poses do not match in each blending result. Ours shows the best results regarding 3D-aware alignment and identity preservation. 
}
\label{fig:identity}
\end{figure}
\fi

}

\begin{document}

\title{3D-aware Blending with Generative NeRFs}

\author{Hyunsu Kim \textsuperscript{1}
\qquad Gayoung Lee \textsuperscript{1}
\qquad Yunjey Choi \textsuperscript{1}
\qquad Jin-Hwa Kim \textsuperscript{1,2}
\qquad Jun-Yan Zhu\textsuperscript{3} \vspace{0.08in}\\
\textsuperscript{1}NAVER AI Lab
\qquad
\textsuperscript{2}SNU AIIS
\qquad
\textsuperscript{3}CMU
}
\maketitle
    

\newcommand{\paragrapht}[1]{\vspace{5pt} \noindent\textbf{#1}}

\def\r{\bm{r}}
\def\m{\mathbf{m}}
\def\LPIPS{\mathcal{L}_{\text{LPIPS}}}
\def\Lone{\mathcal{L}_{\text{1}}}
\def\Ltwo{\mathcal{L}_{\text{2}}}

\newcommand{\step}[1]{\textit{Step #1}\xspace}

\newcommand{\mask}[1]{\mathbf{m}_{\text{#1}}}
\renewcommand{\L}[1]{\mathcal{L}_{\text{#1}}}
\renewcommand{\G}[1]{G  _{\text{#1}}}
\newcommand{\cam}[1]{\mathbf{c}_{\text{#1}}}
\newcommand{\I}[1]{\text{I}_{\text{#1}}}
\newcommand{\Irotate}{\text{I}^{\text{R}}_{\text{ref}}}
\newcommand{\Ialigned}[1]{\tilde{\text{I}}_{\text{#1}}}

\newcommand{\Mesh}[1]{M_{\text{#1}}}
\newcommand{\Ray}[1]{\mathcal{R}_{\text{#1}}}

\newcommand{\Pt}[1]{P_{\text{#1}}}
\newcommand{\Ptnear}[1]{P^'_{\text{#1}}}
\newcommand{\w}[1]{\mathbf{w}_{\text{#1}}}

\newcommand{\MC}{\mathbf{MC}}

\definecolor{ori}{rgb}{0.92, 0.3, 0.46}
\definecolor{ref}{rgb}{0.36, 0.54, 0.76}

\newcommand{\revise}[1]{\textbf{\textcolor{alizarin}{#1}}}

\begin{abstract}

Image blending aims to combine multiple images seamlessly. 
It remains challenging for existing 2D-based methods, especially when input images are misaligned due to differences in 3D camera poses and object shapes.  
To tackle these issues, we propose a 3D-aware blending method using generative Neural Radiance Fields (NeRF), including two key components: 3D-aware alignment and 3D-aware blending. 
For 3D-aware alignment, we first estimate the camera pose of the reference image with respect to generative NeRFs and then perform pose alignment for objects.
To further leverage 3D information of the generative NeRF, we propose 3D-aware blending that utilizes volume density and blends on the NeRF's latent space, rather than raw pixel space. 
Collectively, our method outperforms existing 2D baselines, as validated by extensive quantitative and qualitative evaluations with FFHQ and AFHQ-Cat.
\ifdefined\ARXIV
Please find the code and data on our project \href{https://blandocs.github.io/blendnerf}{page}.
\fi

\end{abstract}

\section{Introduction}
Image blending aims at combining elements from multiple images naturally, enabling a wide range of applications in content creation, and virtual and augmented realities~\cite{zhu2021barbershop,hairnet}. However, blending images seamlessly requires delicate adjustment of color, texture, and shape, often requiring users' expertise and tedious manual processes. To reduce human efforts, researchers have proposed various automatic image blending algorithms, including classic methods~\cite{perez2003poisson,kwatra2003graphcut,burt1987laplacian,uyttendaele2001eliminating} and deep neural networks~\cite{zhu2015learning,wu2019gp,meng2021sdedit}.

\figmotivation
\figcomparison
Despite significant progress, blending two unaligned images remains a challenge. Current 2D-based methods often assume that object shapes and camera poses have been accurately aligned. As shown in \reffig{motivation}c, even slight misalignment can produce unnatural results, as it is obvious to human eyes that foreground and background objects were captured using different cameras. 
Several methods~\cite{hays2007scene, lin2018st, chen2019toward, sbai2021surprising, zhan2019spatial} warp an image via 2D affine transformation. However, these approaches do not account for 3D geometric differences, such as out-of-plane rotation and 3D shape differences. 
3D alignment is much more difficult for users and algorithms, as it requires inferring the 3D structure from a single view. 
Additionally, even though previous methods get aligned images, they blend images in 2D space. Blending images using only 2D signals, such as pixel values (RGB) or 2D feature maps, doesn't account for the 3D structure of objects.


To address the above issues, 
we propose a 3D-aware image blending method based on generative Neural Radiance Fields (NeRFs)~\cite{Chan2022,gu2021stylenerf,chan2021pi,pan2021shading,schwarz2020graf,zhou2021cips}.
Generative NeRFs learn to synthesize images in 3D using only collections of single-view images. 
Our method projects the input images to the latent space of generative NeRFs and performs 3D-aware alignment by novel view synthesis. We then perform blending on NeRFs' latent space. Concretely, we formulate an optimization problem in which a latent code is optimized to synthesize an image and volume density of the foreground close to the reference while preserving the background of the original. 

\reffig{comparison} shows critical differences between our approach and previous methods. \reffig{comparison}a shows a classic 2D blending method composing two 2D images without alignment. We then show the performance of the 2D blending method can be improved using our 3D-aware alignment with generative NeRFs as shown in \reffig{comparison}b. To further exploit 3D information, we propose to compose two images in the NeRFs' latent space instead of 2D pixel space. \reffig{comparison}c shows our final method.

We demonstrate the effectiveness of our 3D-aware alignment and 3D-aware blending (volume density) on unaligned images.
Extensive experiments show that our method outperforms both classic and learning-based methods regarding both photorealism and faithfulness to the input images. Additionally, our method can disentangle color and geometric changes during blending, and create multi-view consistent results. 
To our knowledge, our method is the first general-purpose 3D-aware image blending method capable of blending a diverse set of unaligned images.

\vspace{7mm}
\section{Related Work}
\paragraph{Image blending}
aims to compose different visual elements into a single image.  
Seminal works tackle this problem using various low-level visual cues, such as image gradients~\cite{perez2003poisson,jia2006drag,tao2010error,farbman2009coordinates,szeliski2011fast}, frequency bands~\cite{burt1987laplacian,brown2003recognising},  color and noise transfer~\cite{xue2012understanding,sunkavalli2010multi}, and segmentation~\cite{kwatra2003graphcut,rother2004grabcut,agarwala2004interactive,levin2007closed}. Later, researchers developed data-driven systems to compose objects with similar lighting conditions, camera poses, and scene contexts~\cite{lalonde2007photo,chen2009sketch2photo,hays2007scene}. 

Recently, various learning-based methods have been proposed, including blending deep features instead of pixels~\cite{suzuki2018neuralcollage,collins2020editing,fruhstuck2019tilegan} or designing loss functions based on deep features~\cite{zhang2020deep,zhang2021deep}. Generative Adversarial Networks (GAN) have also been used for image blending~\cite{wu2019gp,zhu2020indomaingan,collins2020editinginstyle,kim2021exploiting,chai2021latent,zhu2021barbershop,shi2022semanticstylegan}. For example, In-DomainGAN~\cite{zhu2020indomaingan} exploits GAN inversion to achieve seamless blending, and StyleMapGAN~\cite{kim2021exploiting} blends images in the spatial latent space. Recently, SDEdit~\cite{meng2021sdedit} proposes a blending method via diffusion models.
The above learning-based methods tend to be more robust than pixel-based methods. But given two images with large pose differences, both may struggle to preserve identity or generate unnatural effects.

In specific domains like faces~\cite{yang2011expression,dale2011video,nguyen2022deepfakesurvey,xu2022high} or hair~\cite{zhu2021barbershop,hairnet,chung2022hairfit,kim2022style}, multiple methods can swap and blend unaligned images. 
However, these methods are limited to faces or hair, and they often need 3D face morphable models~\cite{blanz1999morphable, ferrari2021sparse}, or multi-view images~\cite{nagrani2017voxceleb,kim2021k} to provide 3D information. Our method offers a general-purpose solution that can handle a diverse set of objects without 3D data.

\paragraph{3D-aware generative models.}
Generative image models learn to synthesize realistic 2D images~\cite{goodfellow2014generative,song2021scorebased,dhariwal2021diffusion,van2016conditional,chen2020generative}.
However, the original formulations do not account for the 3D nature of our visual world, making 3D manipulation difficult. 
Recently, several methods have integrated implicit scene representation, volumetric rendering, and GANs into generative NeRFs~\cite{schwarz2020graf,chan2021pi,niemeyer2021giraffe,deng2022gram}. Given a sampled viewpoint, an image is rendered via volumetric rendering and fed to a discriminator. 
For example, EG3D~\cite{Chan2022} uses an efficient 3D representation called tri-planes, and StyleSDF~\cite{or2022stylesdf} merges the style-based architecture and the SDF-based volume renderer. Multiple works~\cite{Chan2022,or2022stylesdf,zhou2021cips,gu2021stylenerf} have developed a two-stage model to generate high-resolution images. 
With GAN inversion methods~\cite{zhu2016generative,roich2022pivotal,feng2022near,parmar2022spatially,daras2021solving,zhu2020improved,dinh2022hyperinverter,bau2020semantic}, we can utilize these 3D-aware generative models to align and blend images and produce multi-view consistent 3D effects.



\paragraph{3D-aware image editing.}
Classic 3D-aware image editing methods can create 3D effects given 2D photographs~\cite{karsch2011rendering,chen20133,kholgade20143d}. However, they often require manual efforts to reconstruct the input's geometry and texture. Recently, to reduce manual efforts, researchers have employed generative NeRFs for 3D-aware editing. For example, EditNeRF~\cite{liu2021editing} uses separate latent codes to edit the shape and color of a NeRF object. 
NeRF-Editing~\cite{yuan2022nerf} proposes to reflect geometric edits in implicit neural representations. CLIP-NeRF~\cite{wang2022clip} uses a CLIP loss~\cite{radford2021learning} to ensure that the edited result corresponds to the input condition. In SURF-GAN~\cite{kwak2022injecting}, they discover controllable attributes using NeRFs for training a 3D-controllable GAN. Kobayashi et al.~\cite{kobayashi2022decomposing} enable editing via semantic scene decomposition. While the above works tackle various image editing tasks, we focus on a different task -- image blending, which requires both alignment and harmonization. 
Compared to previous image blending methods, our method addresses blending in a 3D-aware manner.



\section{Method}
\figglobalalign
\figblending

We aim to perform 3D-aware image blending using only 2D images, with target masks from users for both original and reference images. Our method consists of two stages: 3D-aware alignment and 3D-aware blending. Before we blend, we first align the pair of images regarding the pose. 
In \refsec{align}, we describe \textit{pose alignment} for entire objects and \textit{local alignment} for target regions. Then, we apply the 3D-aware blending method in the generative NeRF's latent space in \refsec{blend}. A variation of our blending method is illustrated in \refsec{oursWpoisson}. We combine Poisson blending with our method to achieve near-perfect background preservation.
We use EG3D~\cite{Chan2022} as our backbone, although other 3D-aware generative models, such as StyleSDF~\cite{or2022stylesdf}, can also be
\ifdefined\ARXIV
applied. See \refapp{stylesdf} for more details.
\else
applied; see Section E in the supplement. 
\fi

\subsection{3D-aware alignment}
\label{sec:align}


\textit{Pose alignment} is a requisite process of our blending method, as slight pose misalignment of two images can severely degrade blending quality as shown in \reffig{motivation}. 
To match the reference image $\I{ref}$ to the pose of the original image $\I{ori}$, we use a generative NeRF $\G{}$ to estimate the camera pose $\cam{}$ and the latent code $\w{}$ of each image. 
In \step{1} in \reffig{figglobalalign}, we first train and freeze a CNN encoder (\ie pose estimation network) to predict the camera poses of input images. During training, we can generate a large number of pairs of camera poses and images using generative NeRF and train the encoder $E$ using a pose reconstruction loss $\L{pose}$ as follows:
\begin{align}
\L{pose} = \mathbb{E}_{\w{}, \cam{}}{\lVert \cam{} - E(\G{RGB}(\w{},\cam{})) \rVert}_1,
    \label{eq:encoder}
\end{align}
where $\G{RGB}$ is an image rendering function with the generative NeRF $\G{}$, and ${\lVert \cdot \rVert}_1$ is the L1 distance.
The latent code $\w{}$ and camera pose $\cam{}$ are randomly drawn.

With our trained encoder, we estimate the camera poses $\cam{ori}$ and $\cam{ref}$ (defined as Euler angles $\cam{} \in SO3$) of the original and reference images, respectively. 
Given the estimated camera poses, we project input images $\I{ori}$ and $\I{ref}$ to the latent codes $\w{ori}$ and $\w{ref}$ using Pivotal Tuning Inversion (PTI)~\cite{roich2022pivotal}. 
We optimize the latent code $\w{}$ using the reconstruction loss $\L{rec}$ as follows:
\begin{align}
    \L{rec} = {\lVert \I{} - \G{RGB}(\w{},\cam{}) \rVert}_1 + \LPIPS \big(\I{}, \G{RGB}(\w{},\cam{}) \big),
    \label{eq:inversion_w_main}
\end{align}
where $\LPIPS$ is a learned perceptual image patch similarity (LPIPS)~\cite{zhang2018lpips} loss. For more accurate inversion, we fine-tune the generator $\G{}$. Inversion details are described in
\ifdefined\ARXIV
\refapp{suppInversion}.
\else
Section B in the supplement.
\fi
Finally, as shown in \step{2} of \reffig{figglobalalign}, we can align the reference image as follows:
\begin{align}
    \text{I}^{\text{R}}_{\text{\textcolor{ref}{ref}}} = \G{RGB}(\w{\textcolor{ref}{ref}},\cam{\textcolor{ori}{ori}}).
\end{align}


While \textit{pose alignment} can align two entire objects, further alignment in editing regions may still be necessary due to variations in scale and translation between object instances.
 To align target editing regions such as the face, eyes, and ears, we can further employ \textit{local alignment} in the loosely aligned dataset (AFHQv2). 
The Iterative Closest Point (ICP) algorithm~\cite{besl1992method,chen1992object} is applied to meshes, which can adjust their scale and translation. 
 For further details, please refer to
\ifdefined\ARXIV
\refapp{local_align}.
\else
Section C in the supplement.
\fi

\subsection{3D-aware blending}
\label{sec:blending}
We aim to find the best latent code $\w{edit}$ to synthesize a seamless and natural output. To achieve this goal, we exploit both 2D pixel constraints (RGB value) and 3D geometric constraints (volume density). With the proposed image-blending and density-blending losses, we optimize the latent code $\w{edit}$, by matching the foreground with the reference and the background with the original.
\ifdefined\ARXIV
Algorithm details such as optimization latent spaces and input masks can be found in \refapp{blendingDetails}.
\fi

\paragraph{Image-blending} algorithms are often designed to match the color and details of the original image (\ie background) while preserving the structure of  the reference image (\ie foreground)~\cite{perez2003poisson}. 
As shown in  \reffig{blending}, our image-blending loss matches the color and perceptual similarity of the original image using a combination of L1 and LPIPS~\cite{zhang2018lpips}, while matching the reference image's details using LPIPS loss alone. L1 loss in the reference can lead to overfitting to the pixel space. 
Let $\I{edit}$ be the rendered image from the latent code $\w{edit}$. We define the image-blending loss 
as follows:
\begin{align}
    \L{image} &= {\lVert(\mathbf{1} - \m) \circ \I{edit} - (\mathbf{1} - \m) \circ \I{ori} \rVert}_1 \nonumber\\ 
    &+ \lambda_1\LPIPS \big((\mathbf{1} - \m) \circ \I{edit}, (\mathbf{1} - \m) \circ \I{ori} \big) \nonumber\\ 
    &+ \lambda_2\LPIPS (\m \circ \I{edit}, \m \circ \I{ref}), \label{eq:color_edit}
\end{align}
where $\circ$ denotes element-wise multiplication. Here, $\lambda_1$ and $\lambda_2$ balance each loss term.

\paragraph{Density-blending}
\label{sec:blend}
is our key component in 3D-aware image blending. If we use only image-blending loss, the blending result easily falls blurry and may not reflect the reference object correctly. Especially, a highly structured object such as hair is hard to be blended in the 3D NeRF space without volume density, as shown in \reffig{densityResult}.
By representing each image as a NeRF instance, we can calculate the density $\bm{\sigma}$ of a given 3D location $\bm{x} \in \R^3$. Let $\Ray{ref}$ and $\Ray{ori}$ be the set of rays $\bm{r}$ passing through the interior and exterior of the target mask $\mathbf{m}$, respectively. For the 3D sample points along the rays $\Ray{ref}$, we aim to match the density field between the reference and our output result, as shown as the sample points in a green ray in \reffig{blending}.
For 3D sample points in $\Ray{ori}$, we also match the density field between the original and the result, as shown as the sample points in a red ray in \reffig{blending}. 
Let $G_{\sigma}(\w{}; \bm{x})$ be the density of a given 3D point $\bm{x}$ with a given latent code $\mathbf{w}$. Our density-blending loss can be formulated as follows:
\begin{align}
    \mathcal{L}_{\textrm{density}} =& 
    \sum_{\bm{r} \in \Ray{ref}} \sum_{\bm{x} \in \bm{r}} {\lVert G_{\sigma}(\w{edit}; \bm{x}) - G_{\sigma}(\w{ref}; \bm{x})\rVert}_1 \nonumber\\ 
    &+ \sum_{\bm{r} \in \Ray{ori}} \sum_{\bm{x} \in \bm{r}} {\lVert G_{\sigma}(\w{edit}; \bm{x}) - G_{\sigma}(\w{ori}; \bm{x})\rVert}_1.
    \label{eq:density_edit}
\end{align}

Our final objective function includes both image-blending loss and density-blending loss: 
\begin{equation}
\label{eq:blend}
\begin{aligned}
    \mathcal{L} = \lambda \mathcal{L}_{\text{image}} + \mathcal{L}_{\text{density}},
\end{aligned}
\end{equation}
where $\lambda$ is the hyperparameter that controls the contribution of the image-blending loss. If our user wants to blend the shape together without reflecting the color of reference, $\lambda_2$ in \refeq{color_edit} is set to $0$. Otherwise, we can set $\lambda_2$ to a positive number to reflect the reference image's color and geometry as shown in \reffig{disentangle}.

\subsection{Combining with Poisson blending}
\label{sec:oursWpoisson}

While our method produces high-quality blending results, incorporating Poisson blending~\cite{perez2003poisson} further improves the preservation of the original image details. \reffig{figOurs_w_wo_PB} shows the effect of Poisson blending with our method.
We perform Poisson blending between the original image and the blended image generated by our 3D-aware blending method. 
Our blending method is modified in two ways. 1) In the initial blending stage, we only preserve the area around the border of the mask instead of all parts of the original image, as we can directly use the original image in the Poisson blending stage. We can reduce the number of iterations from $200$ to $100$, as improved faithfulness scores are easily achieved; see m$L_2$ and LPIPS$_m$ in Tables \ref{tbl:baseline_edit_celeba} and \ref{tbl:baseline_edit_afhq}.
2) Instead of using the latent code of the original image $\w{ori}$ as the initial value of $\w{edit}$, we use the latent code of the reference image $\w{ref}$. This allows us to instantly reflect the identity of the reference image and only optimize $\w{edit}$ to reconstruct a small region near the mask boundary of the original image. Note that this is an optional choice, as our method \emph{without} Poisson blending has already outperformed all the baselines, as shown in Tables \ref{tbl:userstudyCelebA} and \ref{tbl:userstudyAFHQ}. 



\figOursWWOPB

\tabBaselineEditCelebAHQ
\tabBaselineEditAFHQvtwo

\section{Experiments}

In this section, we show the advantage of our full method over several existing methods and ablated baselines. In \refsec{expr_setup}, we describe our experimental settings, including baselines, datasets, and evaluation metrics. In \refsec{expr_comparison}, we show both quantitative and qualitative comparisons.  
In addition to the automatic quantitative metrics, 
our user study shows that our method is preferred over baselines regarding photorealism. In \refsec{expr_ablation}, we analyze the effectiveness of each module via ablation studies. Lastly, \refsec{expr_nerf} shows useful by-products of our method, such as generating multi-view images and controlling the color and geometry disentanglement. 
\ifdefined\ARXIV
We provide experimental details in \refapp{suppexpr} and runtime of our method in Appendix~\ref{sec:suppInversion}, \ref{sec:local_align}, and \ref{sec:blendingDetails}.
\else
Please see the supplement for experimental details, video results on the webpage, additional results, \etc.
\fi

\subsection{Experimental setup}
\label{sec:expr_setup}
\paragraph{Baselines.} We compare our method with various image blending methods using only 2D input images. For classic methods, we run Poisson blending~\cite{perez2003poisson}, a widely-used gradient-domain editing method. We also compare with several recent learning-based methods~\cite{chai2021latent,kim2021exploiting,karras2021alias,meng2021sdedit}. Latent Composition~\cite{chai2021latent} utilizes the compositionality in GANs by finding the latent code of the roughly collaged inputs on the manifold of the generator. StyleMapGAN~\cite{kim2021exploiting} proposes the spatial latent space for GANs to enable local parts blending by mixing the spatial latent codes. Recently, Karras \etal~\cite{karras2021alias} proposed StyleGAN3, which provides rotation equivariance. Therefore, we additionally show their blending results by finding the latent code of the composited inputs on the StyleGAN3-R manifold. Both $\mathcal{W}$ and $\mathcal{W}+$ of StyleGAN3-R latent spaces are tested. SDEdit~\cite{meng2021sdedit} is a diffusion-based blending method that produces a natural-looking result by denoising the corrupted image of a composite image.



\paragraph{Datasets.} 

We use FFHQ~\cite{karras2019stylegan} and AFHQv2-Cat datasets~\cite{choi2020starganv2} for model training. We use pose alignment for both datasets and apply further local alignment to the loosely aligned dataset (AFHQ).

To test blending performance, we use CelebA-HQ~\cite{karras2017progressivegan} for the FFHQ-trained models and AFHQv2-Cat test sets for the AFHQ-trained models. We randomly select 250 pairs of images from each dataset for an original and reference image. We also create a target mask for each pair to automatically simulate a user input using pretrained semantic segmentation networks~\cite{yu2018bisenet, zhang2021datasetgan,chen2017rethinking}. We blend 5 and 3 semantic parts in each pair of images for CelebA-HQ and AFHQ, respectively. The total number of blended images in each method is 1,250 (CelebA-HQ) and 750 (AFHQv2-Cat). We also include results on ShapeNet-Car dataset~\cite{chang2015shapenet} to show that our method works well for non-facial data.

\paragraph{Evaluation metrics.} 
For evaluation metrics, we use masked $L_2$, masked LPIPS~\cite{zhang2018lpips} and Kernel Inception Score (KID)~\cite{kid}. Masked $L_2$ (m$L_2$) is the $L_2$ distance between the original image and the blended image on the exterior of the mask, measuring the preservation of non-target areas of the original image. Unlike background regions, a pixel-wise loss is too strict for the target area changed during blending. We measure the perceptual similarity metric (LPIPS)~\cite{zhang2018lpips} for the blended regions, which are called masked LPIPS (LPIPS$_m$) used in previous methods~\cite{huh2020transforming, meng2021sdedit}. Kernel Inception Score (KID)~\cite{kid} is widely used to quantify the realism of the generated images regarding the real data distribution. We compute KID between blended images and the training dataset using the \texttt{clean-fid} library~\cite{parmar2022aliased}.

\paragraph{User study.} 
To further examine the effectiveness of our 3D-aware blending method, we conduct a user study for photorealism. Our target goal is to edit the original image, so we exclude baselines that show highly flawed preservation of the original image. Human evaluates pairwise comparison of blended images between our method and one of the baselines. The user selects more real-looking images. We collect  5,000 comparison results via Amazon Mechanical Turk (MTurk). 
\ifdefined\ARXIV
See \refapp{suppuserstudy} for more details.
\fi









\subsection{Comparison with baselines}
\label{sec:expr_comparison}

Here we compare our method with baselines in two variations. In the \textit{w/o align} setting, we do not apply our 3D-aware alignment to baselines. In the \textit{w/ align} setting, we align the reference image with our 3D-aware alignment. This experiment demonstrates the effectiveness of our proposed method. 1) Our alignment method consistently improves all baselines in all evaluation metrics: KID, LPIPS$_m$, and masked $L_2$. 2) Our 3D-aware blending method outperforms all baselines, including those that use our alignment method. We also report the combination of our method and Poisson blending to achieve better background preservation, as the perfect inversion is still hard to be achieved in GAN-based methods. 

\reftbl{baseline_edit_celeba} shows comparison results in CelebA-HQ. The left side of the table includes all the baselines without our 3D-aware alignment. All metrics are worse than the right side of the table (w/ alignment). This result reveals that alignment between the original and reference image affects overall editing performance. 
\reftbl{baseline_edit_afhq} shows comparison results in AFHQv2-Cat. It shows the same tendency as \reftbl{baseline_edit_celeba}. 
\ifdefined\ARXIV
More comparison results are included in \refapp{suppAdditional}.
\fi

Our method performs well regarding all metrics. Combined with Poisson blending, our method outperforms all baselines. Poisson blending and StyleMapGAN ($16\times16$, $32\times32$) show great faithfulness to the input images but suffer from artifacts. Latent Composition, StyleMapGAN ($8\times8$), and StyleGAN3 $\mathcal{W}$ produce realistic results but far from the input images. The identities of the original and reference images have changed, which is reflected by a worse LPIPS$_m$ and m$L_2$.
SDEdit fails to reflect the reference image and shows worse LPIPS$_m$.
StyleGAN3 $\mathcal{W}+$ often shows entirely collapsed images. Our method preserves the identity of the original image and reflects the reference image well while producing realistic outputs. 









\UserCelebA
\UserAFHQ

\paragraph{User study.}

We note that KID has a high correlation with background preservation. Unfortunately, it fails to capture the boundary artifacts and foreground image quality, especially for small foregrounds. To further evaluate the realism score of results, we conduct a human perception study for our method and other baselines, which shows great preservation scores m$L_2$. As shown in Tables \ref{tbl:userstudyCelebA} and \ref{tbl:userstudyAFHQ}, MTurk participants prefer our method to other baselines regarding the photorealism of the results. Our method, as well as the combination of ours with Poisson blending, outperforms the baselines.
SDEdit with our 3D-aware alignment shows a comparable realism score with ours, but it can not reflect the reference well, as reflected in worse LPIPS$_{m}$ score in \reftbl{baseline_edit_celeba}. Similar to Tables \ref{tbl:baseline_edit_celeba} and \ref{tbl:baseline_edit_afhq}, Mturk participants prefer baselines with alignment to their unaligned counterparts.

\figablationalign
\figPoissonBaseline
\densityResult
\subsection{Ablation study}
\label{sec:expr_ablation}
\paragraph{3D-aware alignment} is an essential part of image blending. As discussed in \refsec{expr_comparison}, our alignment provides consistent improvements in all baseline methods. Moreover, it plays a crucial role in our blending approach. \reffig{ablationalign} shows the importance of 3D-aware alignment, where the lack of alignment in the reference images result in degraded blending results (\reffig{ablationalign}c). Specifically, the woman's hair appears blurry, and the size of the cat's eyes looks different. Aligned reference images can generate realistic blending results (\reffig{ablationalign}e) in our 3D-aware blending method.

\paragraph{Density-blending loss} gives rich 3D signals in the blending procedure. \refsec{blending} explains how we can exploit volume density fields in blending. Delicate geometric structures, such as hair, can not be easily blended without awareness of 3D information. \reffig{densityResult} shows an ablation study of our density-blending loss. In the bottom left, the hair looks blurry in the blended image, and the mesh of the result shows shorter hair than that in the reference image. In the bottom right, the well-blended image and corresponding mesh show that our density-blending loss contributes to capturing the highly structured object in blending.
\ifdefined\ARXIV
We provide an additional ablation study using StyleSDF~\cite{or2022stylesdf} in \refapp{stylesdf}.
\fi

\paragraph{Combination with Poisson blending.} 
In Tables \ref{tbl:baseline_edit_celeba} and \ref{tbl:baseline_edit_afhq}, 
we report the combination of our method and Poisson blending. It shows Poisson blending further enhances the performance of our method in all automatic metrics: KID, m$L_2$, and LPIPS$_m$. In the realism score of human perception, ours with Poisson blending enhance the score, as shown in green numbers of Tables \ref{tbl:userstudyCelebA} and \ref{tbl:userstudyAFHQ}. However, combining Poisson blending with each baseline does not have meaningful benefits, as shown in ~\reffig{poisson_baseline}. Baselines still show artifacts or fail to reflect the identity of the reference.



\figdisentangle

\subsection{Additional advantages of NeRF-based blending}
\label{sec:expr_nerf}

In addition to increasing blending quality, our 3D-aware method enables additional capacity: color-geometry disentanglement and multi-view consistent blending. As shown in \reffig{disentangle}, we can control the influence of color in blending. The results with $\mathcal{L}_{\text{image}}$ have a redder color than the results without the loss. If we remove or assign a lower weight to the image-blending loss on reference ($\lambda_2$ in \refeq{color_edit}), we can reflect the geometry of the reference object more than the color. In contrast, we can reflect colors better if we give a larger weight to $\lambda_2$. Note that we always use the image-blending loss on the original image to preserve it better.
\ifdefined\ARXIV
An additional ablation study using StyleSDF is included in \refapp{stylesdf}.
\fi

A key component of generative NeRFs is multi-view consistent generation. After applying the blending procedure described in \refsec{blending}, we have an optimized latent code $\w{edit}$. Generative NeRF can synthesize a novel view blended image using $\w{edit}$ and a target camera pose. \reffig{multiview} shows the multi-view consistent blending results in CelebA-HQ, AFHQv2-Cat, and ShapeNet-Car~\cite{chang2015shapenet}. 
\ifdefined\ARXIV
We provide more multi-view results for EG3D and StyleSDF in \refapp{suppAdditional}.
\else
In Section I in the supplement and the attached website, we provide more multi-view blending results and videos for EG3D and StyleSDF~\cite{or2022stylesdf}.
\fi

\figmultiview

\section{Discussion and Limitations} 
\label{sec:discussion}

Our method exploits the capability of NeRFs to align and blend images in a 3D-aware manner only with a collection of 2D images. Our 3D-aware alignment boosts the quality of existing 2D baselines. 3D-aware blending exceeds improved 2D baselines with our alignment method and shows additional advantages such as color-geometry disentanglement and multi-view consistent blending. We hope our approach paves the road to 3D-aware blending. Recently, 3DGP~\cite{3dgp} presents a 3D-aware GAN, handling non-alignable scenes captured from arbitrary camera poses in real-world environments. Since our approach relies solely on a pre-trained generator, it can be readily extended to blend unaligned multi-category datasets such as ImageNet~\cite{deng2009imagenet}.

Despite improvements over existing blending baselines, 
our method depends on GAN inversion, which is a bottleneck of the overall performance regarding quality and speed. 
\reffig{figLimitInv} shows the inversion process can sometimes fail to accurately reconstruct the input image. We cannot obtain an acceptable inversion result if an input image is far from the average face generated from the mean latent code $\w{avg}$. We also note the camera pose inferred by our encoder should not be overly inaccurate.
Currently, the problem is being addressed by combining our method with Poisson blending. However, more effective solutions may be available with recent advances in 3D GAN inversion techniques~\cite{xie2022high,ko20233d}. In the future, to enable real-time editing, we could explore training an encoder~\cite{dinh2022hyperinverter,kim2021exploiting} to blend images using our proposed loss functions.


\figLimitInv
\vspace{-3mm}
\paragraph{Acknowledgments.} We would like to thank Seohui Jeong, Che-Sang Park, Eric R. Chan, Junho Kim, Jung-Woo Ha, Youngjung Uh, and other NAVER AI Lab researchers for their helpful comments and sharing of materials. All experiments were conducted on NAVER Smart Machine Learning (NSML) platform~\cite{kim2018nsml, sung2017nsml}.

\clearpage 
{\small
\bibliographystyle{ieee_fullname}
\bibliography{main}
}

\clearpage
\appendix

\ifdefined\ARXIV
\paragraph{Overview of Appendix.} 
\else
\paragraph{Overview of the supplementary material.}
\fi
\begin{itemize} 
\item  Please refer to the code, data, and results on our website: \url{https://blandocs.github.io/blendnerf}.
  \item Experimental details are described in \refapp{suppexpr}.
  \item Details of inversion are described in \refapp{suppInversion}.
  \item Details of local alignment are described in \refapp{local_align} with an extra user study.
  \item Details of 3D-aware blending are described in \refapp{blendingDetails}.
  \item 3D-aware blending in StyleSDF is described in \refapp{stylesdf}. Our method can be applied to Signed Distance Fields (SDF) beyond NeRFs.
  \item Details of user studies are described in \refapp{suppuserstudy}.
  \item Failure cases are described in \refapp{limit}.  
  \item Societal impact is discussed in \refapp{societal}.
  \item Additional qualitative results are in \refapp{suppAdditional}.
  
\end{itemize}

\section{Experimental details}
\label{sec:suppexpr}

\paragraph{Baselines.} 
\begin{itemize}
\item 
{\bf Poisson Blending}~\cite{perez2003poisson} is implemented in OpenCV~\cite{opencv_library}. We use \texttt{cv2.seamlessClone} in the \texttt{cv2.NORMAL_CLONE} cloning type. 
\item
{\bf StyleGAN3}:  As there is no official projection code in StyleGAN3~\cite{karras2021alias}, we use unofficial implementation\footnote{https://github.com/PDillis/stylegan3-fun}. We follow the hyperparameters of StyleGAN2~\cite{karras2020analyzing} official projection algorithm\footnote{https://github.com/NVlabs/stylegan2-ada-pytorch} and set the number of optimization iterations as 1,000. 

\item {\bf StyleMapGAN}~\cite{kim2021exploiting} introduce \textit{stylemap}, which has spatial dimensions in the latent space. We use the official pretrained networks with $8 \times 8$ and $16 \times 16$ \textit{stylemap} for AFHQ~\cite{choi2020starganv2} and $32 \times 32$ \textit{stylemap} for FFHQ~\cite{karras2019stylegan}. 
\item 
{\bf SDEdit}~\cite{meng2021sdedit} transforms a noise-added image into a realistic image through iterative denoising. The total denoising step is a sensitive hyperparameter that decides blending quality. If it is too small, blending results are faithful to the input images but less realistic. If it is too large,  blending results are less faithful to the input images but more realistic. We carefully select the number of iterations for the best quality; 300 for the small editing parts (eyes, nose, and lip) and 500 for the large editing parts (face and hair). To exploit the FFHQ-pretrained model~\cite{baranchuk2021label}\footnote{https://github.com/yandex-research/ddpm-segmentation}, we implement guided-diffusion~\cite{dhariwal2021diffusion} version of SDEdit. 
\item 
{\bf Latent Composition}~\cite{chai2021latent} requires a mask to decide which area preserve. In our blending experiment, we need to preserve both the original and reference image, so we use a mask for the entire image.
\end{itemize}

\paragraph{Datasets.} We select FFHQ~\cite{karras2019stylegan} and AFHQv2-Cat~\cite{choi2020starganv2} for our comparison experiments. FFHQ has $1024 \times 1024$ images, and AFHQ has $512 \times 512$ images. In SDEdit~\cite{meng2021sdedit,baranchuk2021label}, the pretrained model is trained on $256 \times 256$ FFHQ. Our backbone network EG3D~\cite{Chan2022} is trained on $512 \times 512$ FFHQ. 
\ifdefined\ARXIV
   In \reftbl{baseline_edit_celeba},
\else
   In Table 1 of the paper,
\fi
we upsample the blending results of SDEdit and EG3D using bilinear interpolation for a fair comparison. Other baselines in FFHQ and all methods in AFHQ have the same resolution with the corresponding datasets. As the StyleMapGAN network is trained on AFHQv1, we fine-tune the networks using AFHQv2-Cat. As EG3D uses a different crop version of FFHQ compared to the original FFHQ, we fine-tune the EG3D using the original crop version of FFHQ. We use ShapeNet-Car~\cite{chang2015shapenet} ($128 \times 128$) to further demonstrate the effectiveness of our method. 

\paragraph{Metrics.} 
\ifdefined\ARXIV
   In Tables~\ref{tbl:baseline_edit_celeba} and~\ref{tbl:baseline_edit_afhq},
\else
   In Tables 1 and 2 of the main paper,
\fi
we use the masked LPIPS (LPIPS$_m$)~\cite{zhang2018lpips} to evaluate the faithfulness to the reference object. It needs a reference image to compute the score, and we use aligned reference images as pseudo-ground-truth images in both experiments: without and with 3D-aware alignment. Additionally, we always apply the alignment to our method in 
\ifdefined\ARXIV
   Tables~\ref{tbl:baseline_edit_celeba} and~\ref{tbl:baseline_edit_afhq}.
\else
   Tables 1 and 2 of the main paper.
\fi

\paragraph{Hyperparameters.} 

In the 3D-aware blending, we optimize the latent code $\w{edit} \in \R^{14\times512}$ of the $\mathcal{W}+$ space~\cite{abdal2019image2stylegan} for 200 iterations. The initial value of $\w{edit}$ is the latent code of the original image $\w{ori}$. Adam~\cite{kingma2014adam} optimizer is used with 0.02 learning rate, $\beta_1$ = 0.9, and $\beta_2$ = 0.999. We reformulate 
\ifdefined\ARXIV
   \refeq{color_edit}
\else
   Eqn. 4 in the main paper 
\fi
to specify the hyperparameters as follows:
\begin{align}
    \L{image} =& {\lVert (\mathbf{1} - \m) \circ \I{edit} - (\mathbf{1} - \m) \circ \I{ori} \rVert}_1  \nonumber\\ 
    &+ \lambda_1 \LPIPS \big((\mathbf{1} - \m) \circ \I{edit}, (\mathbf{1} - \m) \circ \I{ori} \big) \nonumber\\ 
    &+ \lambda_2 \lambda_m \LPIPS (\m \circ \I{edit}, \m \circ \I{ref}), \label{eq:supp_color_edit}
\end{align}
where $\lambda_1=1$, $\lambda_2=0.5$ for AFHQ, $\lambda_2=0.1$ for FFHQ except hair ($\lambda_2=0.3$). $\lambda_m = \frac{3 \cdot H \cdot W}{\lVert \m \rVert_0}$~is a weighting parameter to give a high weight for the small target blending region $\m$; $H$ and $W$ denotes the height and width of the image, $\m$ is a binary mask $\m \in \{0, 1\}^ {3 \times H \times W}$, and $||\cdot||_0$ is the $L_0$ norm that counts the total number of non-zero elements.
\ifdefined\ARXIV
   \refeq{density_edit}
\else
   Eqn. 5 in the paper
\fi
 is reformulated as follows:
\begin{align}
    \mathcal{L}_{\textrm{density}} =& 
    \lambda_m \sum_{\bm{r} \in \Ray{ref}} \sum_{\bm{x} \in \bm{r}} {\lVert G_{\sigma}(\w{edit}; \bm{x}) - G_{\sigma}(\w{ref}; \bm{x}) \rVert}_1 \nonumber\\ 
    &+ \sum_{\bm{r} \in \Ray{ori}} \sum_{\bm{x} \in \bm{r}} {\lVert G_{\sigma}(\w{edit}; \bm{x}) - G_{\sigma}(\w{ori}; \bm{x})\rVert}_1.
    \label{eq:supp_density_edit}
\end{align}
The results of L1 loss in the image- and density-blending are normalized by the number of elements. We set $\lambda$ in 
\ifdefined\ARXIV
   \refeq{blend}
\else
   Eqn. 6 in the paper 
\fi
to $10$. 
\section{Inversion details}
\label{sec:suppInversion}
We train our encoder to predict the camera pose of an image. It has a similar structure to the EG3D discriminator~\cite{Chan2022} but does not include minibatch discrimination~\cite{salimans2016improved} and involves adjustment of the number of input and output channels.
Given that training images for generative NeRFs~\cite{Chan2022,or2022stylesdf} are pre-aligned with respect to scale and translation, the camera pose of the input image can be simplified as a rotation matrix; EG3D also samples camera poses on the surface of a sphere.
As directly predicting the camera extrinsics $\in \R^{4\times4}$ is not easy, we convert the extrinsics to Euler angles $\in \R^{3}$. During inference, we transform it back to the camera extrinsics after obtaining the Euler angles from the encoder, which produces a more accurate pose estimation. Let the camera pose of the input image be $\cam{}$.

We adopt Pivotal Tuning Inversion (PTI)~\cite{roich2022pivotal} as our inversion method. In the first stage, we optimize the latent code $\w{} \in \mathcal{W}$ using the reconstruction loss $\L{rec}$ as follows:
\begin{align}
    \L{rec} = {\lVert \I{} - \G{RGB}(\w{},\cam{}) \rVert}_1 +  \LPIPS \big(\I{}, \G{RGB}(\w{},\cam{}) \big),
    \label{eq:inversion_w}
\end{align}
where $\I{}$ is an input image and $\G{RGB}$ is an image rendering function based on the generative NeRF $\G{}$. ${\lVert \cdot \rVert}_1$ is the L1 distance and $ \LPIPS$ is a learned perceptual image patch similarity (LPIPS)~\cite{zhang2018lpips} loss. We optimize the latent code $\w{} \in \R^{512}$ for 300 iterations.

In the second stage, we fine-tune the generative NeRF $\G{}$ using the same reconstruction loss $\L{rec}$ and an additional regularization loss $\L{reg}$ as follows:
\begin{align}
    \L{reg} = & {\lVert G^f_{\text{RGB}}(\w{s},\cam{}) - \G{RGB}(\w{s},\cam{})\rVert}_1 \nonumber\\ 
    &+ \LPIPS \big(G^f_{\text{RGB}}(\w{s},\cam{}), \G{RGB}(\w{s},\cam{}) \big),
    \label{eq:inversion_g}
\end{align}
where $G^f_{\text{RGB}}$ represents the frozen version of $\G{RGB}$, and $\w{s}$ is a randomly sampled latent code followed by linear interpolation with $\w{}$. 
The interpolation parameter is also randomly sampled in $[0, 1]$. The final loss function for optimizing $\G{}$ is as follows:
\begin{align}
    \mathcal{L} = \L{rec} + \lambda_{\text{reg}} \L{reg},
    \label{eq:inversion_all}
\end{align}
where $\lambda_{\text{reg}}=0.1$ and we optimize the $\G{}$ for 100 iterations.




\paragraph{Runtime.} As described above, our inversion method consists of two-stage optimization. For a single image, the first stage (latent code) takes 23.6s, and the second stage (generative NeRF) takes 20.4s. 
We test the runtime on a single A100 GPU.

\suppfiglocalalign

\section{Local alignment}
\label{sec:local_align}

Local alignment is a fine-grained alignment between the target regions of two images. Even though we have matched two images through pose alignment, the scale and translation of target regions (\eg face, eyes, ears, \etc) might need to be further aligned, as the location and size of each object part differ across two object instances. \reffig{suppfiglocalalign} illustrates our local alignment algorithm. In \step{1}, we need to obtain 3D meshes $\Mesh{ori}$ and $\Mesh{ref}$ of input images using the Marching Cube algorithm~\cite{lorensen1987marching} and the density fields.
In \step{2}, we first cast rays through the interior of the target region $\mask{}$ and determine the intersected 3D points with the mesh $\Mesh{ori}$. We define the set of points as a 3D point cloud $\Pt{ori}$. Similarly, we can get the reference's 3D point cloud $\Pt{ref}$. 

Given two sets of 3D point clouds $\Pt{ref}$ and $\Pt{ori}$, we use the Iterative Closest Point (ICP) algorithm~\cite{besl1992method,chen1992object} to obtain the 3D transformation matrix $\mathbf{W} \in \R^{4\times4}$. When we sample the 3D points to align the reference image, we transform the 3D point coordinates by multiplying the coordinate of sampled points with $\mathbf{W}$. We only use uniform scaling and translation in ICP, as we have already matched the rotation through \textit{pose alignment}
\ifdefined\ARXIV
in \refsec{align}.
\else
as described earlier in Section 3.1 of the main paper.
\fi
Finally, we can generate the fully aligned reference image $\Ialigned{ref}$ as follows:
\begin{align}
    \Ialigned{ref} = \G{RGB}(\w{ref},\cam{ori}; \mathbf{W}).
\end{align}

\myparagraph{Iterative Closest Point (ICP)}~\cite{besl1992method,chen1992object} is an algorithm to minimize the difference between two 3D point clouds. It is an iterative optimization process until we meet the threshold $\tau$ of difference or the maximum number of iterations $K=20$. 
Please note that we do not use rotation in ICP as we align the rotation for entire objects in \textit{pose alignment}, and refer to the exact details of ICP in our code.

\ablationWarping

\suppfigablationWarping

\paragraph{3D transformation in the NeRF space.}

$\mathbf{W}$ is the 3D transformation matrix computed by the ICP algorithm. It transforms the point cloud of the blending region in the reference, aligning with the point cloud of the blending region in the original. A Neural Radiance Field learns a mapping function $f$ that outputs density $\sigma$ and radiance $\textbf{c}_{\text{RGB}}$ for a given 3D point $\bm{x} \in \R^3$;
$f(\bm{x}) = (\sigma, \textbf{c}_{\text{RGB}})$. 3D points are sampled along a ray, which depends on the camera pose. Let us assume that we define the matrix $\mathbf{W}$ as reducing scale. We cannot directly reduce the object scale in NeRF because the mapping function $f$ is fixed. 
Instead, we can sample points more widely to reduce the object scale relatively. Renewed density $\sigma'$ and radiance $\textbf{c}'_{\text{RGB}}$ can be computed using the inverse matrix of $\mathbf{W}$ as follows $f(\mathbf{W}^{-1} \bm{x}) = (\sigma', \textbf{c}'_{\text{RGB}})$.

\paragraph{Ablation study for local alignment.}

Pose alignment is a critical part, but it alone might not be enough to handle loosely aligned images such as AFHQv2-Cat. \reffig{ablationWarping} shows an ablation study of our 3D local alignment method. The cat in the original image has a smaller face than the reference cat. If we just apply pose alignment only, the blending result (bottom left in \reffig{ablationWarping}) will create a much bigger face than humans normally expect. After locally aligning the face of the reference, we can obtain a natural-looking result with proper scale (bottom right in \reffig{ablationWarping}). \reffig{suppablationWarping} shows additional ablation results.
We further examine local alignment by conducting a user study. In $\mathbf{60.3}\%$ of cases, users prefer our blending results with local alignment compared to those without alignment. Please see the details of the user study in \refapp{suppuserstudy}.


\suppfigchoiceBlending

\paragraph{Runtime.}
Pose alignment takes 1.7s, and local alignment takes 4.2s, 4.6s, 5.9s for ears, eyes, face in AFHQv2-Cat~\cite{choi2020starganv2}. 
The larger blending part takes more time to align. Our local alignment implementation uses the \texttt{Trimesh}\footnote{https://trimsh.org/trimesh.html} library, which provides pre-defined functions for triangular meshes on the CPU. Implementing it with GPU-based libraries can further reduce alignment runtime.

\section{Details of 3D-aware Blending}
\label{sec:blendingDetails}

\paragraph{Choice for the optimization space.} In our blending method, we optimize the latent code $\w{} \in R^{14\times512}$ in the extended latent space $\mathcal{W}+$~\cite{abdal2019image2stylegan}. There are other options for optimization, such as an intermediate latent space $\mathcal{W}$~\cite{karras2019stylegan} or generator $\G{}$. \reffig{choiceBlending} shows the comparison among optimization spaces in 3D-aware blending. $\mathcal{W}$ space shows realistic blending results but lacks faithfulness to the reference object. The blended image can not capture the green eyes of the reference object, as shown in the first row of \reffig{choiceBlending}.
Optimizing $\G{}$ shows faithfulness to the reference object but lacks realism. The boundaries of results look particularly odd. $\mathcal{W}+$ space shows great results in both realism and faithfulness, our method adopts to optimize the $\mathcal{W}+$ latent space.

\paragraph{Masks used in alignment and blending.}

At first, we apply \textit{pose alignment} to rotate the reference to match the pose of the original. A user selects a mask $\m$ for the target blending region of the original image. The user selects another mask  $\m'$ for the target blending region of the reference image, or it can be derived automatically using previous works~\cite{zhu2016generative,HORN1981185}.
We blend images using the union of the masks, $\m \cup \m'$, and we redefine $\m$ as this union in
\ifdefined\ARXIV
   \reffig{blending}.
\else
   Figure 4 in the main paper.
\fi
Before blending in AFHQ, $\m$ and $\m'$ are used for \textit{local alignment} to align target regions of two images, as shown in \reffig{suppfiglocalalign}. 
After local alignment, the target reference mask $\m'$ is also modified. 

\paragraph{Runtime comparison with baselines.}
Poisson blending~\cite{perez2003poisson}, StyleMapGAN~\cite{kim2021exploiting}, Latent Composition~\cite{chai2021latent} take less than a second to blend images at $1024 \times 1024$ resolution, as they 
use low-level visual cues~\cite{perez2003poisson} or encoders~\cite{kim2021exploiting,chai2021latent}. SDEdit~\cite{meng2021sdedit} and StyleGAN3~\cite{karras2021alias} both require an iterative process. SDEdit takes 29.5s for 500 iterations (face and hair) and 20.0s for 300 iterations (nose, eyes, lip) at $256^2$ image resolution. StyleGAN3 takes 69.8s for $512^2$ and 106.5s for $1024^2$ resolution. Our method takes 26.9s at $512^2$ and we can further reduce the time to 
12.1s by combining ours with Poisson blending
\ifdefined\ARXIV
   (\refsec{oursWpoisson}).
\else
   (Section 3.3).
\fi 
All runtimes are measured in the same device with a single A100 GPU. Ours is faster than other optimization-based baselines but slower than the encoder-based methods. In the future, we can directly train an encoder using our image- and density-blending losses to reduce the runtime.
\suppfigStyleSDFblend
\suppfigStyleSDFdensity
\suppfigStyleSDFimage
\section{3D-aware blending in StyleSDF}
\label{sec:stylesdf}

StyleSDF~\cite{or2022stylesdf} generates high-fidelity view-consistent images in $1024\times1024$ resolution. The main difference between StyleSDF and EG3D~\cite{Chan2022} is StyleSDF uses Signed Distance Fields (SDF) as 3D representation. Besides, StyleSDF does not require camera pose labels to train the generator, unlike EG3D. In our 3D-aware blending method, we can use the SDF value $d$ as a 3D signal, similar to the density $\sigma$ in EG3D. If we assume a non-hollow surface, the SDF value can be converted to the density $\sigma$ as follows:
\begin{equation}
\label{eq::stylesdf}
\begin{aligned}
    \sigma(\bm{x}) = \frac{1}{\alpha} \cdot \text{Sigmoid} ( \frac{-d(\bm{x})}{\alpha}),
\end{aligned}
\end{equation}
where $\bm{x} \in \R^3$ is a 3D location and $\alpha$ is a learned parameter about the tightness of the density around the surface boundary. We use the same blending loss functions used in EG3D, except we replace the density $\sigma$ with SDF $d$.

We demonstrate that our method can be applied to other 3D-aware generative models beyond EG3D. 
\reffig{suppfigStyleSDFblend} shows our 3D-aware blending results in StyleSDF using generated images. \reffig{suppfigStyleSDFdensity} and \reffig{suppfigStyleSDFimage} show ablation studies of our blending loss terms: $\L{density}$ and $\L{image}$, respectively. The ablation studies show a similar tendency with EG3D experiments in the paper. Without the density-blending loss, we cannot blend highly structured objects like hair. If a user does not want to reflect the reference color, we remove or give a low weight to the image-blending loss on reference: $\lambda_2$ in 
\ifdefined\ARXIV
\refeq{color_edit}.
\else
Eqn. 4 in the main paper.
\fi

\section{User study details}
\label{sec:suppuserstudy}
We conduct extensive user studies to show the effectiveness of our methods in the realism score of human perception on the Amazon Mechanical Turk (MTurk) platform. 
We refer to user study pipelines of SDEdit~\cite{meng2021sdedit} and modify the template\footnote{https://github.com/phillipi/AMT\_Real\_vs\_Fake} from the previous work~\cite{zhang2016colorful}. 
The instruction page is shown in \reffig{suppfigAMTintro}, and MTurk workers participate in surveys comparing the results of two methods, as shown in Figures \ref{fig:suppfigAMTbaseline} and \ref{fig:suppfigAMTwarping}.

Each evaluation set consists of 25 pairwise comparison questions, with an additional five questions used to detect deceptive workers. We only invite workers with a HIT Approval Rate greater than $98 \%$. Each set takes approximately 2 to 3 minutes and offers a reward of \$0.5.

\myparagraph{Comparison with baselines} are conducted in 
\ifdefined\ARXIV
   Tables~\ref{tbl:userstudyCelebA} and \ref{tbl:userstudyAFHQ}.
\else
   Tables~3 and 4 of the main paper. 
\fi
\reffig{suppfigAMTbaseline} shows the comparison page. There are two blending results: one for our method and the other for one of the baselines. The order of images is randomly shuffled, and a worker is instructed to select a more realistic image. In CelebA-HQ~\cite{karras2017progressivegan} and AFHQv2-Cat~\cite{choi2020starganv2}, we use 60 and 40 evaluation sets, respectively; CelebA-HQ for 1,500 and AFHQ for 1,000 pairwise comparisons. We set the same number of evaluation sets to report the combination of our method and Poisson blending.

\myparagraph{Ablation study of 3D local alignment} is conducted in \refapp{local_align}. The user study page is shown in \reffig{suppfigAMTwarping}. Experimental settings are almost similar to previous studies: Tables~3 and 4 of the main paper. We show two blending results of our method with and without \textit{local alignment}. Note that pose alignment is used in both of the results. We use 20 evaluation sets in AFHQv2-Cat; 500 pairwise comparisons.

\suppfigAMTintro
\suppfigAMTbaseline
\suppfigAMTwarping

\suppfigOverlap

\section{Failure cases}
\label{sec:limit}

Besides inversion as we described in the main paper, we present other failure cases in our blending method.
\suppfigLimitICP
\reffig{overlap} shows our image blending result from a large mask to a small mask. Undesirable effects (yellow box) of the reference image have been introduced to the final result. One potential solution for future work is to inpaint the original image before blending.
\reffig{suppfigLimitICP} shows another failure case of local alignment in the Iterative Closest Point (ICP) algorithm~\cite{besl1992method,chen1992object}. ICP is an approach to aligning the two point clouds: $\Pt{ori}$ and $\Pt{ref}$ for the original and reference objects, respectively.
It iteratively minimizes the distance between each 3D point in $\Pt{ori}$ and its nearest point in $\Pt{ref}$, but it may fall into the local extremum.
$\Pt{ref}$ shrinks to a point $\tilde{P}_{\text{ref}}$ as discussed in the previous work~\cite{Wang_2019_ICCV}. 
To mitigate this issue, we restrict the minimum and maximum scaling factor to $[0.75,1.25]$. In the future, we may utilize recent pairwise registration techniques~\cite{choy2020deep,Wang_2019_ICCV} instead of ICP for better local alignment.

\section{Societal impact}
\label{sec:societal}

The societal impact of image blending shares similar issues raised in the other generative models and view synthesis techniques~\cite{hancock2021social}. For instance,  a malicious user may use image blending to manipulate the expression and identity of a real person or create a scene that does not exist in real life. Most of the existing watermarking and visual forensics methods focus on 2D image content~\cite{yu2019attributing}. Adopting visual forensics methods for 3D-aware content seems to be an important future work.

\suppfigADDcomparison

\section{Additional results}
\label{sec:suppAdditional}

\paragraph{Comparison with StyleFusion and Barbershop.}
In order to show the advantages of our 3D-aware approach, we compare our method with additional baselines: StyleFusion~\cite{kafri2021stylefusion} and Barbershop~\cite{zhu2021barbershop}. Previous methods struggle to blend fairly misaligned images, and they acknowledged this limitation in their papers.
As shown in \reffig{identity}, latent-based methods often fail to preserve identity, as projecting images into low-dimensional latent space remains challenging. For example, 
 Latent Composition~\cite{chai2021latent} and StyleFusion altered identities from the input images. Other works, such as StyleMapGAN~\cite{kim2021exploiting} and Barbershop, use the spatial latent space to capture image details, but as a trade-off, they are worse at blending misaligned images. In contrast, our method preserves identity while maintaining 3D consistency between parts. Our method addresses these issues by 1) 3D-aware alignment and 2) blending with 3D-aware constraints, including pixel RGBs and volume density from the aligned reference.

 \paragraph{Comparison with SDEdit in AFHQ.}
In addition to comparing CelebA-HQ~\cite{karras2017progressivegan}, we also conduct a comparison with SDEdit using the AFHQv2-Cat dataset~\cite{choi2020starganv2}. However, since pretrained diffusion models on AFHQ are not publicly available, we use the LSUN-Cat model\footnote{https://github.com/openai/guided-diffusion}. \reffig{sdeditAFHQ} shows SDEdit fails to preserve the reference well in both datasets.

\suppfigSDEditAFHQ

\paragraph{More qualitative results.} We show additional qualitative results of our 3D-aware blending. Figure~\ref{fig:eyeglasses} shows blending results of highly structured objects such as eyeglasses. Figures~\ref{fig:suppfigCompareCelebA} and \ref{fig:suppfigCompareAFHQ} show blending comparisons with baselines. Our results show outstanding blending results than other baselines.
By virtue of NeRF, we can synthesize novel-view images. Figures~\ref{fig:suppfigMultiviewCelebA}--\ref{fig:suppfigMultiviewStyleSDF} show multi-view consistent blending results. Please see 
our project \href{https://blandocs.github.io/blendnerf}{page} for the multi-view consistent blending videos.

\suppfigEyeglasses


\clearpage
\suppfigCompareCelebA
\suppfigCompareAFHQ
\suppfigMultiviewCelebA
\suppfigMultiviewShapeNet
\suppfigMultiviewAFHQ
\suppfigMultiviewStyleSDF

\end{document}